\PassOptionsToPackage{table,dvipsnames}{xcolor}
\documentclass{article} %
\usepackage{iclr2027_conference,times}

\usepackage{amsmath,amsfonts,bm}

\def\eqref#1{equation~\ref{#1}}

\def\1{\bm{1}}

\DeclareMathAlphabet{\mathsfit}{\encodingdefault}{\sfdefault}{m}{sl}
\SetMathAlphabet{\mathsfit}{bold}{\encodingdefault}{\sfdefault}{bx}{n}

\usepackage{hyperref}
\usepackage{cleveref} 
\usepackage{url}
\usepackage{xurl}
\usepackage{amsmath,amssymb,amsfonts,amsthm,mathtools}
\usepackage{fvextra,upquote}   
\usepackage[most]{tcolorbox}
\input{promptboxes/style.tex}
\usepackage{listings}
\usepackage{tikz}
\usetikzlibrary{calc}
\usepackage{natbib}
\usepackage{graphicx}
\usepackage{booktabs}
\usepackage{tikz}
\usepackage{subcaption}
\usepackage{multirow}
\usepackage{tabularx}
\definecolor{HeaderGray}{gray}{0.92}
\usepackage{booktabs}
\usepackage{float}
\usepackage{xcolor,colortbl}
\usepackage{subcaption}
\usepackage{wrapfig}
\usepackage{enumitem}
\usepackage{adjustbox}
\usepackage{breqn}

\newcommand{\rev}[1]{\textcolor{black}{#1}}
\newtcolorbox{PromptBox}[2][]{title={#2},#1}

\title{Few-Shot Demonstrations Elicit the Use of In-Context World Representations in LLMs
}

\author{
\hspace{8ex}
Kohsei Matsutani$^{1\,\dagger}$\hspace{3ex}
Gouki Minegishi$^{1}$\hspace{3ex}
Core Francisco Park$^{2\,3}$ \\
\hspace{11ex}
\textbf{Takeshi Kojima}$^{1}$\hspace{4ex}
\textbf{Yusuke Iwasawa}$^{1}$\hspace{4ex}
\textbf{Yutaka Matsuo}$^{1}$ \\[0.4em]
\hspace{8ex}
$^{1}$The University of Tokyo
\hspace{2ex}
$^{2}$Harvard University
\hspace{2ex}
$^{3}$Prior Computers \\[0.25em]
\hspace{15ex}
$^{\dagger}$\texttt{kohsei.matsutani@weblab.t.u-tokyo.ac.jp}
}

\iclrfinalcopy %
\begin{document}

\maketitle
\lhead{Preprint}
\vspace{-2mm}
\begin{abstract}
Large language models (LLMs), when acting as agents, are expected to take observed data in context, infer the latent state space underlying the world, and leverage it for downstream prediction.
However, prior work demonstrated that LLMs struggle to use representations learned in context on a graph tracking task, where the model needs to construct a representation of the graph governing data generation process and use it for subsequent predictions. In this paper, we show that extending this to few-shot settings, where each demonstration is generated from a different world with either the same or different graph topologies, enhances its prediction on \rev{9 models from 7 model families}. To understand this improvement, we linearly probe a low-dimensional world representation that encodes graph information in the hidden states. Notably, we find that few-shot demonstrations relocate the world representation and increase its predictive use. Specifically, for each model, these world representations shift in directions nearly orthogonal to their original subspace, and interventions on these representations selectively impair performance more than interventions on other subspaces.
Consistent with this insight, we show that few-shot demonstrations with observations from different worlds improve performance on ARC-AGI-1\&2, web agent tasks, and Othello. Our findings elucidate the role and internal mechanisms of few-shot demonstrations in in-context world modeling and provide implications for their further improvement.

\end{abstract}

\section{Introduction}

As large language models (LLMs) are increasingly deployed as agents in open-ended environments, they must infer previously unseen world, such as the semantics of entities and the rules governing the data generation process, from observations, and take optimal actions based on their inferred model of the world \citep{andreas2024worldmodels,li2025does}.
For example, in coding, an agent may need to infer the structure of unfamiliar codebases and the semantics of their APIs from the surrounding code and execution feedback \citep{jimenez2024swebench,yang2024sweagent}. In-context learning (ICL) \citep{brown2020language}, which enables adaptation to the world from observed data without updating any parameters, is important to understand especially from a continual learning perspective \citep{wang2024comprehensive}.
We refer to this general problem setting—inferring a world from observations and performing a potentially different task over the inferred world—as \textit{in-context world modeling}.

\begin{figure}[t]
  \centering
  \includegraphics[width=\linewidth]{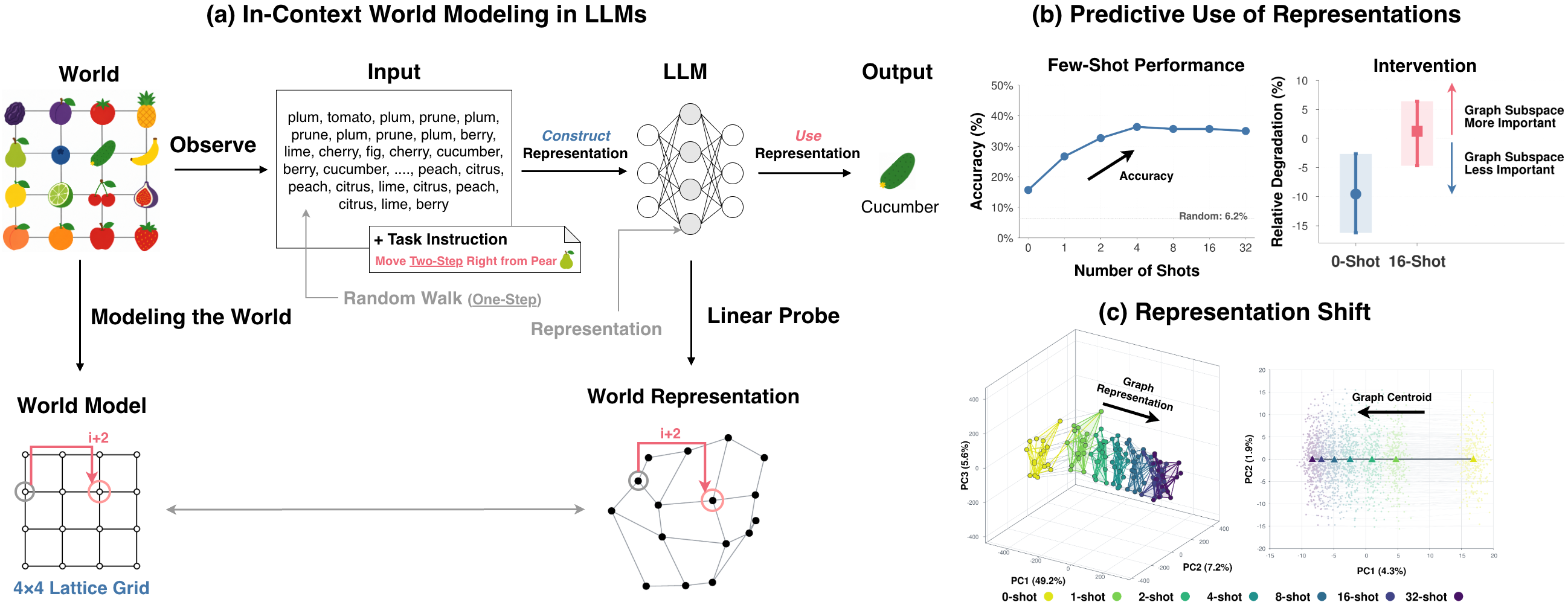}
  \caption{\textbf{(a) Conceptual Framework of In-Context World Modeling.} An LLM is provided with random walks generated on a graph topology, where each node is assigned a fruit, and instructed to move two steps to the right. We probe the internal representations to determine whether the model has constructed the latent graph. A linearly decodable subspace of these representations is termed a world representation. See \Cref{fig:prompt-graph-prediction} for the actual prompt. In this paper, we treat cucumbers as fruits. \textbf{(b) LLMs Learned to Use World Representation with Few Shot Demonstrations.} Few shot demonstrations consisting of random walks, task instructions, and answers from worlds with the same topology but different assignments of fruits to nodes improve task performance. Interventions on the world representation subspace cause larger performance degradation in the few shot setting than in the zero shot setting relative to interventions in random directions. The results are from Qwen3-4B-Base on a $4 \times 4$ Lattice Grid. \textbf{(c) Few-Shot Demonstrations Shift World Representations.} As the number of shots increases, the graph representation shifts orthogonally to the representation subspace. The left shows a 3D PCA of hidden states of the nodes, and the right shows a 2D PCA of mean hidden states over the random walks in the test set. The results are from \Cref{fig:shot-vector-illustration}.}
  \vspace{-4mm}
  \label{fig:fig1}
\end{figure}

One possible mechanism underlying in-context world modeling is the formation of \textit{in-context representations}. \citet{park2025iclr} showed that scaling the context length enables LLMs to construct in-context representations with novel semantics. Specifically, LLMs are given, in context, a one-step random walk generated from a latent graph (a $4 \times 4$ Lattice Grid in \Cref{fig:fig1} (a)) and exhibit energy minimization that reflects the topology in their intermediate states. However, \citep{lepori2026language} argued that even when such representations are successfully constructed, LLMs still fail to solve subsequent next-token prediction tasks defined over the graph. Concretely, after observing the random walk generated from a latent graph, LLMs are given few-shot examples of predicting the node two steps to the right of a given node and are then asked to make the same prediction for a new node. They showed that both open-source and closed-source models fail on this task.
This gap between world representation construction and predictive use of the representation exposes a fundamental limitation of LLM agents that rely on in-context memory without explicit state tracking \citep{lepori2025racing,baldelli2026llms,mozer2026topological}, raising the question of when they can overcome this limitation and why.

In this paper, we show that extending the zero-shot setting of \citet{lepori2026language} with few-shot demonstrations enables LLMs to solve this in-context world modeling task, as shown in \Cref{fig:fig1} (b). Each demonstration contains a random walk generated from a different world that shares the same or different topology and assigns a different vocabulary to its nodes (see \Cref{fig:prompt-illustration} for the prompt illustration).

Motivated by these findings, we use linear probing to analyze a low-dimensional subspace encoding the latent world topology. We find that increasing the number of few-shot demonstrations improves task performance, while both probing accuracy and the Dirichlet Energy (DE) computed from the raw hidden states remain unchanged. This suggests that LLMs construct the world representation equally well regardless of the number of shots (demonstrations). Instead, we find that these representations occupy different subspaces, which influences how they are read out. As the number of shots increases, the representation shifts toward directions roughly orthogonal to itself and progressively converges (\Cref{fig:fig1} (c)). We further intervene on the world representation and find that such interventions degrade predictions more with few-shot demonstrations than in the zero-shot setting. This provides causal evidence that world representations contribute more strongly to prediction with few-shot demonstrations, alongside their systematic relocation across subspaces. Finally, we show that providing observations from different worlds coupled with their task instructions as few-shot demonstrations improves task performance on ARC-AGI-1\&2, web agent tasks, and Othello, using Qwen, Gemma, and Ministral models.

In contrast to conventional few-shot learning, where each demonstration typically specifies an input-output mapping, as in regression \citep{garg2022what,oswald2023transformers} and classification \citep{min2022rethinking}, where a canonical example is predicting ''red'' from ''apple'' \citep{hendel2023context}, we consider an in-context world modeling setting for LLM agents, where each demonstration consists of observations from a world and a task defined over its latent structure. Crucially, our analysis reveals that successful prediction is associated not with improved construction of the world representation itself, but with its relocation to a subspace that elicits downstream readout.

More broadly, we hope that our findings provide support for a growing paradigm of agentic training that leverages ICL \citep{song2026expanding,hubotter2026reinforcement,zhao2026selfdistilled,shi2026experiential,ye2026onpolicy,ye2026online,lin2026scaling,song2026reward}, in which agents condition on observations, trajectories, and outcomes collected through interactions with the environment and generate further rollouts, enabling reinforcement learning (RL) or self-distillation \citep{shenfeld2026selfdistillation} without relying on an external teacher model. Our results may also help clarify the mechanisms underlying such transfer of capabilities acquired in context into model weights.

This paper is organized as follows. In \Cref{section:preliminaries}, we explain in-context world modeling, introduce the graph tracking task, and describe the probing, representation analysis, and intervention methods used throughout the paper. In \Cref{subsection:construction}, we show that LLMs construct representations of the latent topology regardless of the number of shots. In \Cref{subsection:relocation}, we show that the world representation subspace shifts as the number of shots increases. \rev{In \Cref{subsection:intervention}, we perform interventions on representations and analyze representations and attention across different numbers of shots.} In \Cref{subsection:real-task}, we show that these few-shot demonstrations improve performance on ARC-AGI-1\&2, web agent tasks, and Othello. Finally, in \Cref{section:discussion}, we discuss broader connections to related work, limitations, and directions for future work. We cover related work in \Cref{appendix:related-work}.

\section{Preliminaries}\label{section:preliminaries}

\subsection{World Representation in LLMs}
\textbf{In-Context World Modeling.}
As written in \Cref{appendix:related-work}, the definition of a ``world model'' varies across the literature. In this paper, following \citep{li2025does}, we say that an LLM contains a world model if a relevant property of the underlying world from which the data are generated can be recovered from its internal representations and those representations causally contribute to the model's predictions. We refer to such internal representations as \textit{world representations}. From this perspective, we define \textit{in-context world modeling} as inferring a world from contextual observations and performing a different task over it. See \Cref{fig:fig1} (a) for the illustration. We identify world representations using linear probing \citep{alain2018understanding,belinkov2022probing} and test whether they are causally used for prediction by intervening on them and measuring the resulting change in prediction \citep{elazar2021amnesic}. We refer to the former as \textit{representation construction} and the latter as \textit{predictive use of representation}. See \Cref{appendix:world-model} for a detailed explanation.

\textbf{Graph Tracking.}
We adapt the \textit{adaptive world modeling} task of \citet{lepori2026language}, a variant of the \textit{in-context graph tracing} task introduced by \citet{park2025iclr}. As illustrated in \Cref{fig:fig1} (a), each node in a graph topology, specifically $4 \times 4$ Lattice Grid, $5 \times 5$ Lattice Grid, $3 \times 5$ Lattice Grid, $4 \times 4$ Triangle Grid, and $4 \times 4$ Torus Grid is assigned a fruit vocabulary item such as ''apple'' or ''orange''. We regard each such assignment as a world. Even with the same topology, different assignments of fruit to the nodes define different worlds. The model is then given, as observations, a comma-separated sequence of fruit tokens generated by a one-step random walk over the graph. Following these observations, the model receives few-shot examples consisting of node pairs that specify a two-step movement to the right on the graph, such as ''lime'' to ''fig''. Although this transformation could instead be specified in natural language as ''move two steps to the right,'' the meaning of ''right'' changes under rotations or reflections of the topology. We therefore follow \citet{lepori2026language} and specify the task through few-shot node pair examples. 
We view this setting as \textit{in-context world modeling}, where a model infers a graph topology from observations provided in-context and performs a task on that graph. Typical ICL as in \citet{park2025iclr} is a special case where the task providing information about the world is identical to the task being tested.
The length of random walk is sampled uniformly from 48 to 80. See \Cref{appendix:systhetic-task} for the detailed setup and \Cref{fig:prompt-graph-prediction} for an example prompt.
\subsection{Experimental Design}\label{subsection:experimental-design}

\begin{wrapfigure}[18]{r}{0.27\textwidth}
  \centering
  \includegraphics[width=\linewidth]{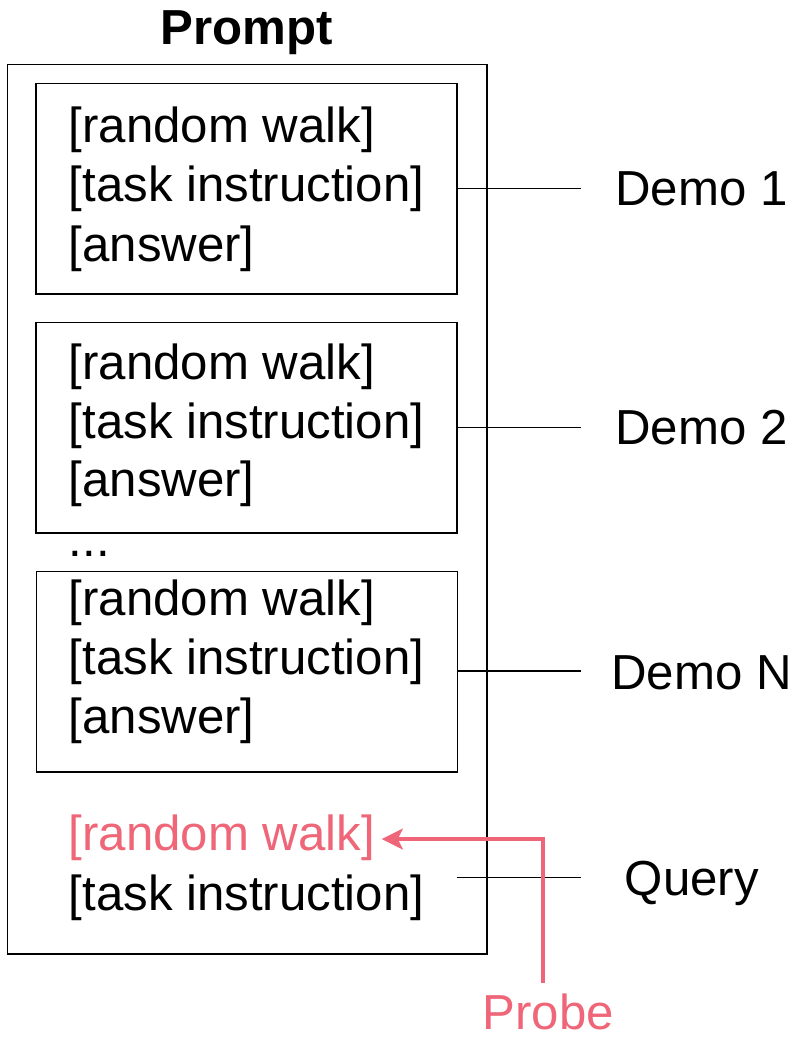}
  \caption{\textbf{Illustration of Few-Shot Prompt.} See \Cref{fig:prompt-graph-prediction} for the prompt example.}
  \label{fig:prompt-illustration}
\end{wrapfigure}

\textbf{Few-Shot Prompting.}
As illustrated in \Cref{fig:prompt-illustration}, each few-shot demonstration consists of observation data and a task instruction, namely a random walk generated from a latent graph and a set of node pairs specifying a two-step movement to the right. Note that these few-shot demonstrations are distinct from the few-shot node pairs used to specify the task instruction. We refer to each demonstration as one shot. All shots share the same graph topology but are generated from different worlds, with different assignments of fruits to the nodes. Across different shot settings, we keep the final query fixed to enable a controlled comparison. The zero-shot setting corresponds to \citet{lepori2026language}. See the prompt example in \Cref{fig:prompt-graph-prediction}.

\textbf{Probe.}
We use linear probing to quantify how well LLMs construct the latent topology from the observed \textcolor[HTML]{EE6677}{random walk in the query} (\Cref{fig:prompt-illustration}). For each node $(i,j)$, we obtain its layer $l$ hidden state $h^{(l)}_{i,j}\in \mathbb{R}^{d}$ by averaging the activations over its subtokens and all occurrences in the random walk of the query. We train a probe $W^{(l)} \in \mathbb{R}^{d \times r}$ on 500 samples and use disjoint 100 samples for validation. Assuming that each graph topology admits a low-dimensional representation \citep{shai2024transformers,tigges2024llm,engels2025not}, we set $r=2,2,2,2,4$ for $4 \times 4$ Lattice Grid, $5 \times 5$ Lattice Grid, $3 \times 5$ Lattice Grid, $4 \times 4$ Triangle Grid, and $4 \times 4$ Torus Grid, respectively. The probe maps each hidden state to $z^{(l)}_{i,j}={W^{(l)}}^\top h^{(l)}_{i,j}\in\mathbb{R}^r$. For each pair of nodes $u=(i,j)$ and $v=(i',j')$, we predict their adjacency as
$p^{(l)}_{u,v}=\sigma\big(\beta^{(l)}-\lVert z^{(l)}_{u}-z^{(l)}_{v}\rVert_2^2\big),$
where $\sigma(x)=(1+\exp(-x))^{-1}$ and $\beta^{(l)}$ is a trainable scalar offset. We evaluate the probe using the AUROC for predicting whether pairs of graph nodes are adjacent, computed by sweeping the decision threshold and integrating the resulting ROC curve with the trapezoidal rule. Intuitively, the probe learns to project high-dimensional hidden states into two-dimensional coordinates from which the adjacency matrix of the lattice grid can be recovered. See \Cref{appendix:probing-representation-intervention} for the detailed explanation and the rationale behind the experimental setup.
\citet{park2025iclr} used Dirichlet Energy (DE) as a measure of representation construction, computed directly from raw hidden states without probing. We instead introduce probing to enable more detailed representation analyses and intervention experiments. For completeness, we also report DE in \Cref{subsection:construction}.

\textbf{Causal Intervention.}
To test whether the probed world representations are causally used for prediction, we perform intervention experiments comparing zero-shot and few-shot settings. We measure how much prediction degrades when graph information is removed from the rank-$r$ world representation subspace, relative to an intervention on an unrelated rank-$r$ random subspace orthogonal to it.
To remove graph information, we first center each node representation by subtracting the mean representation $\bar{h}^{(l)}$ across all nodes. We then intervene at each layer as $h_{i,j}^{(l)} \leftarrow h_{i,j}^{(l)}-\Pi_{\mathcal W}^{(l)}(h_{i,j}^{(l)}-\bar{h}^{(l)})$, where $\mathcal{W}^{(l)}$ is the subspace spanned by the left singular vectors by the singular value decomposition (SVD) of $W^{(l)}$, which we call \textit{world representation subspace}, and $\Pi_{\mathcal W}^{(l)}$ is the orthogonal projection matrix onto $\mathcal W^{(l)}$. Following \citet{elazar2021amnesic}, we iteratively retrain the probe and apply the intervention. This is necessary because the intervened subspace is extremely low-dimensional, with $r=2, 4$ compared with ambient dimensions $d=2560,4096,5120$, so a single projection may leave graph information outside the identified subspace that can be repaired \citep{mcgrath2023hydra}. To establish a matched baseline for subspace intervention, we compare interventions on the probed world representation subspace with interventions on orthogonal random subspaces of the same dimension. For brevity, we refer to the former as the \textit{graph intervention} and the latter as the \textit{random intervention}. \rev{This comparison aims to isolate the causal contribution of the probed representations while accounting for differences in context, capacity for repair, intervention magnitude, and baseline correct answer logits between zero-shot and few-shot settings.}
We emphasize that this comparison controls for nonspecific random subspace removal, rather than for interventions on other task-relevant representations. 

See \Cref{appendix:probing-representation-intervention} for the detailed explanation, \Cref{appendix:rank} for the experiments on different rank-$r$, and \Cref{appendix:seed} the sensitivity to seeds.

\section{Few-Shot Demonstrations Improve Downstream Predictions}\label{section:few-shot}

\begin{figure}[t]
  \vspace{-3mm}
  \centering
  \includegraphics[width=\linewidth]{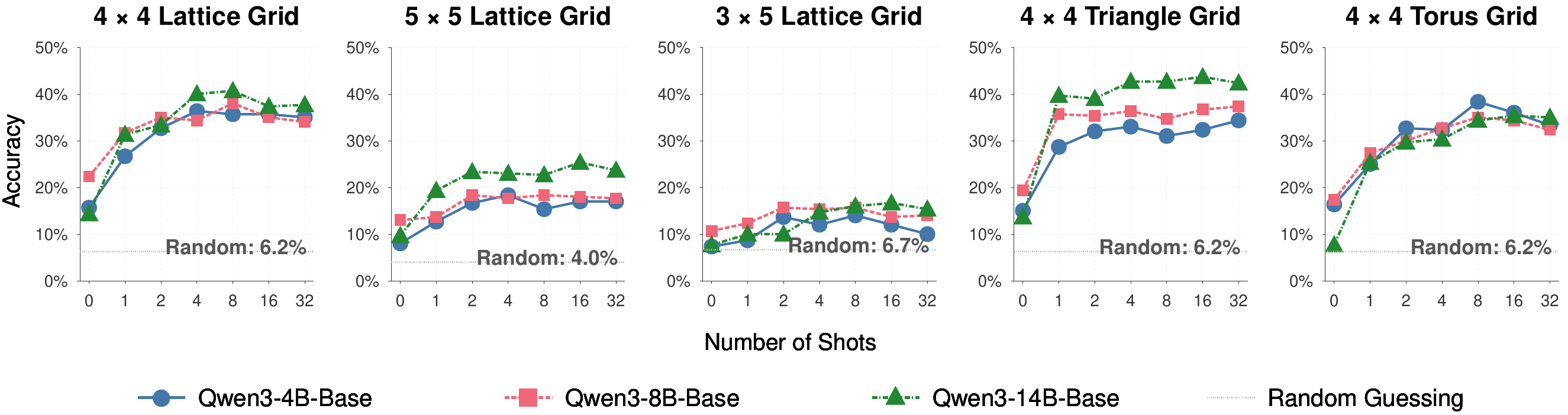}
  \caption{\textbf{Few-Shot Performance.} Accuracy of three Qwen3 models on the two-step-right prediction task across five graph topologies. Each point reports accuracy on 300 test samples with 0, 1, 2, 4, 8, 16, or 32 few-shot demonstrations. Dotted gray lines indicate random guessing over the candidate nodes for each topology. \citet{lepori2026language} corresponds to the zero-shot case.}
  \label{fig:few-shot-performance}
  \vspace{-3mm}
\end{figure}

In this section, we show that, in the graph tracking task, increasing the number of demonstrations improves downstream performance in predicting the node two steps to the right. We evaluate Qwen3-4B-Base, Qwen3-8B-Base, and Qwen3-14B-Base \citep{yang2025qwen3} on the $4 \times 4$ Lattice Grid, $5 \times 5$ Lattice Grid, $3 \times 5$ Lattice Grid, $4 \times 4$ Triangle Grid, and $4 \times 4$ Torus Grid using 300 samples per graph. We consider 0, 1, 2, 4, 8, 16, and 32-shot settings while keeping the query fixed across settings. Predictions are scored by selecting the node with the highest logit among the candidate node set. See \Cref{appendix:systhetic-task} for details.

\Cref{fig:few-shot-performance} shows performance across different numbers of shots. Across all models and topologies, increasing the number of few-shot examples improves performance relative to the 0-shot setting. The performance gains gradually saturate as the number of shots increases, consistent with prior observations in many-shot ICL \citep{yuan2024focused,zhang2025more}. 
Note that the \textit{Metalearning AWM Prompt} of \citet{lepori2026language} uses up to three demonstrations with the same topology and identical node assignments, which might allow copying or elimination across shots. In contrast, our demonstrations come from different worlds that share the same topology but use different node assignments. See \Cref{appendix:other-families} for the results from the Llama-3 \citep{grattafiori2024llama3}, Gemma-4 \citep{gemmateam2026gemma4}, and Ministral-3 \citep{liu2026ministral3} models.

More strikingly, in \Cref{appendix:few-shot-performance}, we further show that few-shot demonstrations improve performance even when the graph which generate their random walks differ from that of the query.
\rev{We discuss other prompting baselines in \Cref{appendix:few-shot-ablation}, evaluations on closed models in \Cref{appendix:few-shot-performance-closed-models}, and performance improvements by candidate elimination in \Cref{appendix:candidate-selection}.}

\section{Few-Shot Demonstrations Shift World Representations}\label{section:representation}
\begin{wrapfigure}[14]{r}{0.52\textwidth}
  \centering
  \vspace{-9mm}
  \includegraphics[width=\linewidth]{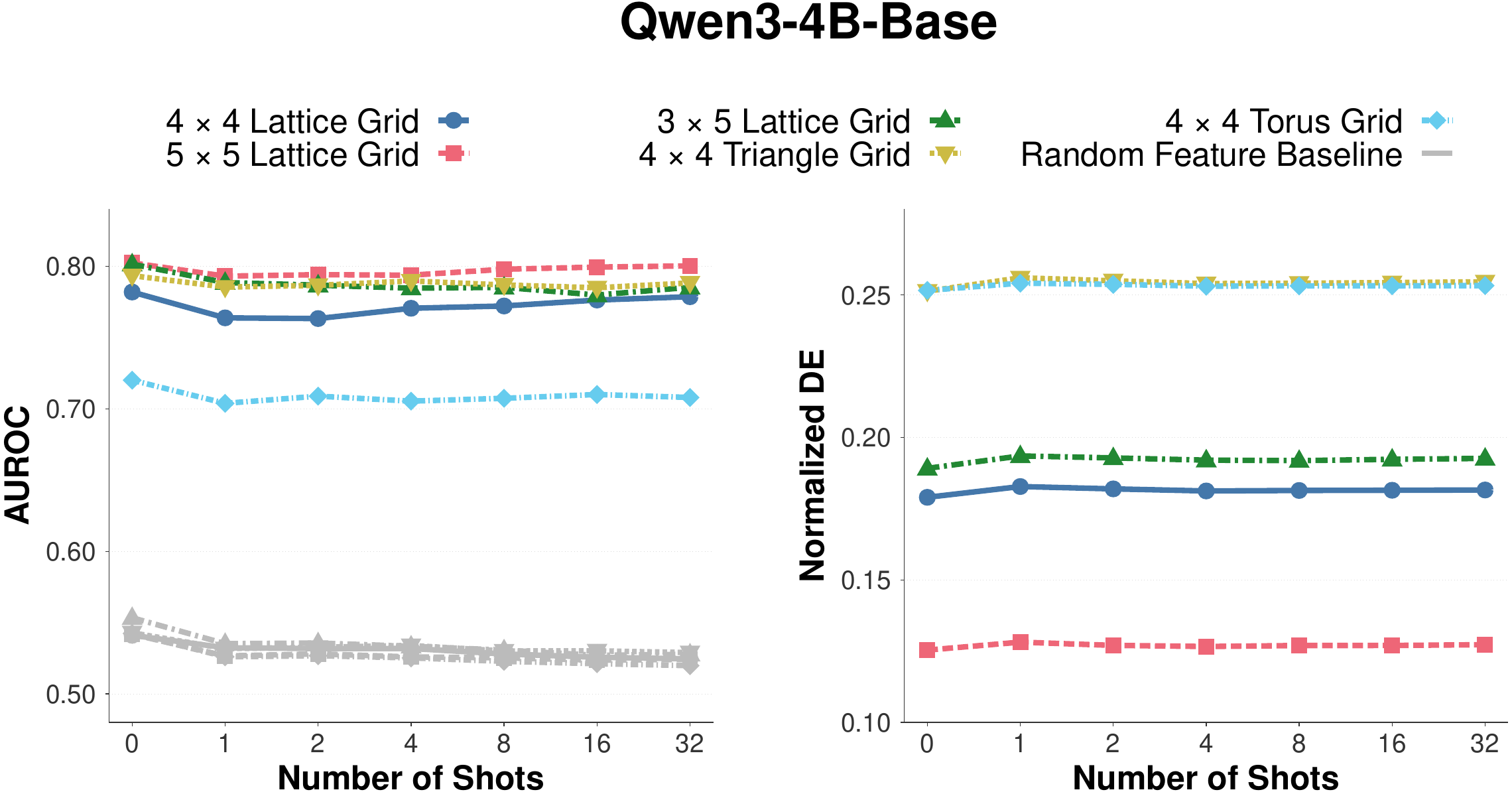}
  \caption{\textbf{Probe Performance and Normalized Dirichlet Energy (DE).} Probe AUROC and normalized DE of hidden states across shots and five graph topologies for layer 20 of Qwen3-4B-Base. Gray lines show random feature probe baselines.}
  \label{fig:construction}
\end{wrapfigure}
\subsection{LLMs Construct World Representations}\label{subsection:construction}
We first show through linear probing that LLMs construct the graph topology underlying the random walk equally well regardless of the number of shots.

As described in \Cref{subsection:experimental-design}, we measure whether LLMs construct the graph by evaluating graph adjacency prediction with AUROC. We also report Dirichlet Energy (DE) \citep{park2025iclr}, which measures the alignment between node representations and the graph structure. At layer $l$, DE is computed as $\sum_{u}\sum_{v}A_{u,v}\lVert h^{(l)}_{u}-h^{(l)}_{v}\rVert_2^2,$ where $u=(i,j)$, $v=(i',j')$, and $A_{u,v}$ denotes the adjacency matrix entry between nodes $u$ and $v$. We report normalized DE by dividing this quantity by the sum of squared Euclidean distances between the hidden states of all node pairs.

\Cref{fig:construction} shows that probing AUROC and normalized DE remain largely unchanged across different numbers of shots for all three models and five topologies. As a random feature baseline, we replace the activations with isotropic Gaussian over five random seeds, which yields AUROC values around 0.5. The clearly higher AUROC indicates that the latent graph topology is recoverable from their internal states. Together, these results suggest that LLMs construct an internal representation of the world behind the random walk in the query regardless of the number of shots.
We additionally show in \Cref{fig:relative-node-position} that the relative positions of nodes in the graph representation of a query do not change much as the number of shots increases.
\begin{figure}[t]
  \vspace{-3mm}
  \centering
  \includegraphics[width=\linewidth]{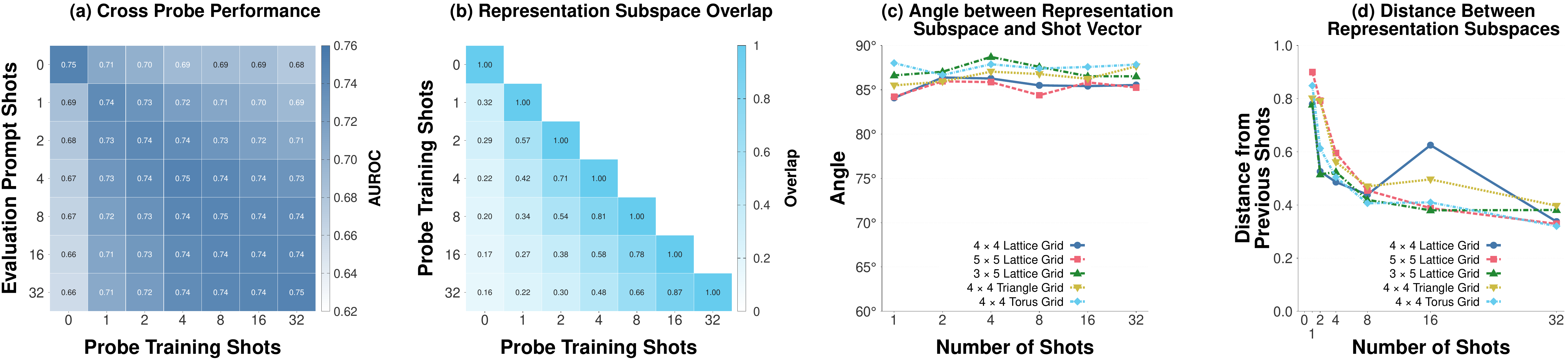}
  \caption{\textbf{(a) Cross Probe Performance.} AUROC of probes trained at one shot count and evaluated at another, averaged over five topologies. \textbf{(b) World Representation Subspace Overlap.} Overlap between probe subspaces across shot counts, averaged over the five topologies. \textbf{(c) Angle Between Shot Vectors and the World Representation Subspace.} Angles to the corresponding probe subspaces, shown separately for each topology. \textbf{(d) Distance between Representation Subspaces across Shots.} Normalized projection distance between probe subspaces at consecutive tested shot counts, shown separately for each topology. All results are from layer 36 of Qwen3-4B-Base.}
  \label{fig:relocation}
\end{figure}
\subsection{Few-Shot Demonstrations Relocate World Representations}\label{subsection:relocation}
\begin{wrapfigure}[18]{r}{0.55\textwidth}
  \vspace{-3mm}
  \centering
  \includegraphics[width=\linewidth]{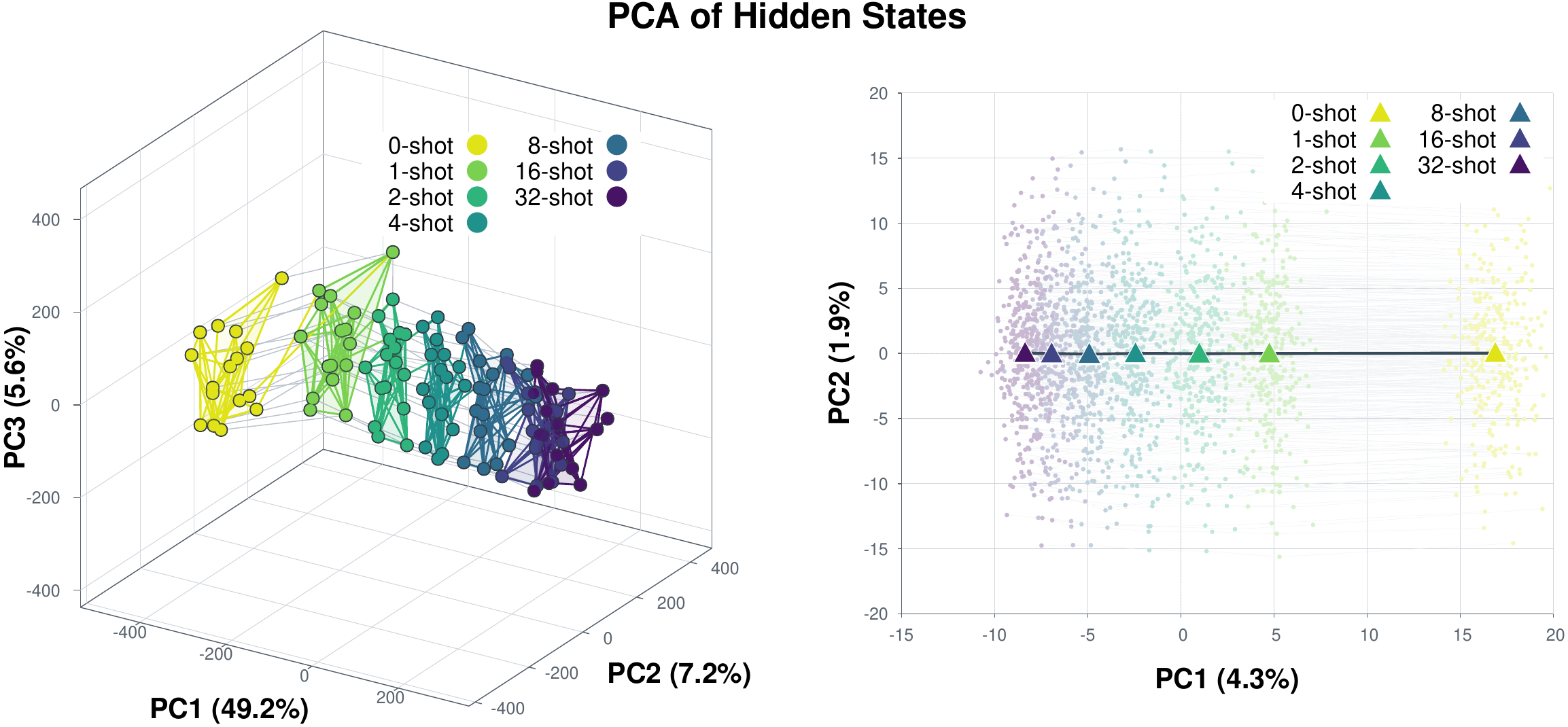}
  \caption{\textbf{(Left) PCA of Shift in Hidden States of a Random Walk.} PCA of $4 \times 4$ Lattice Grid at layer 36 of Qwen3-8B-Base \textbf{(Right) PCA of Shifts in Hidden-State Centroids across Random Walks.} PCA of 300 test samples per shot for $4 \times 4$ Lattice Grid at layer 20 of Qwen3-8B-Base. Points represent individual random walks and triangles indicate their means. In each panel, PCA is fit jointly to all activations shown.}
  \label{fig:shot-vector-illustration}
\end{wrapfigure}
Here, we provide lines of evidence that the world representation shifts as the number of shots increases.

For three Qwen3 models, we train probes at three layers under 0, 1, 2, 4, 8, 16, and 32-shot settings, using 500 samples for training and 100 for validation. We then evaluate cross-shot probing on a 300 sample test set and report the AUROC in \Cref{fig:relocation} (a). Probes trained on few-shot representations fail to correctly decode graphs from zero-shot representations, and decoding performance also degrades as the gap between the numbers of shots increases. This indicates that changing the number of shots changes the probe readout space, at least in part, toward different directions even for the same query. See \Cref{fig:cross-probe} for the full results.

We further measure the overlap between world representation subspaces $\mathcal{W}^{(l)}$ and ${\mathcal{W}'}^{(l)}$ learned by probes trained with different numbers of shots, by computing $\operatorname{tr}(\Pi_{\mathcal{W}}^{(l)}{\Pi'}_{\mathcal{W}'}^{(l)})/{r}.$
As shown in \Cref{fig:relocation} (b), the subspace overlap decreases as the number of shots changes, with a particularly large discrepancy between zero-shot and few-shot settings. This suggests that the linearly decodable graph representation subspace shifts as the number of shots increases. See \Cref{fig:subspace-overlap} for the full results.

Then, we compute the mean shift $\Delta \bar{h}^{(l)}$ between corresponding few-shot and zero-shot hidden states, averaged over all nodes in the test set. We then measure its angle to the world representation subspace $\mathcal{W}^{(l)}$ as $\theta^{(l)} = \arccos(\|\Pi_{\mathcal{W}}^{(l)} \Delta\bar{h}^{(l)}\|_2 /{\|\Delta\bar{h}^{(l)}\|_2 }).$ As shown in \Cref{fig:relocation} (c), the hidden states shift induced by increasing number of shots, where we refer to the shift $\Delta \bar{h}^{(l)}$ as the \textit{Shot Vector}, is nearly orthogonal to the world representation subspace. See \Cref{fig:fs-zs} for the full results.
Importantly, this interpretation relies not only on near-orthogonality in the high-dimensional activation space, but also on changes in cross probe performance and subspace overlap, which suggests that few-shot demonstrations alter the geometry of representations rather than simply shifting hidden states along existing subspace directions, with the Shot Vector capturing only this displacement.

\Cref{fig:shot-vector-illustration} visualizes the hidden states of nodes in a random walk for one sample, and the mean hidden states of random walks across 300 test samples, using PCA. As the number of shots increases, the representations systematically shift. The line segments connecting corresponding points, which form part of the Shot Vector, are also aligned in a similar direction. More visualizations are provided in \Cref{fig:pca-3d-example,fig:pca-example}.

Note that \textit{Task Vector} \citep{hendel2023context} is extracted as the hidden state at the final token of a few-shot prompt, whereas \textit{Function Vector} \citep{todd2024function} is constructed by averaging the final token of causally important attention heads across multiple few-shot prompts and summing them. In contrast, our \textit{Shot Vector} is defined as the difference between the centroids of the hidden states corresponding to observation inputs under few-shot and zero-shot settings. We show that the Shot Vector is different from the Task Vector in \Cref{appendix:shot-vector}.

We further compute the distance between world representation subspaces for adjacent shot settings among 0, 1, 2, 4, 8, 16, and 32 shots as ${\lVert \Pi_{\mathcal{W}}^{(l)}-\Pi{'}_{\mathcal{W}'}^{(l)}\rVert_{\mathrm{F}}}/{\sqrt{2r}}$. As shown in \Cref{fig:relocation} (d), this distance decreases as the number of shots increases, which indicates that the world representation subspace gradually converges in distance. See \Cref{fig:distance} for the full results.

All in all, these results suggest that increasing the number of shots induces a shift in the hidden states of the random walk, which we call the Shot Vector. This shift occurs approximately orthogonally to the rank-$r$ world representation subspace, corresponding to a change in the readout space of the world representation. See \Cref{appendix:other-families} for the results from the Llama-3, Gemma-4, and Ministral-3 models.

\section{Predictive Use of World Representations}\label{section:causality}
\subsection{Intervention Experiments}\label{subsection:intervention}
\begin{figure}[t]
  \centering
  \includegraphics[width=\linewidth]{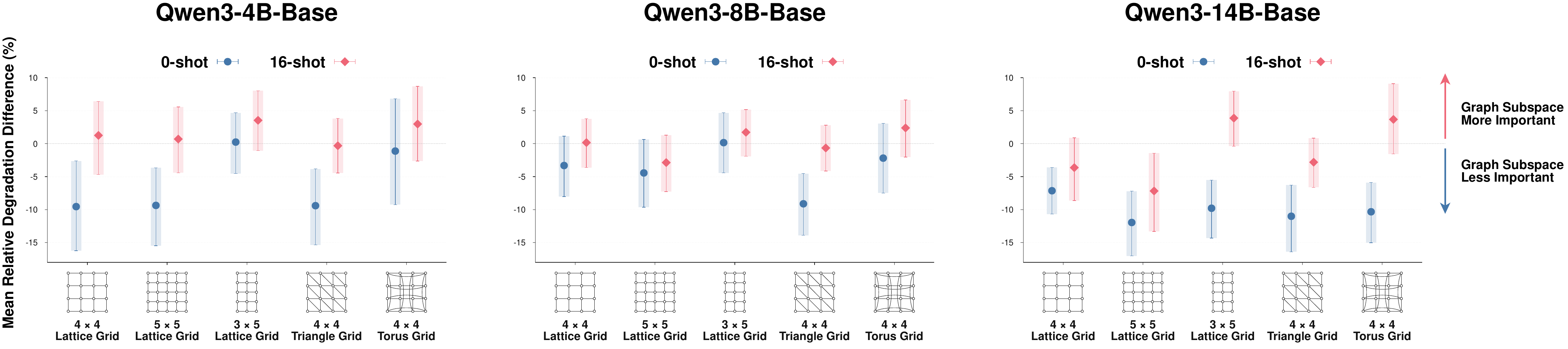}
  \caption{\textbf{Intervention Results.} Graph subspace erasure at iteration 20 in three Qwen3 models. Interventions target node subtokens in the query’s random walk at every Transformer block output. The random intervention uses an orthogonal subspace with matched rank and perturbation norm per node. Points show the mean difference in the probability of correct answer token reduction between graph and random interventions, normalized by each query’s unperturbed probability, across 300 test samples. Probabilities are normalized over candidate answer sequence likelihoods. Positive values indicate greater degradation from graph erasure. Horizontal bars show 95\% bootstrap confidence intervals obtained by resampling test samples 5000 times.}
  \label{fig:intervention}
  \vspace{-4mm}
\end{figure}

Having shown that few-shot demonstrations shift the world representation, we next test whether the shifted representation contributes to subsequent predictions. Following \citet{ravfogel2020null,elazar2021amnesic}, we eliminate graph information from the world representation subspace, from which graph topology can be decoded. To compare graph interventions across zero-shot and few-shot settings, we measure the reduction in the probability assigned to the correct token relative to a intervention
in the random orthogonal subspace.
\begin{figure}[t]
  \centering

  \begin{subfigure}[t]{0.49\linewidth}
    \centering
    \includegraphics[width=\linewidth]{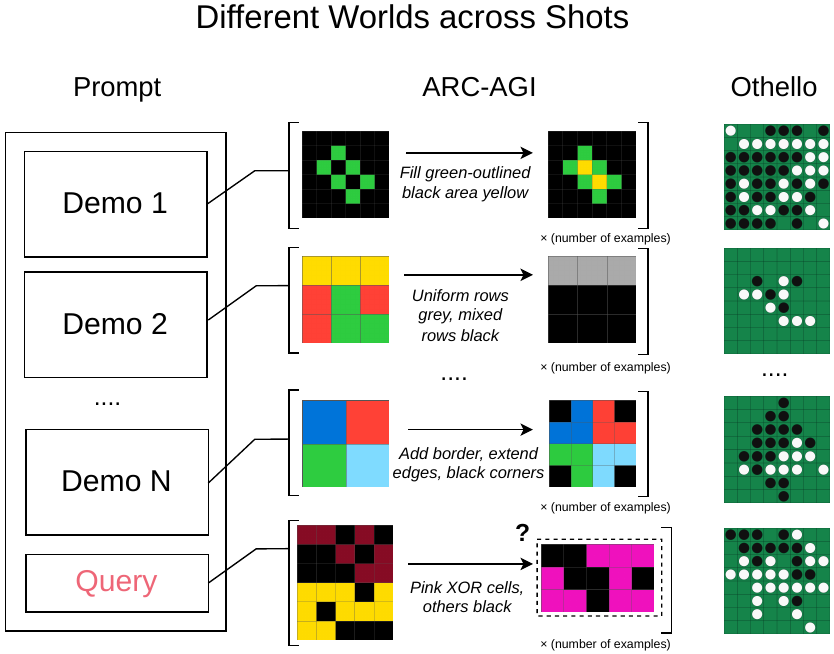}
    \caption{}
    \label{fig:arc-illustration}
  \end{subfigure}
  \hfill
  \begin{subfigure}[t]{0.49\linewidth}
    \centering
    \includegraphics[width=0.95\linewidth]{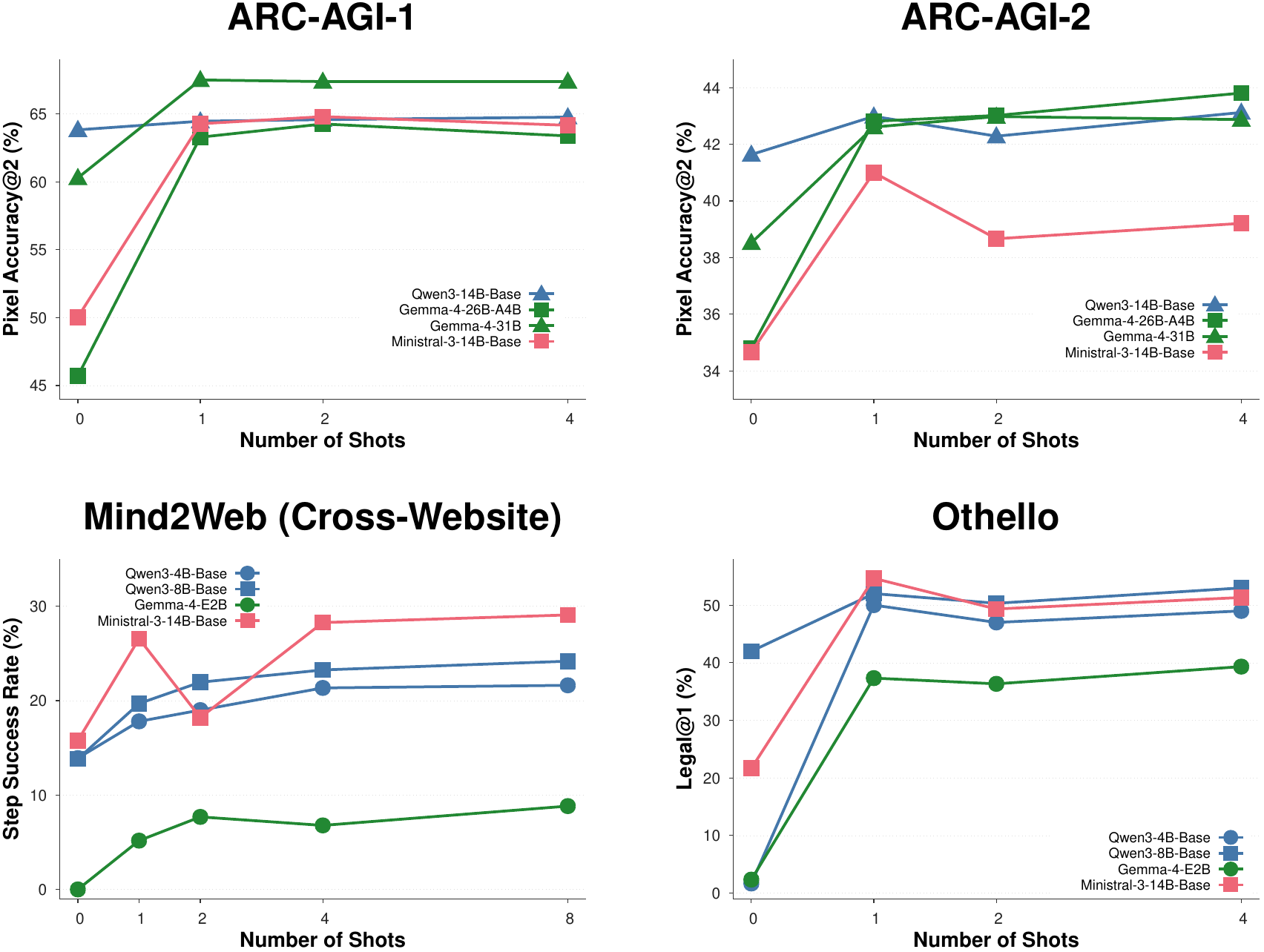}
    \caption{}
    \label{fig:real-task-performance}
  \end{subfigure}
  \caption{\textbf{(a) ARC-AGI and Othello with Different Worlds across Shots.} In ARC-AGI, the world is the rule that maps inputs to outputs. In Othello, the world is the board state. \textbf{(b) Performance across Different Numbers of Shots on ARC-AGI-1\&2, Mind2Web, and Othello.} \rev{For reference, copying the input grid achieves 51.44\% and 57.24\% on ARC AGI 1 and 2, respectively, while filling a same-sized grid with color 0 achieves 37.36\% and 16.53\%. For Othello, uniform sampling from empty squares and all 64 squares yields 41.36\% and 15.89\%, respectively.}}
  \label{fig:real-world}
  \vspace{-5mm}
\end{figure}

\begin{wrapfigure}[15]{r}{0.32\textwidth}
  \vspace{-5mm}
  \centering
  \includegraphics[width=\linewidth]{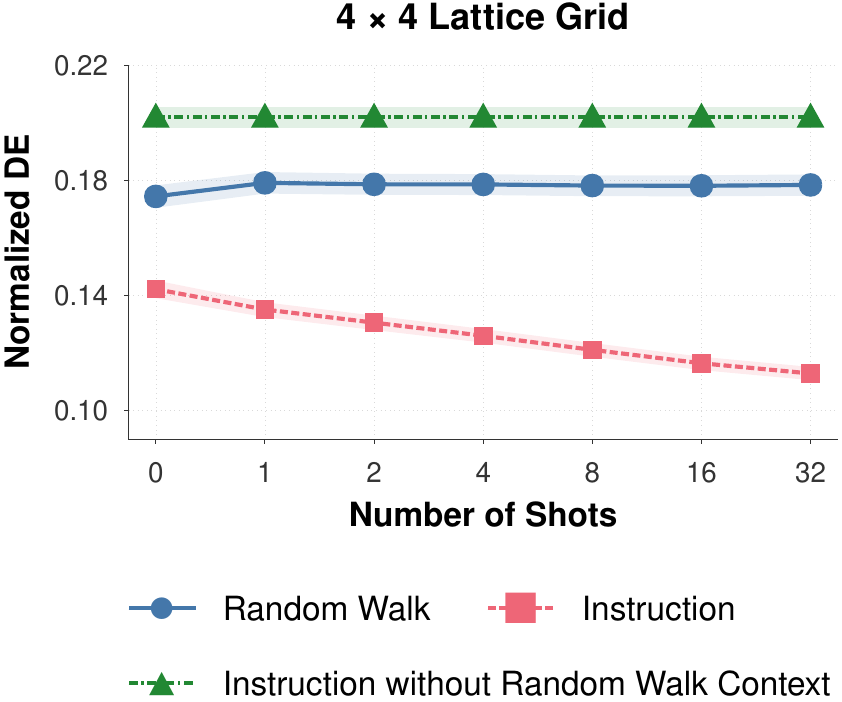}
  \caption{\rev{\textbf{Normalized Dirichlet Energy (DE) of Instruction}. Results from the final layer of Qwen3-4B-Base.}}
  \label{fig:de-instruction-main}
\end{wrapfigure}
This control accounts for differences in baseline task performance and is matched to the graph intervention in the number of activation dimensions perturbed and all other intervention settings.

\Cref{fig:intervention} shows the difference in the degradation of the probability assigned to the correct token between the graph intervention and the random intervention. Larger values indicate that intervening on the world representation subspace causes greater degradation than the random intervention. Compared with the zero-shot setting, this difference increases when few-shot demonstrations are provided. Because representational repair can occur under both interventions, their difference captures the additional degradation from intervening on the graph representation subspace. This result provides evidence that few-shot demonstrations elicit greater, though still partial, predictive use of world representations that encode graph information. See \Cref{appendix:intervention} for the detailed results and \Cref{appendix:other-families} for the results from the Llama-3, Gemma-4, and Ministral-3 models.
\begin{wrapfigure}[13]{r}{0.53\textwidth}
  \centering
  \includegraphics[width=\linewidth]{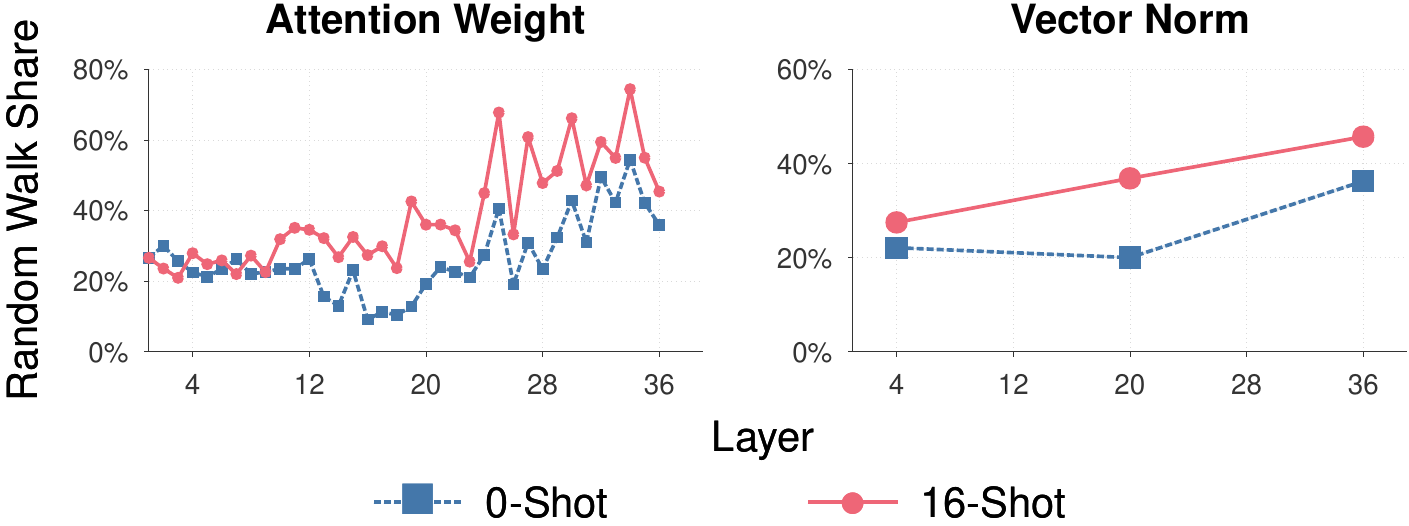}
  \caption{\rev{\textbf{Random Walk Share of Attention Weights and Vector Norms}. Results of Qwen3-4B-Base on $4 \times 4$ Lattice Grid.}}
  \label{fig:attention-main}
\end{wrapfigure}

\rev{\Cref{fig:de-instruction-main} presents the normalized DE of the fruit vocabulary in the few-shot example (two-step move) instructions. While the normalized DE for random walks remains largely constant regardless of the number of shots, that for the instructions decreases. This suggests that providing few-shot demonstrations encourages the task instruction nodes to align more closely with the latent topology formed by the random walks.
See \Cref{fig:de-instruction} for detailed results, and \Cref{fig:cross-probe-instruction} for the cross probe performance of probes trained on random walks and evaluated on instructions.}

\rev{Moreover, \Cref{fig:attention-main} shows the \textit{random walk share} of attention weights and the norms of attention-weighted value vectors (vector norms, cf. \citet{kobayashi2020attention}). The random walk share is the fraction of these values allocated to the random walk out of those allocated to the random walk and instruction in the query, measured at the final input token position. This suggests that few-shot demonstrations shift attention relatively away from the instruction and toward the random walk in the middle to later layers.}
\rev{See \Cref{appendix:attention} for the settings and the results of other models.}

\subsection{Experiments on Practical Tasks}\label{subsection:real-task}

Finally, we demonstrate that the effectiveness of few-shot demonstrations generalizes beyond graph tracking to ARC-AGI-1\&2, web agent tasks, and Othello. 

The ARC-AGI-1 \citep{chollet2019on} test set consists of 400 tasks and 419 test pairs, while ARC-AGI-2 \citep{chollet2026arcagi2} consists of 120 tasks and 167 test pairs. Each task provides input--output pairs labeled as ''Examples.'' These pairs constitute observations from which the model must infer the underlying rule and apply it to a new input. Here, the underlying rule corresponds to the world, which determines the input--output data generation process. For each query in the test set, we provide few-shot demonstrations from different rules (worlds), randomly sampled from the training set. Because the task is challenging for the models evaluated here, exact-match accuracy (pass@2) is sparse. We therefore use pixel-level accuracy@2 as a more sensitive measure of partial task performance. For the web agent task, we use the Mind2Web dataset \citep{deng2023mind2web}, which contains 1373 action steps across 177 Cross-Website tasks. Step success rate is computed for each task and then averaged across tasks. Given a website's HTML, an interaction trajectory, and a task instruction, the model must infer the appropriate action. Here, the states and structure of the website underlying the HTML correspond to the world. We randomly sample few-shot demonstrations from the train set and use the test set for queries. By using cross-website subset, we ensure that the websites (worlds) differ from those in the few-shot demonstrations. For the Othello task, following \citet{li2023emergent,nanda2023emergent}, we represent a game as a sequence of between 8 and 52 moves over the 64 board squares, alternating between the two players. The model predicts the next move, which is considered correct if legal, with each few-shot demonstration and query corresponding to a different board position (world).
We evaluate on 300 samples using Qwen3, Gemma-4, and Ministral-3, selecting model sizes appropriate to the difficulty of each task. See \Cref{fig:arc-illustration} for the prompt illustration and \Cref{appendix:real-world} for the detailed setup.

\Cref{fig:real-task-performance} shows few-shot performance on four benchmarks, where performance improves as the number of demonstrations increases compared with zero-shot settings, despite demonstrations and queries being drawn from different worlds. This consistency across diverse tasks suggests that the benefit of few-shot demonstrations extends beyond graph tracking.

\section{Discussion}\label{section:discussion}
In this work, we investigated how LLMs construct an internal representation of the underlying world from observations and use it for prediction. While such predictive use is difficult in the zero-shot setting \citep{lepori2026language}, we showed that few-shot demonstrations containing observations from different worlds enhance downstream predictions, both in graph tracking, ARC-AGI-1\&2, web agent tasks, and Othello. In the graph tracking setting, our representation analyses and intervention experiments further provide evidence that increasing the number of shots relocates the world representation into a subspace from which it can be more effectively read out for prediction.

\textbf{Intelligence as Skill-Acquisition.}
At a higher level, an important milestone of AI is efficient skill acquisition and adaptation  \citep{legg2007universal,chollet2019on,goldfeder2026ai} rather than merely maximizing fixed benchmarks. This work offers a mechanistic account of how LLM agents use in-context observations to perform tasks in novel environments.

\textbf{ICL in the Era of Agents.}
Previous research on ICL has focused on settings where ground truth input--output pairs are provided, without considering agentic interaction with an environment. This work instead studies a more agentic setting, which we call in-context world modeling, asking how demonstrations composed of observations, task instructions, and outcomes shape both LLM behavior and its internal mechanisms.

\textbf{LLMs Know but Cannot Say.} 
Prior work has shown that LLMs can internally represent truth even when it is not decodable \citep{kadavath2022language,burns2023discovering,azaria2023internal}, and that interventions can make such information decodable \citep{li2024inference}. In line with these findings, this work provides evidence for a gap between representation construction and its predictive use. LLMs can construct representations of world structure from observations yet fail to use them for downstream tasks, but few shot demonstrations can bridge this gap through prompting alone.

\textbf{Task (Function) Vectors in ICL.}
We observe that increasing the number of shots shifts the centroid of the hidden states of the same observed token sequence, and we refer to the direction of this shift as the \textit{Shot Vector}. While Task Vectors \citep{hendel2023context} and Function Vectors \citep{todd2024function} encode input--output mappings \citep{yang2025unifying}, our Shot Vector instead shifts the representations of observed tokens that is associated with greater predictive use of the world representation.

\textbf{Meta-In-Context-Learning.}
Meta-in-context learning \citep{coda-forno2023metaincontext} uses sequential linear regression and two-armed bandit tasks with task-specific demonstrations, showing that LLMs can infer priors over task distributions, such as linearity, from context. In contrast, our few-shot setting examines how LLMs infer how to use world representations for prediction from observations generated in different worlds under the same task.

\textbf{Limitations and Future Work.}
In this work, we define an LLM's world model in terms of its representation of the static structure of the world \citep{li2023emergent,nanda2023emergent,li2025does}. State transitions, temporal dynamics, and more general forms of causal dynamics \citep{andreas2024worldmodels} are beyond the scope of this definition.
We emphasize that performance gains from few-shot demonstrations are not solely due to increased predictive use of world representations, which \rev{is only one of several contributing factors (e.g., learning shortcuts such as candidate elimination). Further work could help disentangle these factors and extend the representation analysis to larger models.} Additionally, we studied the construction and predictive use of world representations by changing model activations through prompting. Future work should examine how RL and SFT affect these processes \citep{chu2025sft,matsutani2026rl}, their relation to generalization, and how in-context tuning \citep{min2022metaicl,chen2022meta} differs from fine-tuning. Our demonstrations used ground-truth answers, but it is also interesting to study in-context reinforcement learning (ICRL) \citep{laskin2023incontext,monea2025llms,song2026reward}, where environmental rewards are provided instead.

\subsection*{AI use statement}

In this work, we used generative AI tools (ChatGPT and Codex) to provide feedback on research methodology or experiments, implement methods, and assist with translation. We did not use generative AI tools to help develop theoretical models or conceptual frameworks, provide critical ingredients for proving mathematical claims, propose or refine hypotheses, support qualitative and thematic data analysis, or interpret results. Generating synthetic datasets, formulating mathematical claims, assisting in the writing of proofs, and cleaning or reformatting datasets were not applicable to this work.

Additionally, we used generative AI tools to create or modify scientific figures or images, suggest experimental parameters, create or edit software code, summarize or analyze existing literature, brainstorm research ideas, source or search for information, edit the paper to improve readability, identify relevant literature, and propose a title or keywords for the paper.

We reviewed all AI-assisted work. We manually reviewed and tested method implementations and code. We independently evaluated suggestions about research methods, experiments, and experimental parameters before adopting them. We checked figures and images against the underlying data and intended scientific content, and verified that translations and edits preserved the intended meaning. We take responsibility for the paper’s final content, including all AI-assisted text, claims, and artifacts.

\bibliography{iclr2027_conference}
\bibliographystyle{iclr2027_conference}

\appendix
\crefname{appendix}{appendix}{appendices}
\Crefname{appendix}{Appendix}{Appendices}

\clearpage
\section{Related Work}\label[appendix]{appendix:related-work}

\paragraph{World Models in LLMs.}

World models have been discussed in different strands of literature, such as cognitive maps \citep{behrens2018cognitive,yildirim2024task}, reinforcement learnings \citep{silver2018a,ha2018world}, abstract predictive representation\citep{lecun2022path}, and video generations \citep{bruce2024genie}.
\citet{andreas2024worldmodels,li2025does,zhang2025when} sought to define what it means for neural networks, especially language models, to learn world models.
Prior works demonstrates that neural networks encode sentiments \citep{radford2017learning}, and state and temporal information \citep{gurnee2024language} by probing internal representations, and assess their causal effects by intervening them. \citet{li2023emergent,nanda2023emergent,hua2024mothello,yuan2025revisiting,chawla2026metaothello} probed board state representations in Othello. \citet{karvonen2024emergent}, \citet{joshi2024checkersgpt}, and \citet{ivanitskiy2023linearly} did so for chess, checkers, and mazes, respectively.

\paragraph{ICL in LLMs.}
The properties of ICL \citep{brown2020language} have been studied from the perspectives of algorithmic characteristics \citep{xie2022an,oswald2023transformers}, mechanistic circuits \citep{olsson2022incontext,hendel2023context}, and language data distribution \citep{chan2022data}.
\citet{park2025iclr} showed that increasing context length induces a sudden reorganization into in-context representations, whereas \citet{lepori2026language} argued that these internal representations may not be accessible for subsequent prediction. 
\citet{mittal2025does,han2025emergence,kobayashi2024transformers,mittal2025incontext} analyzed latent-variable prediction in ICL, while \citet{jiang2025unlocking,jain2024mechanistically} investigated internal behavioral changes under supervised fine-tuning (SFT).

\section{Definition of World Models in LLMs}\label[appendix]{appendix:world-model}

\begin{wrapfigure}[17]{r}{0.60\textwidth}
  \centering
  \includegraphics[width=\linewidth]{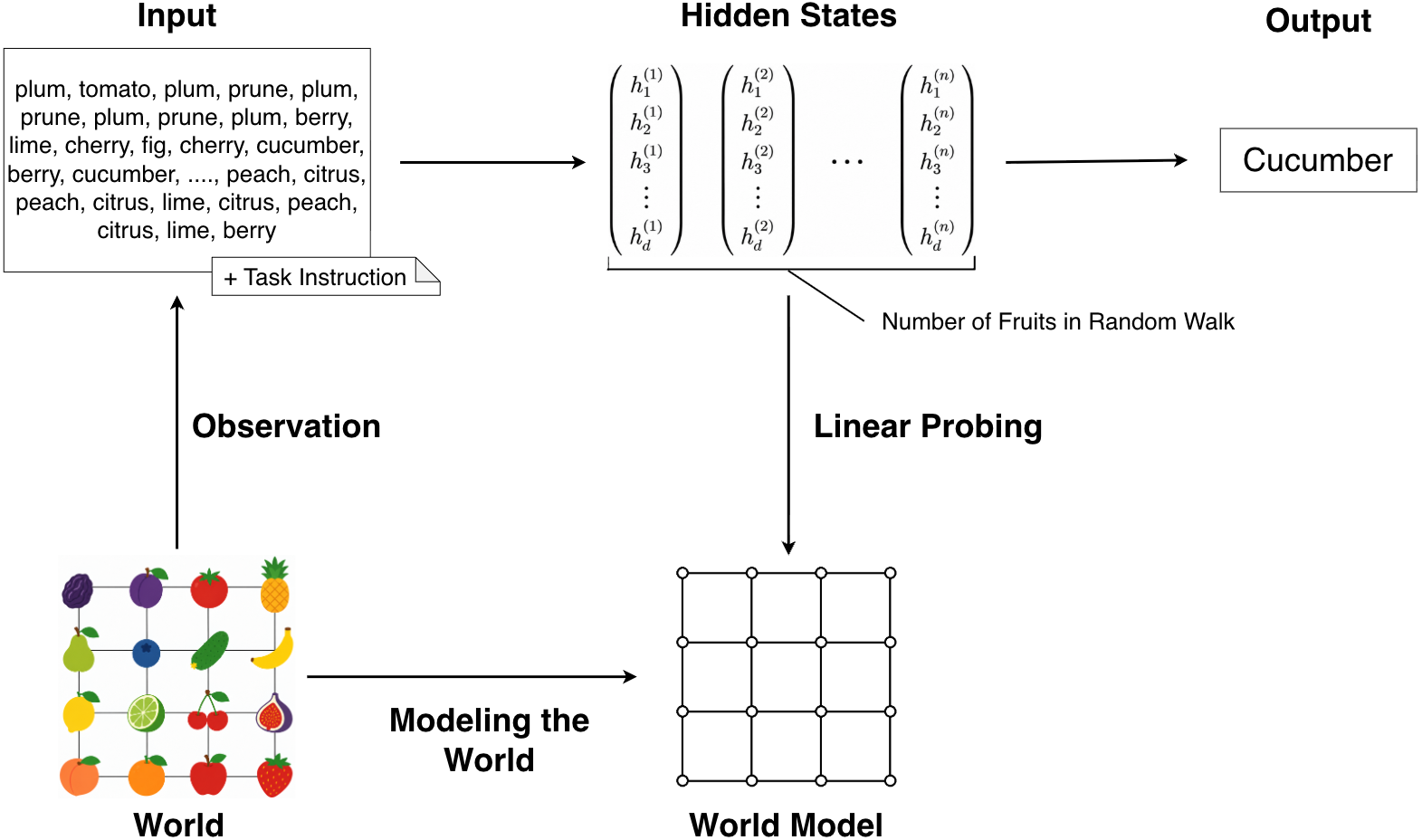}
  \caption{\textbf{Illustration of World Models in LLMs.}}
  \label{fig:wm-llm}
\end{wrapfigure}

Following \citet{andreas2024worldmodels,li2025does,zhang2025when}, we consider the world underlying the generation of the textual observations to be a static configuration.

Although world models can be defined at various levels, our setting corresponds to ''The map'' discussed in \citet{andreas2024worldmodels}. Rather than storing pairwise relations between fruits in a lookup table, the representation encodes a graph topology that captures their relative spatial relationships. Computation over this representation can then be used to solve navigation tasks on the graph. The random walks are generated using one-step transitions, whereas the task requires two-step reasoning, and thus cannot be solved by simply copying a single-hop transition. Our setting does not, however, correspond to ''The orrery,'' which models temporal evolution, or ''The simulator,'' which incorporates physical laws, as discussed in \citet{andreas2024worldmodels}.

Following \citet{li2023emergent,nanda2023emergent,yuan2025revisiting,chawla2026metaothello}, we say that an LLM possesses a world model in this sense if, as visualized in \Cref{fig:wm-llm}, it receives textual random walks generated from a world in which fruits are arranged according to a graph, and the static graph structure of that arrangement can be recovered from the LLM's internal representations using a linear probe. For representations associated with multiple subtokens or repeated occurrences, we use their average. In addition, intervening on these representations should disrupt predictions that were previously possible through them. Under this definition, the model is said to possess a world model. Note that this notion is narrower than more general world models that capture temporal evolution or causal dynamics.

Beyond our narrow notion of world models in LLMs, future work could investigate whether LLMs capture temporal dynamics and causal relationships \citep{gurnee2024language,vafa2024evaluating,rohekar2025a} through ICL, how multi-task training affects representations \citep{park2026convergent,chawla2026metaothello,kawata2025mixture}, how chain-of-thought (CoT) \citep{wei2022chain,kojima2022large,matsutani2026zipping} affects behaviors, and whether similar structures emerge in weight-tied recurrent models \citep{geiping2025scaling}.

\section{Experimental details}

\subsection{Synthetic Task}\label[appendix]{appendix:systhetic-task}

\paragraph{Graph Tracking.}

\begin{figure}[h]
  \centering
  \includegraphics[width=\linewidth]{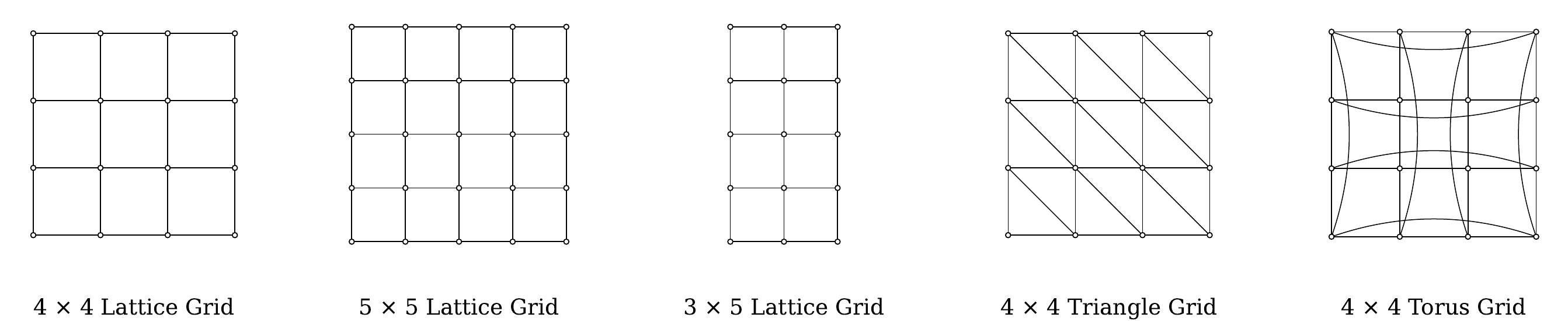}
  \caption{\textbf{Illustration of Five Graph Topologies.} }
  \label{fig:graph-topologies}
\end{figure}

Following the \textit{adaptive world modeling} setup of \citep{lepori2026language}, we use random walks generated on five graph topologies: $4 \times 4$ Lattice Grid, $5 \times 5$ Lattice Grid, $3 \times 5$ Lattice Grid, $4 \times 4$ Triangle Grid, and $4 \times 4$ Torus Grid \Cref{fig:graph-topologies}. On these graphs, we consider a two-step prediction task. Each random walk is generated by randomly sampling between 48 and 80 nodes. Note that no information about the underlying topology is provided in the prompt. The LLM must infer the topology solely from the random walk. Although \citet{lepori2026language} employed the 16-State Line and 25-State Line as latent graphs, we exclude them to rule out a shortcut in which the model simply copies the token from two positions earlier in the sequence.

For the two-step prediction task, following \citep{lepori2026language}, the model is asked to move from grid coordinate $(i,j)$ to $(i+2,j)$. Directly specifying an instruction such as ``move two steps to the right'' does not yield a unique answer under rotations or reflections of the grid. We therefore provide few-shot examples illustrating the mapping $(i,j)\to(i+2,j)$. We provide 6, 10, 3, 6, and 6 examples for the $4 \times 4$ Lattice Grid, $5 \times 5$ Lattice Grid, $3 \times 5$ Lattice Grid, $4 \times 4$ Triangle Grid, and $4 \times 4$ Torus Grid, respectively. For the downstream task, \citet{lepori2026language} considered one, two, and three-step prediction. We use two-step prediction because one-step prediction can be solved by copying an adjacent node from the random walk, whereas three-step prediction is substantially more difficult. We additionally ensure that the expected answer appears at least once in the random walk, and excluded query pairs whenever the correct target appeared exactly two positions before or after the query source in the sequence.

In this paper, under the few-shot setting, we treat each random walk together with its corresponding node pair examples for the task instruction as a single shot and repeat this format across multiple shots. Note that these shots are distinct from the few-shot examples used in the instruction for the two-step prediction task. Prompt examples are in \Cref{fig:prompt-graph-prediction}.

\begin{figure}[h]
  \centering
  \begin{minipage}{0.9\linewidth}
    \begin{PromptBox}{Prompt for Graph Tracking (Two-Step Prediction Task)}
\VerbatimPygments{\PYG}{\PYG}
\input{promptboxes/prompt1.tex}
    \end{PromptBox}
  \end{minipage}
  \caption{\textbf{Prompt for Graph Tracking Task.} A 1-shot example generated from a 4×4 Lattice Grid. The answer is "cucumber". The query corresponds to the illustration in \Cref{fig:fig1} (a).}
  \label{fig:prompt-graph-prediction}
\end{figure}

\clearpage
\subsection{Probing, Representation Analysis and Intervention Details}\label[appendix]{appendix:probing-representation-intervention}

All probing, representation analysis, and intervention experiments described below are conducted on NVIDIA GH200 GPUs.

$t$ denotes the token position index. $l$ denotes the Transformer layer index. $d$ denotes the hidden dimension. $(i,j)$ denote grid coordinates in graph tracking tasks.
Consider a specific sample, $x^{(l)}_{t} \in \mathbb{R}^d$ denotes the raw hidden state at token position $t$ in layer $l$. In this paper, we use Qwen3-4B-Base, Qwen3-8B-Base and Qwen3-14B-Base \citep{yang2025qwen3}, where $d=2560,4096,5120$ and $l=36,36,40$, respectively.

\paragraph{Probing.}
$x_t^{(l)}$ is pooled over subtokens and over occurrences of the same token within the random walk in a query. To normalize these representations, we divide them by a single scalar Root Mean Square (RMS) computed over the activations of all nodes in that layer. We obtain a node representation $h^{(l)}_{i,j} \in \mathbb{R}^d$ where $(i,j)$ denotes node indices. We train probe projection matrix $W^{(l)} \in  \mathbb{R}^{d \times r}$ where $r$ denotes the dimensionality of the latent probe space. We compute probe representation $z^{(l)}_{i,j}={W^{(l)}}^\top h^{(l)}_{i,j}\in \mathbb{R}^{r}$, and then we obtain the probability that nodes $u=(i,j)$ and $v=(i',j')$ are adjacent $p^{(l)}_{u,v}=\sigma\big( \beta^{(l)} - {\lVert z^{(l)}_{u} - z^{(l)}_{v} \rVert}^2_2 \big)$ where $\sigma(x)=(1+\exp(-x))^{-1}$ is the logistic sigmoid function and $\beta^{(l)}$ is trainable scalar offset. Note that we train $W^{(l)}$ to obtain the probe space node representations $z^{(l)}_{i,j}$, while Dirichlet Energy in \citet{park2025iclr} is computed directly on the raw node representations $h^{(l)}_{i,j}$.

To evaluate probing performance, we compute the AUROC for predicting whether pairs of graph nodes are adjacent. AUROC is computed by sweeping the threshold over predicted probabilities and integrating the resulting ROC curve using the trapezoidal rule. \citet{groger2026revisiting} noted that probing accuracies are not directly comparable across layers with different dimensionalities, but we compare different activations within the same model.

We set $r=2,2,2,2,4$ for the $4 \times 4$ Lattice Grid, $5 \times 5$ Lattice Grid, $3 \times 5$ Lattice Grid, $4 \times 4$ Triangle Grid, and $4 \times 4$ Torus Grid, respectively. The first four grids admit two-dimensional coordinate representations, motivating $r=2$. For the $4 \times 4$ Torus Grid, we use the four-dimensional embedding $z(i,j)=(\cos(2\pi i/4),\sin(2\pi i/4),\cos(2\pi j/4),\sin(2\pi j/4))$, which represents each periodic coordinate using a sine and cosine pair. The squared Euclidean distance is $2$ for adjacent nodes and at least $4$ for non-adjacent nodes, motivating $r=4$ for our probe based on distance. See \Cref{appendix:rank} for experiments with different ranks.

\paragraph{Representation Analysis.}

To obtain an orthonormal basis of this subspace, we compute the singular value decomposition
$$W^{(l)}=U^{(l)}\Sigma^{(l)}{V^{(l)}}^{\top} \in \mathbb{R}^{d \times r},$$
where $U^{(l)}\in \mathbb{R}^{d \times r}$, $\Sigma^{(l)}\in \mathbb{R}^{r \times r}$ and $V^{(l)}\in \mathbb{R}^{r \times r}$. Then, we can define world representation space $\mathcal{W}^{(l)}=\operatorname{span}(u_1^{(l)}, \ldots, u_{r}^{(l)}$), where $u_1^{(l)}, \ldots, u_{r}^{(l)}$ are columns of $U^{(l)}$. 
By using orthogonal projection matrix $\Pi_{\mathcal{W}}^{(l)} = U^{(l)}{U^{(l)}}^\top \in \mathbb{R}^{d \times d}$, we can separate in-subspace and out-of-subspace component from an activation $h^{(l)}$ as
$$h^{(l)}=\underbrace{\Pi_{\mathcal{W}}^{(l)} h^{(l)}}_{\text{in-subspace component}}+\underbrace{\left(I-\Pi_{\mathcal{W}}^{(l)}\right)h^{(l)}}_{\text{out-of-subspace component}},$$
where $I$ is the identity matrix. We can use $$0\leq\frac{\big\|\Pi^{(l)}_{\mathcal{W}}h^{(l)}\big\|_2^2}{\left\|h^{(l)}\right\|_2^2}\leq 1$$ to quantify the extent (energy) to which the activation $h^{(l)}$ lies within the world representation subspace.

To quantify the overlap (similarity) of world representation subspaces $S(\mathcal{W}^{(l)}, {\mathcal{W}'}^{(l)})$ obtained from different probes at the same layer $l$, we compute
$$
S(\mathcal{W}^{(l)}, {\mathcal{W}'}^{(l)}) =\frac{\bigl\lVert {U^{(l)}}^\top {U'}^{(l)} \bigr\rVert_{\mathrm{F}}^2}{r}=\frac{\operatorname{tr}\bigl(\Pi_{\mathcal{W}}^{(l)}{\Pi'}_{\mathcal{W'}}^{(l)}\bigr)}{r},
$$
where $\lVert \cdot \rVert_{\mathrm{F}}$ denotes the Frobenius norm. Here, $U^{(l)}$ and ${U'}^{(l)}$ are obtained from two probes trained on different data at the same layer $l$. This quantity is the mean squared cosine of the principal angles between $\mathcal{W}^{(l)}$ and ${\mathcal{W}'}^{(l)}$, ranging from $0$ for orthogonal subspaces to $1$ for identical subspaces.

Similarly, we quantify the distance of world representation subspace $D(\mathcal{W}^{(l)}, {\mathcal{W}'}^{(l)})$ obtained from different probes at the same layer $l$ by computing
$$D(\mathcal{W}^{(l)}, {\mathcal{W}'}^{(l)}) = \sqrt{1-\frac{\bigl\lVert {U^{(l)}}^\top {U'}^{(l)}\bigr\rVert_{\mathrm{F}}^2}{r}} = \frac{\bigl\lVert \Pi_{\mathcal{W}}^{(l)}-{\Pi'}_{\mathcal{W}'}^{(l)}\bigr\rVert_{\mathrm{F}}}{\sqrt{2r}}.$$

Additionally, let $\theta^{(l)}$ be the angle between the activation vector $h^{(l)}$ and the subspace $\mathcal{W}^{(l)}$,
$$\theta^{(l)} = \arccos\Big( \frac{ \big\|\Pi_{\mathcal{W}}^{(l)} h^{(l)}\big\|_2 }{ \big\|h^{(l)}\big\|_2 } \Big) \qquad \text{with} \quad 0 \leq \theta^{(l)} \leq \frac{\pi}{2}.$$

\paragraph{Intervention.}
A single low-rank probe may leave additional linearly decodable adjacency information. Following \citep{elazar2021amnesic}, we repeat the probing and removal procedure. 

At each layer and for each sample, the pooled representations of the nodes observed in the final query walk are centered by subtracting their mean. By writing the mean representation as $\bar{h}^{(l)}$, the graph intervention removes its component in $\mathcal{W}^{(l)}$ as
$$h_{i,j}^{(l)}
  \leftarrow
  h_{i,j}^{(l)}-
  \Pi_{\mathcal W}^{(l)}(h_{i,j}^{(l)}-\bar{h}^{(l)}).$$
Operationally, let $s^{(l)}$ denote the scalar RMS used to normalize the pooled representations. The vector $-s^{(l)}\Pi_{\mathcal W}^{(l)}(h_{i,j}^{(l)}-\bar{h}^{(l)})$ is added to $x_t^{(l)}$ at every selected subtoken position $t$ in every occurrence of node $(i,j)$ within the final query walk.

To compare with this low-rank graph intervention, we intervene random orthonormal directions to the world representation subspace in the same way. At each layer, we sample as many random orthonormal directions as the current dimension of $\mathcal{W}^{(l)}$. So, this random subspace has the same dimension as $\mathcal{W}^{(l)}$. We project the current centered $h_{i,j}^{(l)}-\bar h^{(l)}$ onto this orthogonal random subspace. This projected vector is rescaled separately for every node $(i,j)$ and every layer $l$ so that its norm equals $\big\lVert\Pi_{\mathcal W}^{(l)}(h_{i,j}^{(l)}-\bar{h}^{(l)})\big\rVert_2.$
We then subtract the rescaled random vector from $x^{(l)}_t$ at the same token positions used by the graph intervention.

\subsection{Practical Tasks}\label[appendix]{appendix:real-world}

\paragraph{ARC-AGI-1\&2.}
We use the public evaluation set of ARC-AGI-1 \citep{chollet2019on}. The set contains 400 tasks and 419 test pairs, where 381 tasks have one test pair and 19 have two. For ARC-AGI-2 \citep{chollet2026arcagi2}, we use evaluation set with 120 tasks and 167 test pairs, with 75 tasks containing one test pair, 43 containing two, and 2 containing three. Each test pair was evaluated with an independent prompt, while all native input–output training pairs belonging to its target task were retained. A single shot consists  of the task instruction and a set of input–output pairs provided as observations in a serialized format. We construct a few-shot prompt by concatenating these shots, using demonstrations from the training set and the query from the public evaluation set. Prompt examples are in \Cref{fig:prompt-arc}. 

Prior work has approached ARC-AGI using leave-one-out test-time training (TTT) within a task \citep{akyurek2025the}, a combination of data augmentation and test-time search \citep{franzen2025product}, Vision Transformers (ViTs) that formulate ARC as an image-to-image translation problem \citep{hu2025arc}, and weight-tied recurrent models \citep{shu2026loopvit,liu2026tracevit}.

\begin{figure}[h]
  \centering
  \begin{minipage}{0.9\linewidth}
    \begin{PromptBox}{Prompt for ARC-AGI-1\&2}
\VerbatimPygments{\PYG}{\PYG}
\input{promptboxes/prompt2.tex}
    \end{PromptBox}
  \end{minipage}
  \caption{\textbf{Prompt for ARC-AGI-1\&2.} A 1-shot example. The part marked \textbf{[SKIP]} is omitted for brevity.}
  \label{fig:prompt-arc}
\end{figure}

\paragraph{Web Agent Task.}

We use the cross-website subset of the Mind2Web dataset \citep{deng2023mind2web}, which contains 177 tasks and 1373 action steps. Each query prompt contained the user’s goal, gold previous action history, the HTML of the current webpage, and the top-10 elements from the ranking. Few-shot demonstrations were drawn from the disjoint train set, whereas queries were drawn from the cross-website subset of the test set, since we need to ensure that the world (website) of the query is different from that of the demonstrations. Prompt examples are in \Cref{fig:prompt-web}.

For ARC-AGI-1\&2, we use top-p sampling with temperature 0.7 and top-p 0.95. For Othello and Mind2Web (cross-website subset), we use greedy decoding with temperature 0 and top-p 1. For all tasks, we generate the answer immediately after prefill in the specified format with a sufficient response length.

\begin{figure}[h]
  \centering
  \begin{minipage}{0.9\linewidth}
    \begin{PromptBox}{Prompt for Web Agent Task}
\VerbatimPygments{\PYG}{\PYG}
\input{promptboxes/prompt3.tex}
    \end{PromptBox}
  \end{minipage}
  \caption{\textbf{Prompt for Web Agent Task.} A 1-shot example of web agent tasks from Mind2Web dataset. The answer is "F" and "CLICK".}
  \label{fig:prompt-web}
\end{figure}

\paragraph{Othello.}

\begin{figure}[h]
  \centering
  \begin{minipage}{0.9\linewidth}
    \begin{PromptBox}{Prompt for Othello}
\VerbatimPygments{\PYG}{\PYG}
\input{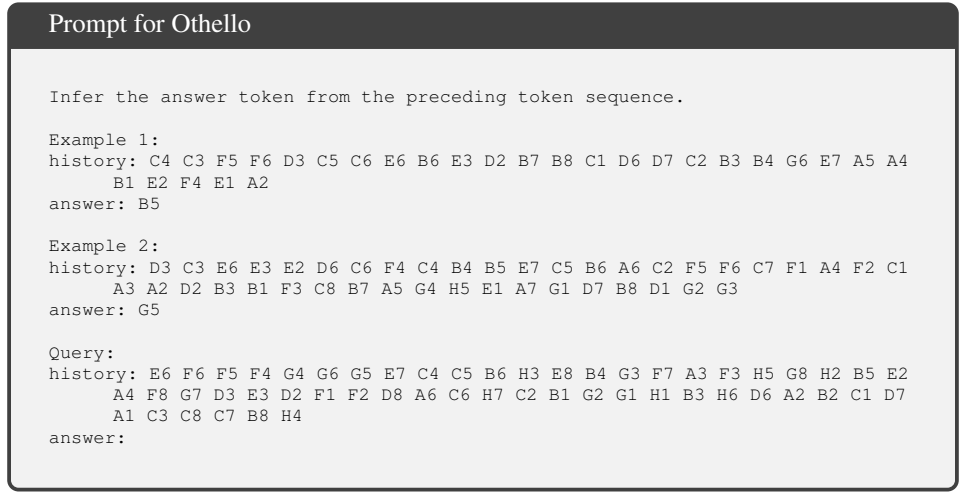}
    \end{PromptBox}
  \end{minipage}
  \caption{\textbf{Prompt for Othello.} A 2-shot example of classical 8×8 Othello. The answer is "A5".}
  \label{fig:prompt-othello}
\end{figure}

Following \citet{li2023emergent,nanda2023emergent}, in the Othello task, the model is given a sequence of 8--52 moves, where each move is represented by a board coordinate such as ``A5'' or ``E3'' and corresponds to a move by either $\mathrm{MINE}$ or $\mathrm{YOURS}$. The model is then asked to predict the next legal move. Although the moves are provided as board coordinates, the model is not explicitly told that they are Othello moves. It must therefore infer the current board state from the history of observed moves in order to predict a legal next move.
Prompt examples are in \Cref{fig:prompt-othello}

\clearpage

\section{Detailed Results}

\subsection{Few-Shot Performance}\label[appendix]{appendix:few-shot-performance}

In addition to \Cref{fig:few-shot-performance}, we show that mixing graph topologies different from that of the query in the demonstrations also improves performance. To evaluate this setting, we randomly sample the topology of each demonstration from four graphs other than the graph of the query.

\begin{figure}[h]
  \centering
  \includegraphics[width=\linewidth]{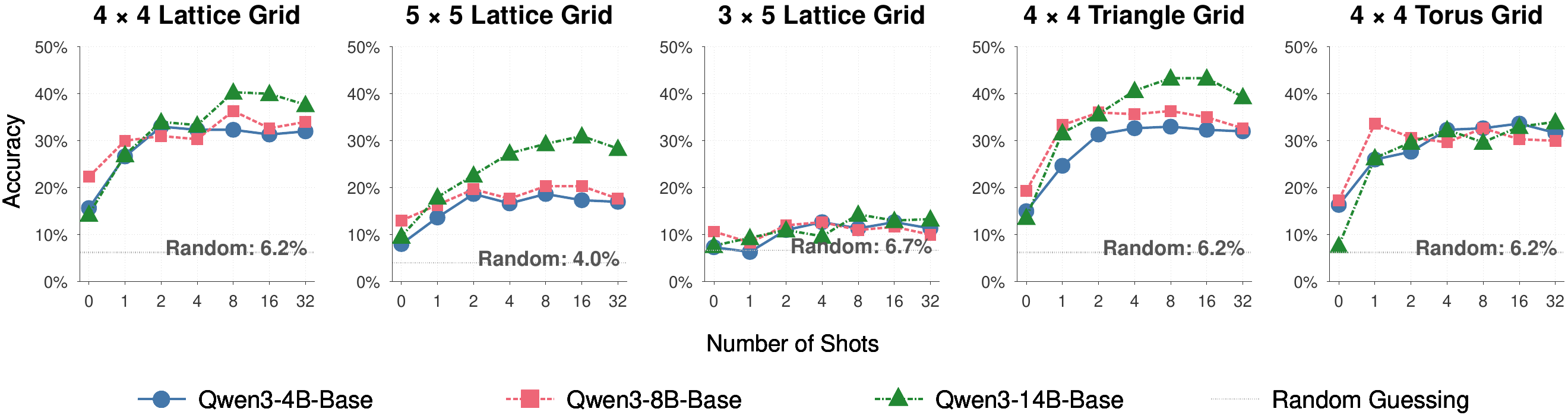}
  \caption{\textbf{Few-Shot Performance where Demonstrations are from Different Graph Topologies.} Accuracy of three Qwen3 models on the two-step-right prediction task across five graph topologies. The demonstrations are randomly sampled from four topologies other than the query topology, with each topology represented equally when the number of shots is 4, 8, 16, or 32. We report performance separately for each graph topology in the query. Each point reports accuracy on 300 test samples with 0, 1, 2, 4, 8, 16, or 32 few-shot demonstrations. Dotted gray lines indicate random guessing over the candidate nodes for each topology.}
  \label{fig:few-shot-performance-appendix}
\end{figure}

\Cref{fig:few-shot-performance-appendix} shows few-shot performance under the settings described above. The graph in each title indicates the query graph, while the demonstrations are sampled from the other graphs. Even under this harder setting, where the demonstrations come from different worlds, few-shot demonstrations improve task performance.

\rev{\subsection{Ablation of Few-Shot Prompting}\label[appendix]{appendix:few-shot-ablation}}

\begin{figure}[h]
  \centering
  \includegraphics[width=\linewidth]{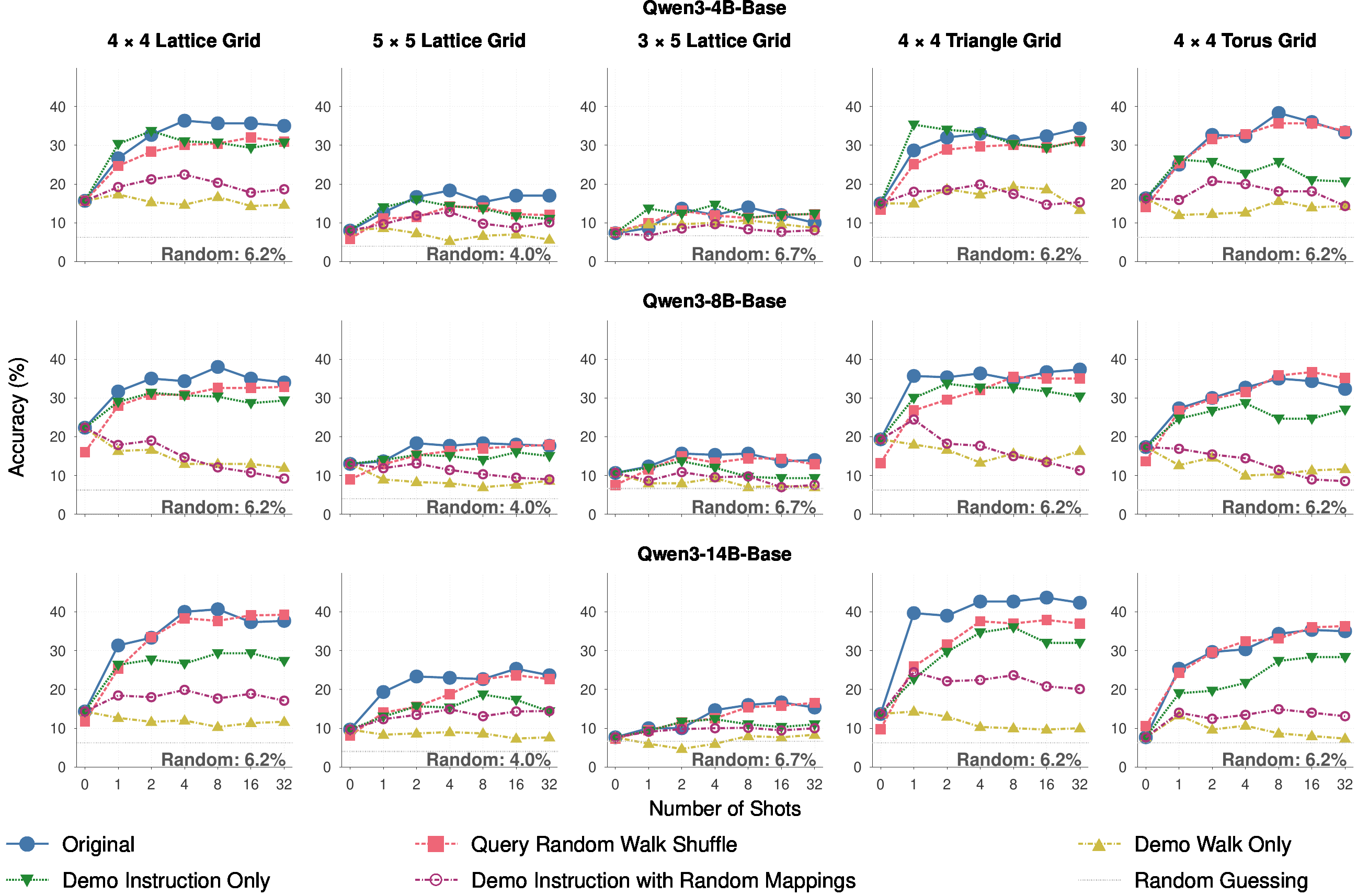}
  \caption{\rev{\textbf{Few-Shot Performance with Different Promptings.} Accuracy of three Qwen3 models on the two-step-right prediction task across five graph topologies. Each point reports accuracy on 300 test samples with 0, 1, 2, 4, 8, 16, or 32 few-shot demonstrations. Dotted gray lines indicate random guessing over the candidate nodes for each topology. While the Original setting (blue) is identical to that in \Cref{fig:few-shot-performance}, Query Random Walk Shuffle (red) shuffles the random walks in the query. In addition, Demo Walk Only (yellow) uses only random walks in the demonstrations (omitting instructions), while Demo Instruction Only (green) includes only instructions (omitting random walks). Demo Instruction with Random Mappings keeps the demonstrations' random walks intact, but replaces the target answers in the instruction mapping pairs with nodes sampled uniformly at random.}}
  \label{fig:few-shot-performance-ablation}
\end{figure}

\rev{We report additional baselines corresponding to \Cref{fig:few-shot-performance} in \Cref{fig:few-shot-performance-ablation}. The additional baselines are as follows:}
\rev{
\begin{itemize}
    \item Original: Identical to the setup in Figure 1.
    \item Query Random Walk Shuffle: Shuffles the random walks in the query.
    \item Demo Walk Only: Includes only random walks in demonstrations (omits instructions).
    \item Demo Instruction Only: Includes only instructions in demonstrations (omits random walks).
    \item Demo Instruction with Random Mappings: Retains demonstration random walks, but replaces target answers with uniformly sampled random nodes.
\end{itemize}
}
\rev{First, while below Original, Query Random Walk Shuffle achieves nearly comparable accuracy. This suggests that even with shuffled walks, models can partially succeed through elimination using demonstration instructions. Since performance may conflate internal graph reasoning with heuristic elimination, we evaluate stronger closed models in \Cref{fig:few-shot-performance-closed-models} to disentangle the two effects (detailed in \Cref{appendix:few-shot-performance-closed-models}). Next, performance drops substantially from Original across Demo Instruction Only, Demo Instruction with Random Mappings, and Demo Walk Only, in that order. This indicates that having both random walks and matching task instructions in demonstrations is crucial for graph tracking task.}

\rev{\subsection{Few-Shot Performance on Closed Models}\label[appendix]{appendix:few-shot-performance-closed-models}}

\rev{We report the evaluation results on three closed models: GPT-6.1-Terra, Gemini-3.8-Flash, and Qwen3.7-Plus, in \Cref{fig:few-shot-performance-closed-models}. To prevent CoT reasoning, we added the instruction "Complete the pattern. Reply with exactly one item from these allowed answers and no explanation: \{candidate fruit words separated by commas\}" to their system prompts. The reasoning effort was configured to "low" for Gemini-3.8-Flash and "none" for the other models, with the maximum response length capped at 16 tokens.}

\rev{As shown in \Cref{fig:few-shot-performance-closed-models}, stronger models benefit substantially from few-shot demonstrations in graph tracking; notably, GPT-6.1-Terra exceeds 80\% accuracy at 32 shots. Although truncating reasoning degrades performance on certain models and topologies, this trend remains consistent. Furthermore, while the gap between Original and Query Random Walk Shuffle was marginal for Qwen3 models (\Cref{fig:few-shot-performance-ablation}), closed models exhibit a clear separation. This demonstrates a few-shot benefit in graph tracking that cannot be explained by elimination alone.}

\begin{figure}[h]
  \centering
  \includegraphics[width=\linewidth]{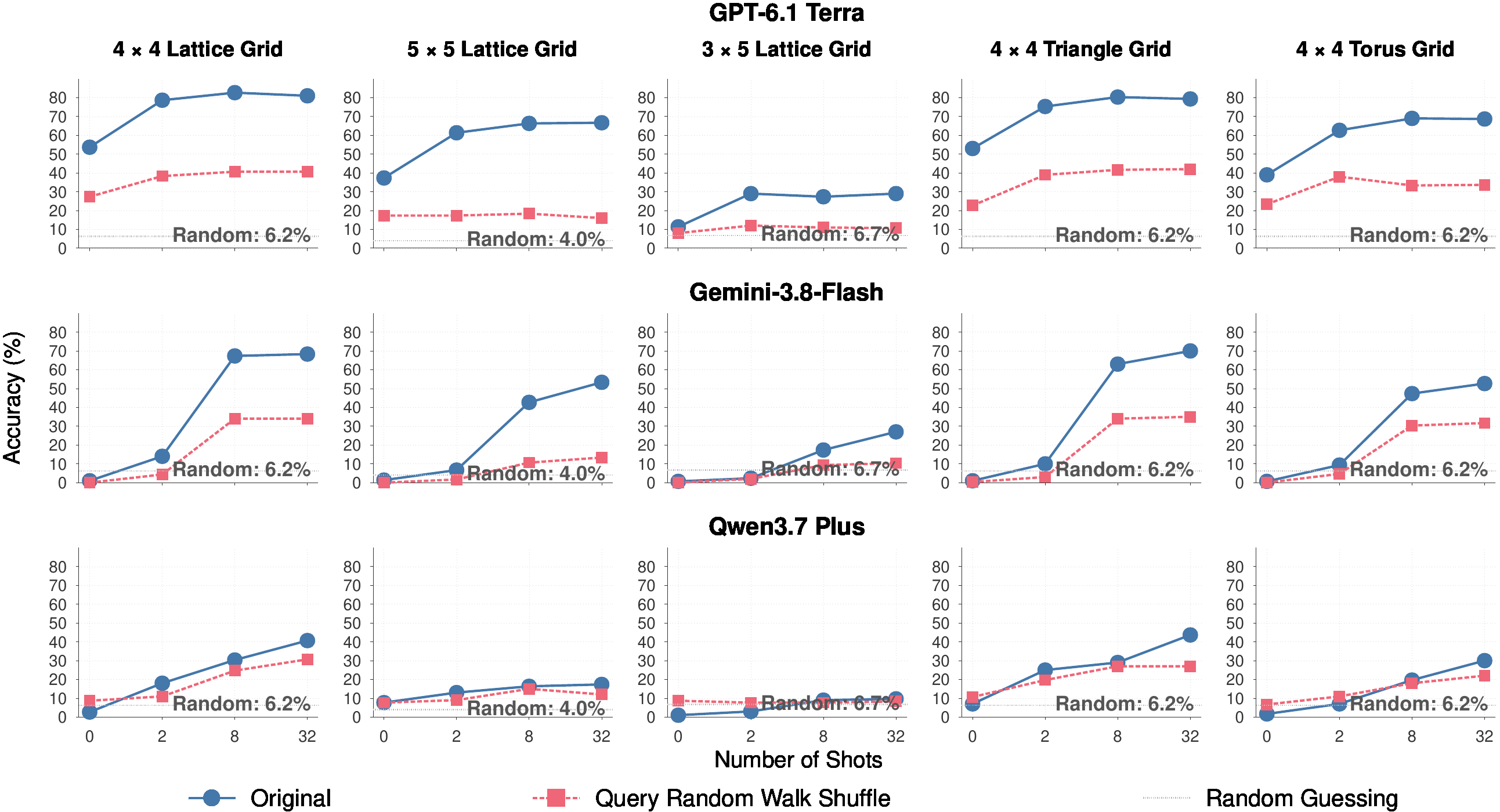}
  \caption{\rev{\textbf{Few-Shot Performance on Closed Models.} Accuracy of three closed models on the two-step-right prediction task across five graph topologies. Each point reports accuracy on 300 test samples with 0, 2, 8, or 32 few-shot demonstrations. Dotted gray lines indicate random guessing over the candidate nodes for each topology. While the Original setting (blue) is identical to that in \Cref{fig:few-shot-performance}, the Query Random Walk Shuffle setting (red) shuffles the random walks in the query.}}
  \label{fig:few-shot-performance-closed-models}
\end{figure}

\subsection{\rev{Removing the Effects of Candidate Elimination}}\label[appendix]{appendix:candidate-selection}

\rev{
In \Cref{appendix:few-shot-ablation,appendix:few-shot-performance-closed-models}, we examined whether models can perform candidate elimination when the query random walk is shuffled. Here, we report Candidate Selection Rate, the probability that the model predicts one of the remaining candidates after elimination, and Conditional Accuracy, the accuracy among predictions that fall within these remaining candidates.}

\begin{figure}[h]
  \centering
  \includegraphics[width=\linewidth]{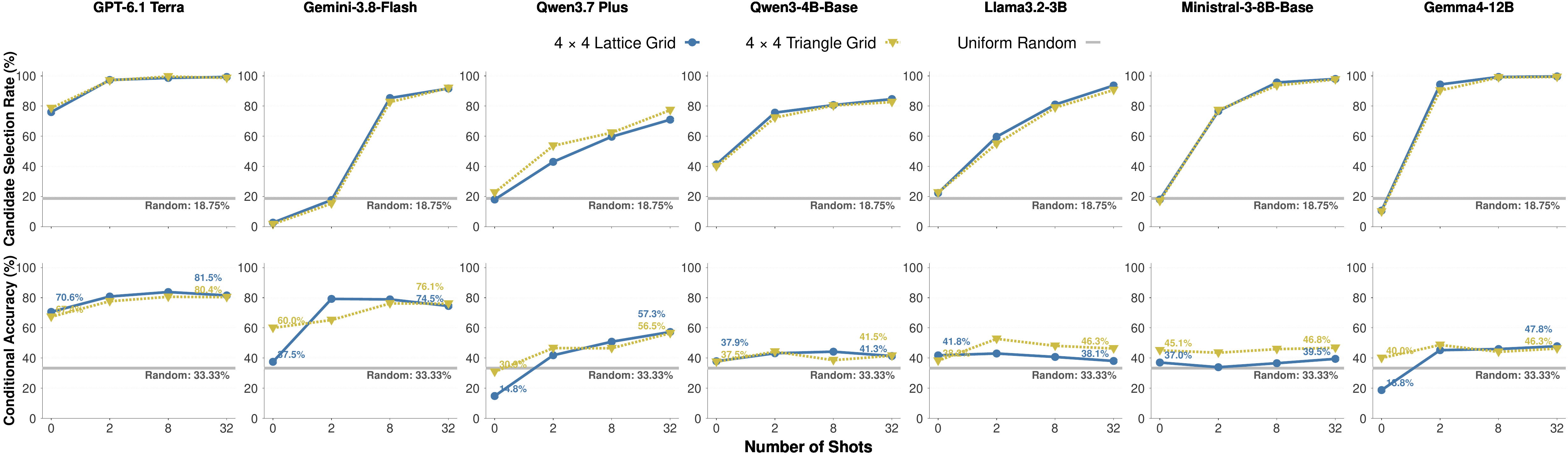}
  \caption{\rev{\textbf{Candidate Selection Rate and Conditional Accuracy.}
Performance of seven models on the two-step-right prediction task in the $4\times4$ Lattice Grid (blue) and Triangle Grid (yellow).
Each condition comprises 300 test samples with 0, 2, 8, or 32 few-shot demonstrations.
For each sample, the six input--output pairs and the query input exclude 13 of the 16 nodes, leaving three candidates, including the correct answer.
\textbf{(Top)} Candidate Selection Rate: the fraction of predictions falling within the three remaining candidates.
\textbf{(Bottom)} Conditional Accuracy: the fraction of correct predictions among samples in which the model predicts one of these three candidates, computed as the number of correct predictions divided by the number of predictions within the candidate set.
Both metrics are computed from the original model predictions without restricting the prediction space to the three remaining candidates.
Gray lines indicate random baselines of $3/16=18.75\%$ and $1/3=33.33\%$, respectively.}
}
  \label{fig:candidate-selection}
\end{figure}

\rev{
As shown in \Cref{fig:candidate-selection}, Candidate Selection Rate increases with the number of shots. Conditional Accuracy is also higher with few-shot demonstrations than in the zero-shot setting and exceeds the random guessing baseline (33.33\%), which indicates that the gains cannot be explained by candidate elimination alone. While the improvement is modest for Llama-3.2-3B, larger models, particularly closed models (GPT-6.1-Terra, Gemini-3.8-Flash, and Qwen3.7-Plus), show substantial increases in Conditional Accuracy as the number of shots increases.
}

\rev{
This suggests that improvements in the two-step prediction task on graphs are driven in large part by candidate elimination, but cannot be solely explained by candidate elimination.
}

\subsection{Probing and Representation Analysis}
\begin{figure}[h]
  \centering
  \includegraphics[width=\linewidth]{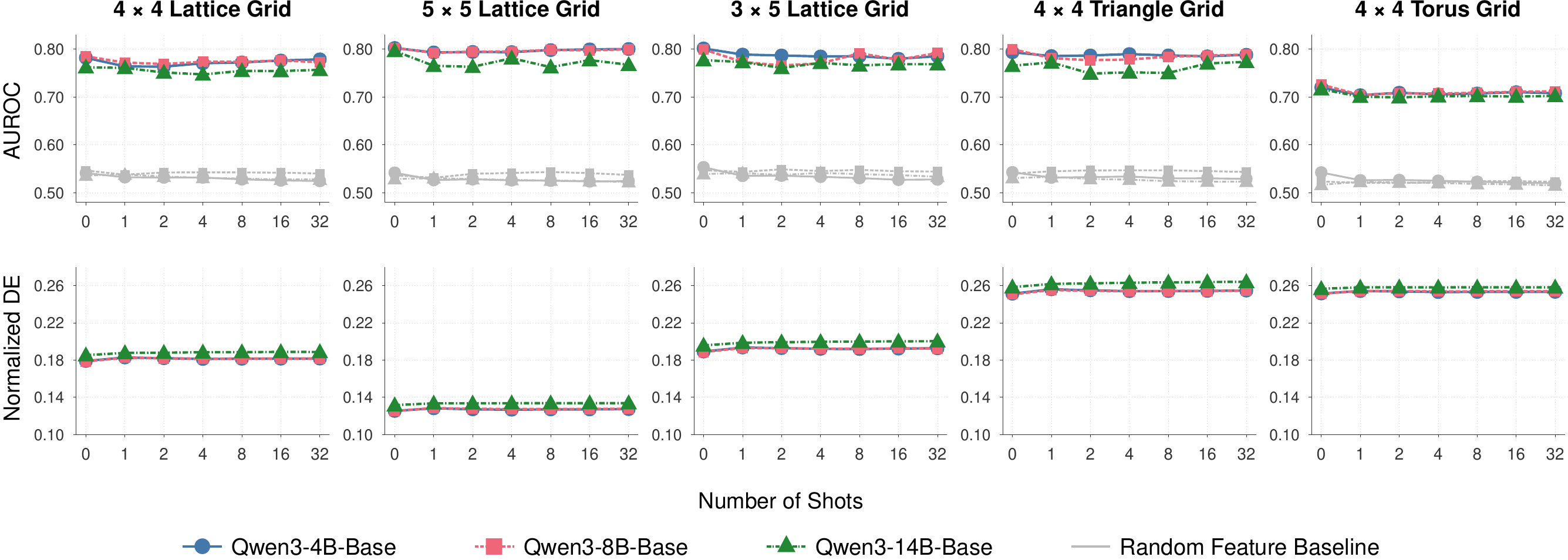}
  \caption{\textbf{Probe Performance and Normalized Dirichlet Energy (DE).} Probe AUROC (Top) and normalized DE of hidden states (Bottom) across shots and five graph topologies for layer 20 of three Qwen3 models. Gray lines show random feature probe baselines. Each condition is evaluated on 300 test samples.}
  \label{fig:auroc-detail}
\end{figure}

For Qwen3-4B-Base, Qwen3-8B-Base, and Qwen3-14B-Base, we report detailed probing results in \Cref{fig:auroc-detail}, cross-probe performance in \Cref{fig:cross-probe}, the overlap of the world representation subspaces in \Cref{fig:subspace-overlap}, PCA of the graph representations in \Cref{fig:pca-3d-example}, and PCA of the graph representation centroids over all samples in the test set in \Cref{fig:pca-example}.
  
Additionally, to examine how the representation structure of the same query, specifically the relative positions of nodes, changes with the number of shots, we computed the similarity of node positions between the 0-shot setting and the 1, 2, 4, 8, 16, and 32-shot settings.
For each comparison between 0-shot and a given number of shots, we compute the Euclidean distances between all pairs of common nodes and normalize each distance vector to unit length. We then compute the squared Euclidean difference between the normalized vectors, average it across comparisons, and take the square root. Smaller values indicate greater preservation of the relative distance structure among nodes. Note that this metric captures the relative positions of nodes and is invariant to translation and rotation.

\begin{figure}[h]
  \centering
  \includegraphics[width=\linewidth]{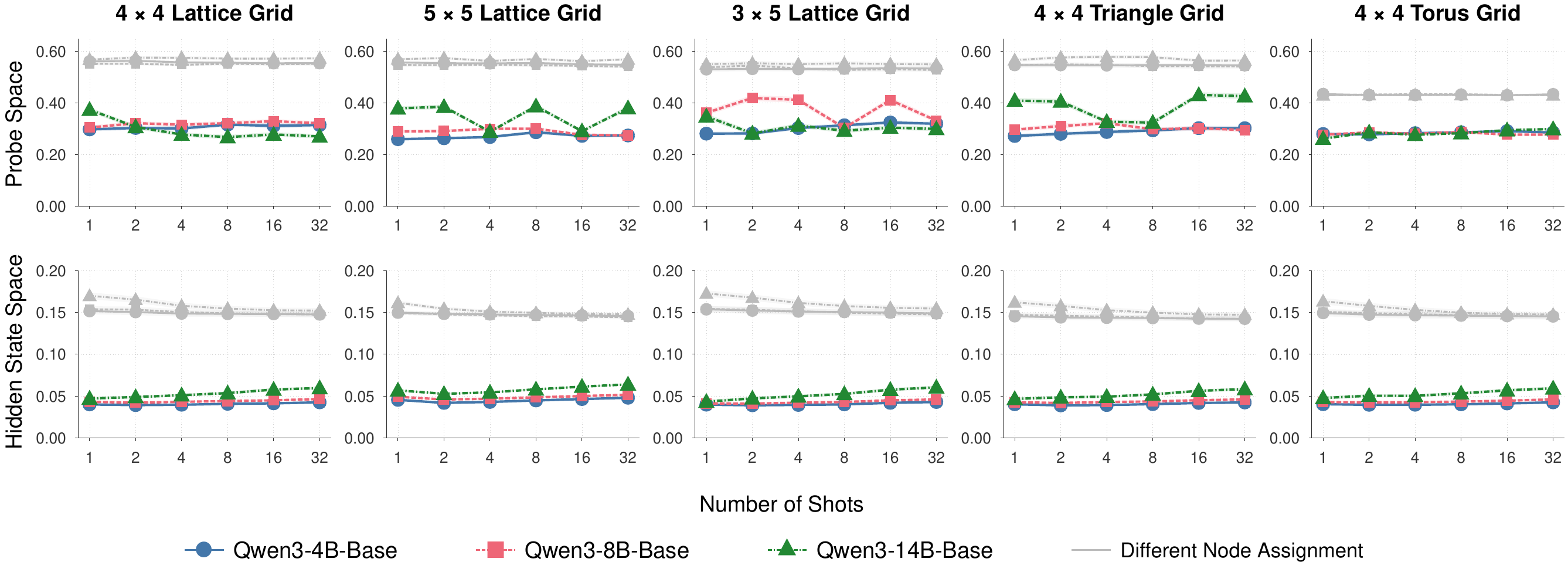}
  \caption{\textbf{Node Distance Geometry Relative to Zero-Shot Setting.} Node distance geometry relative to zero-shot setting using probe projected representations (Top) and full hidden states (Bottom) across shots and five graph topologies for layer 20 of three Qwen3 models. Colored lines compare the same query across shots, while gray lines compare different query families with the same topology but different node vocabulary assignments. Lower values indicate more similar relative node geometry. Each condition uses 300 test samples.}
  \label{fig:relative-node-position}
\end{figure}

\begin{figure}[h]
  \centering
  \includegraphics[width=0.88\linewidth]{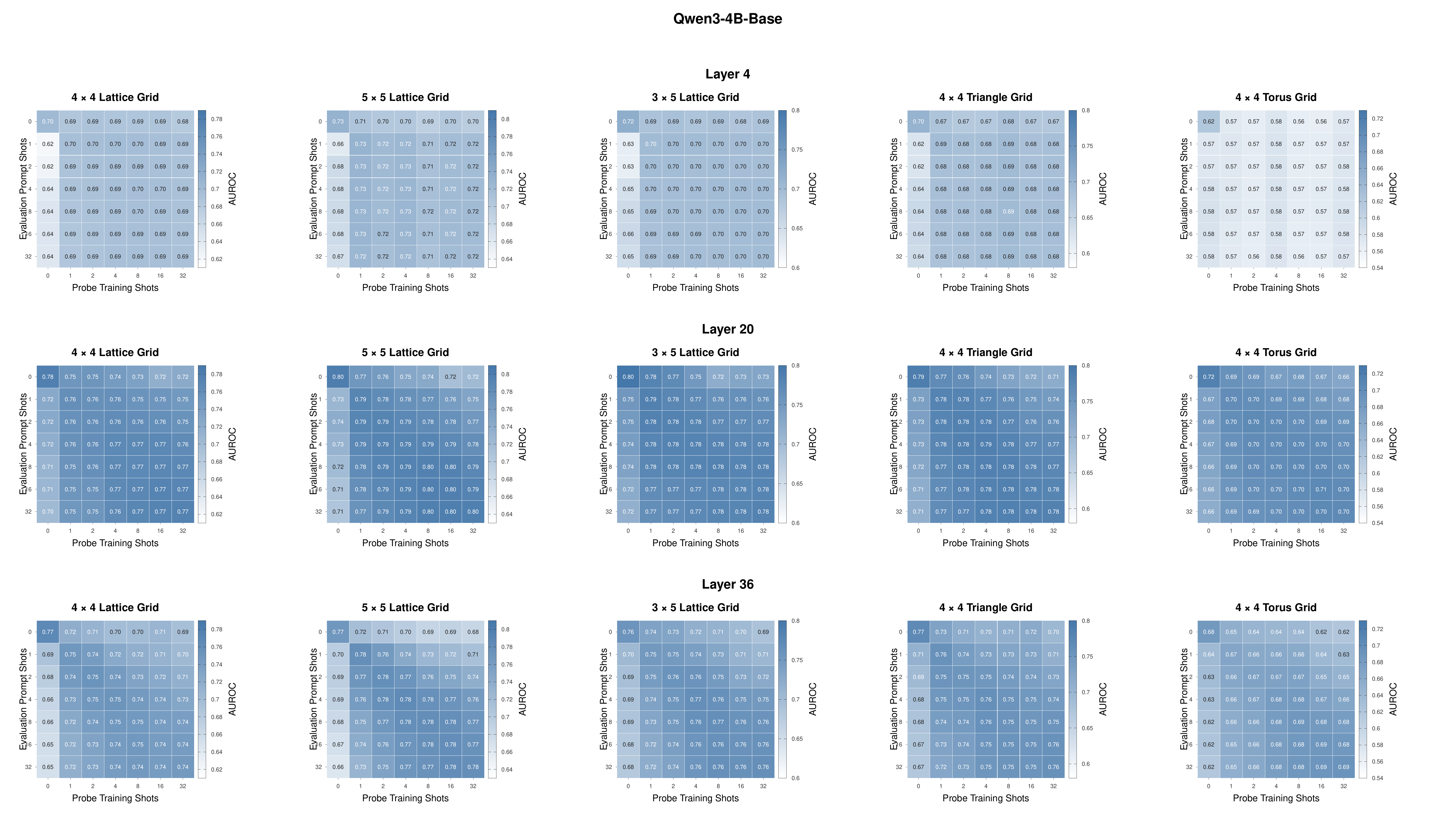}
  \vspace{0.5em}
\includegraphics[width=0.88\linewidth]{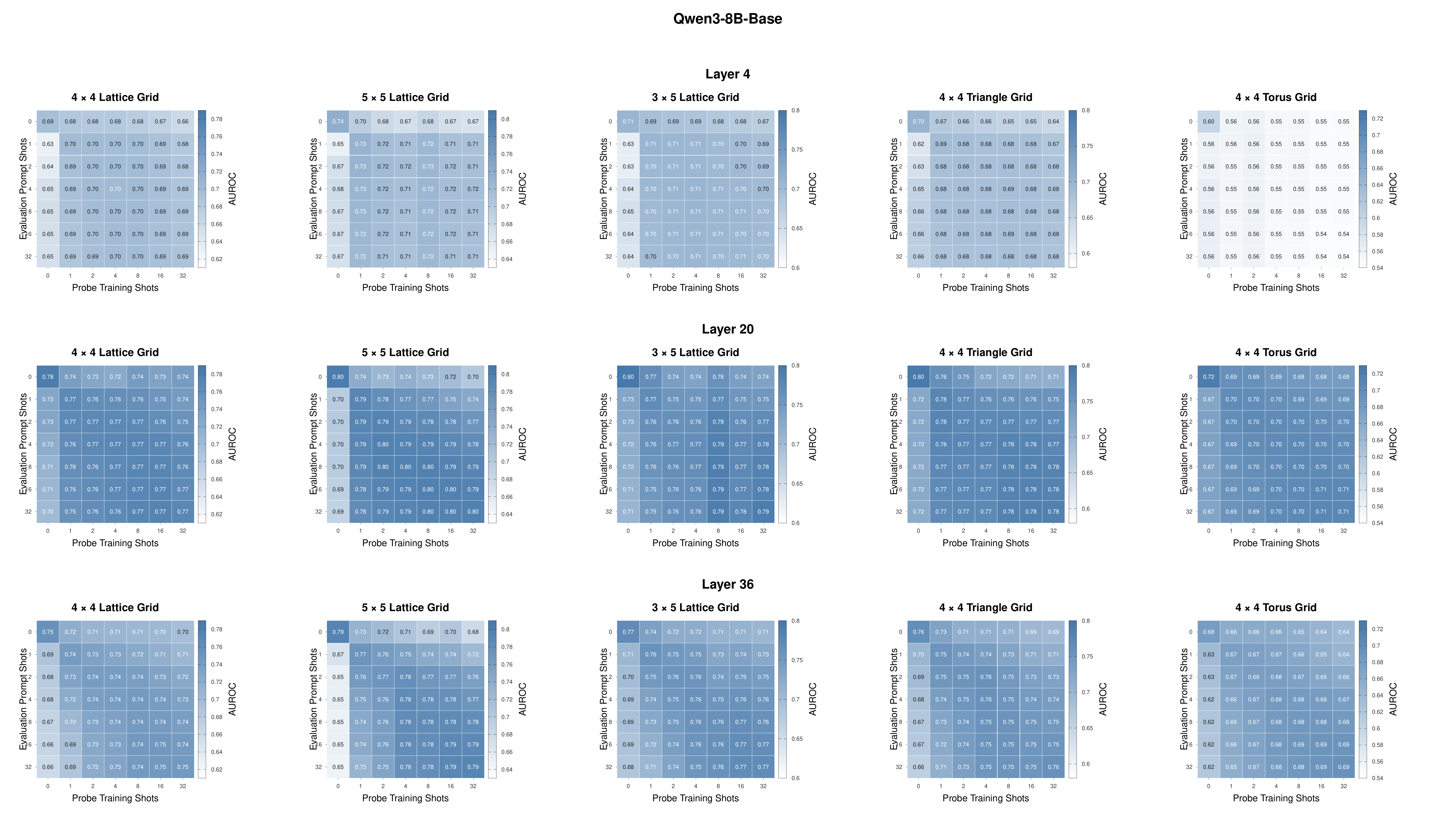}
  \vspace{0.5em}
\includegraphics[width=0.88\linewidth]{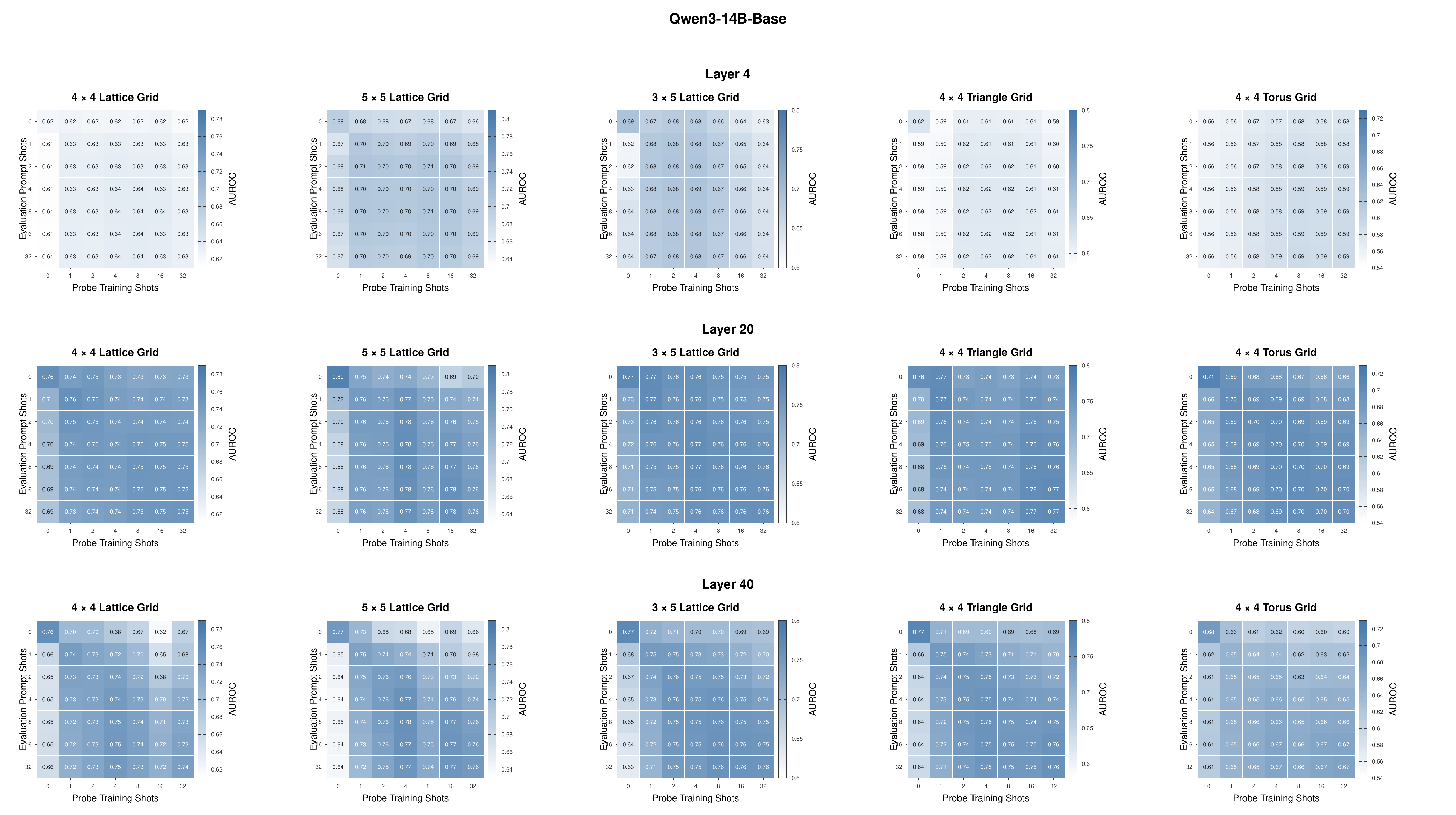}
  \caption{\textbf{Cross Probe Performance.} AUROC of probes trained at one shot count and evaluated at another. Results are from three Qwen3 models on five graph topologies.}
  \label{fig:cross-probe}
\end{figure}

\begin{figure}[h]
  \centering
  \includegraphics[width=0.84\linewidth]{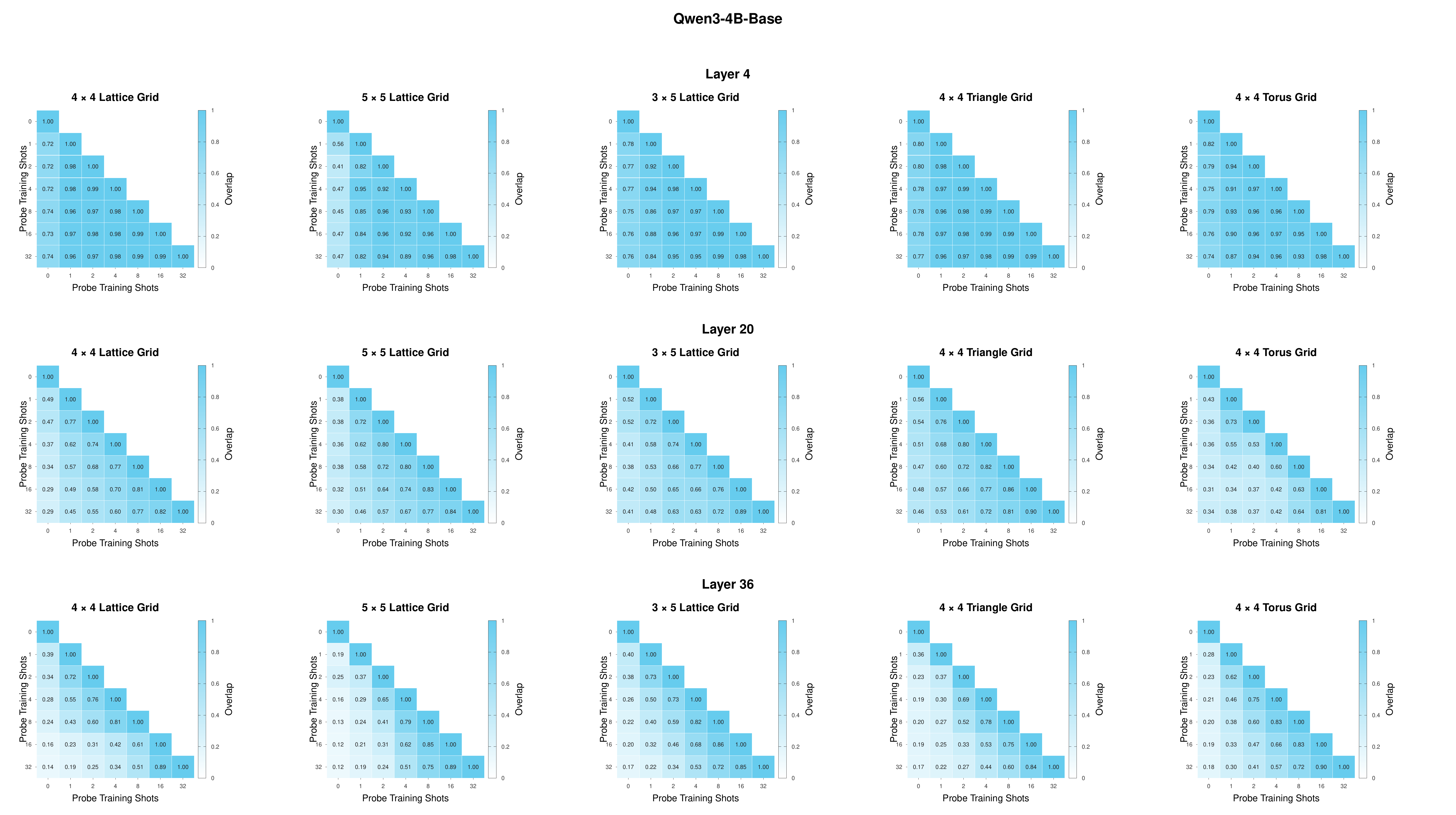}
  \vspace{0.5em}
\includegraphics[width=0.84\linewidth]{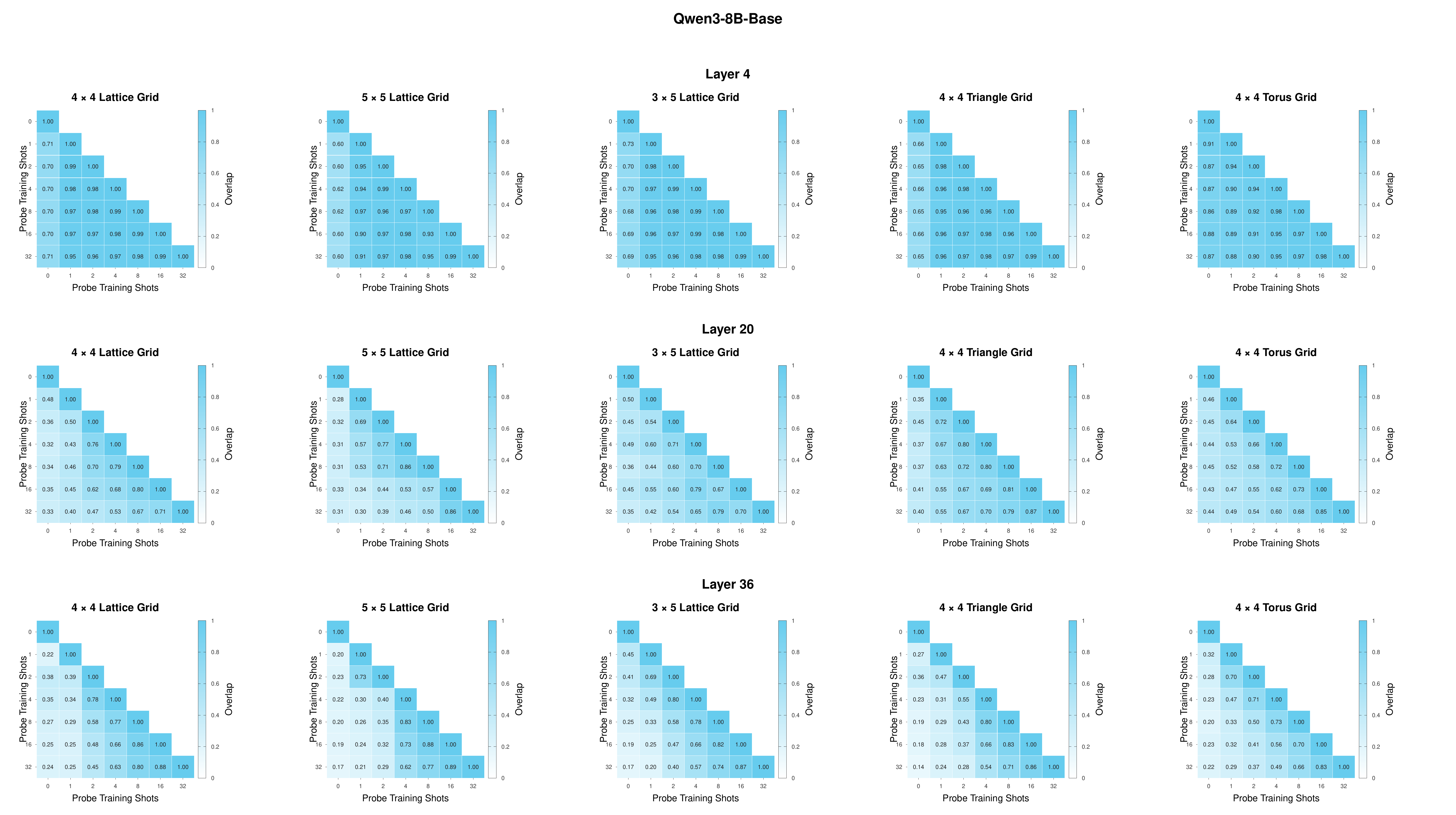}
  \vspace{0.5em}
\includegraphics[width=0.84\linewidth]{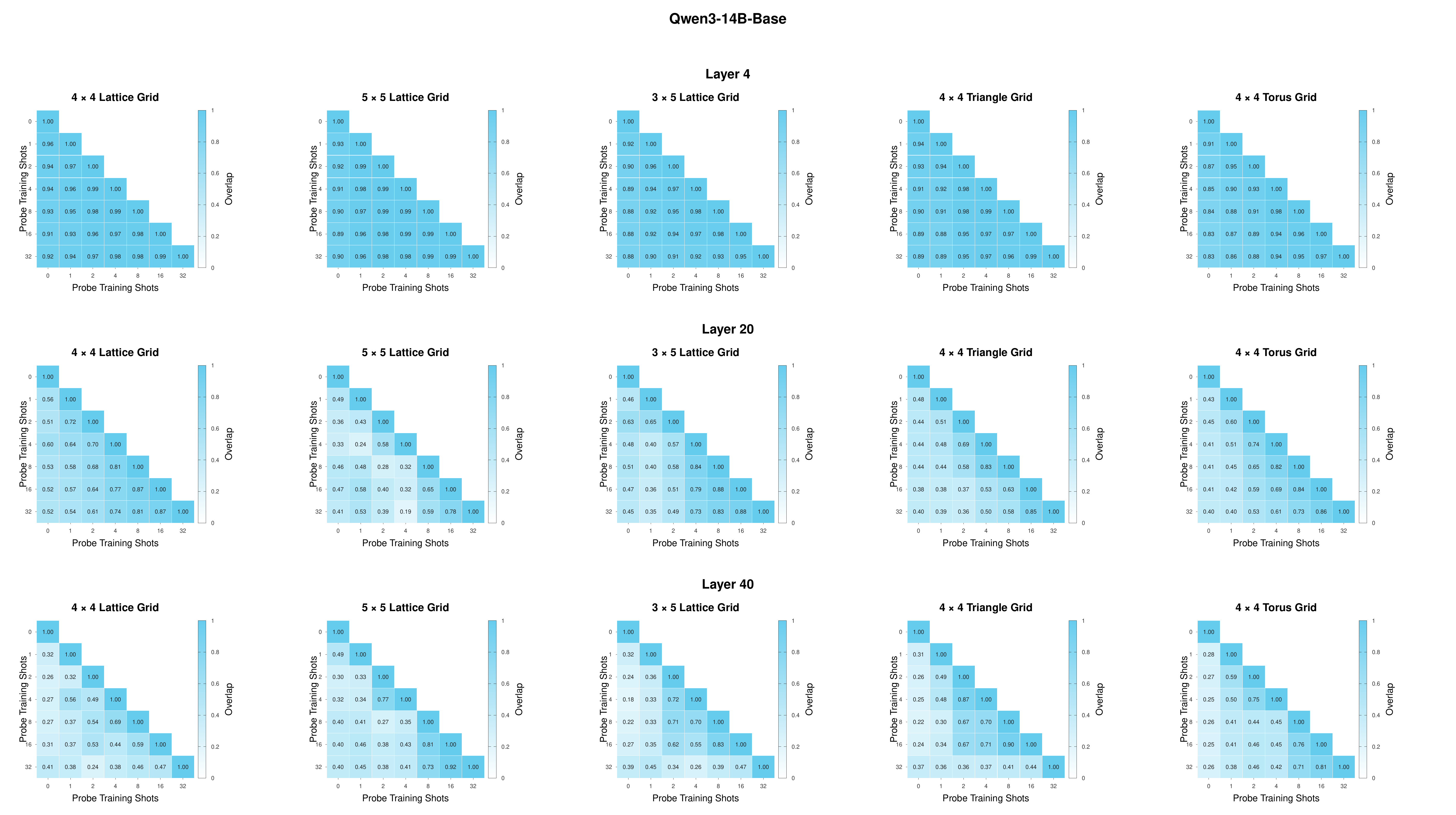}
  \caption{\textbf{Representation Subspace Overlap Results.} Overlap between probe subspaces across shot counts. Results are from three Qwen3 models on five graph topologies.}
  \label{fig:subspace-overlap}
\end{figure}

\begin{figure}[h]
  \centering
  \includegraphics[width=0.90\linewidth]{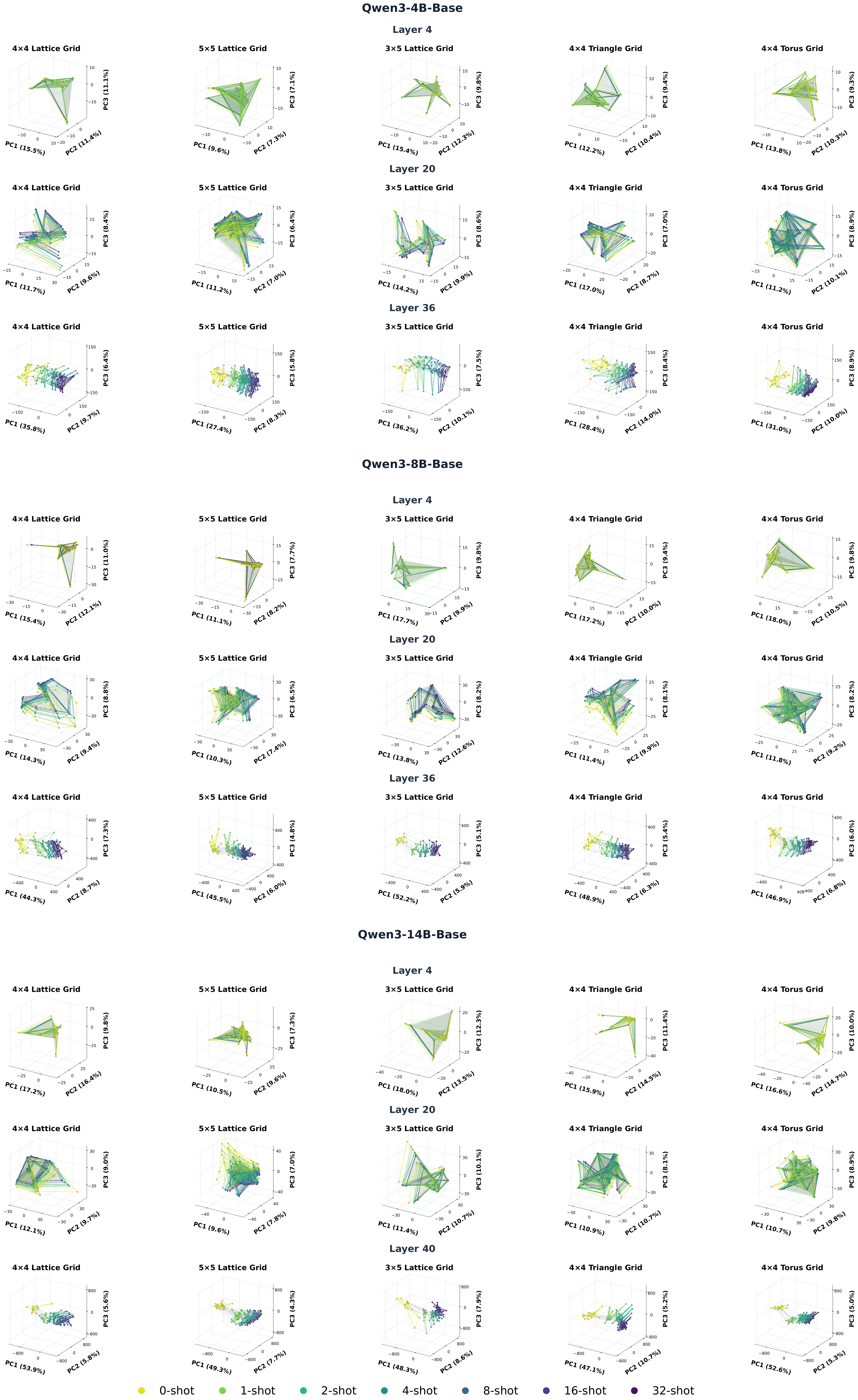}
  \caption{\textbf{PCA of Shift in Hidden States of a Random Walk Across Different Number of Shots.} In each panel, PCA is fit jointly to all activations shown. Results are from three Qwen3 models on five graph topologies.}
  \label{fig:pca-3d-example}
\end{figure}

\begin{figure}[h]
  \centering
  \includegraphics[width=0.90\linewidth]{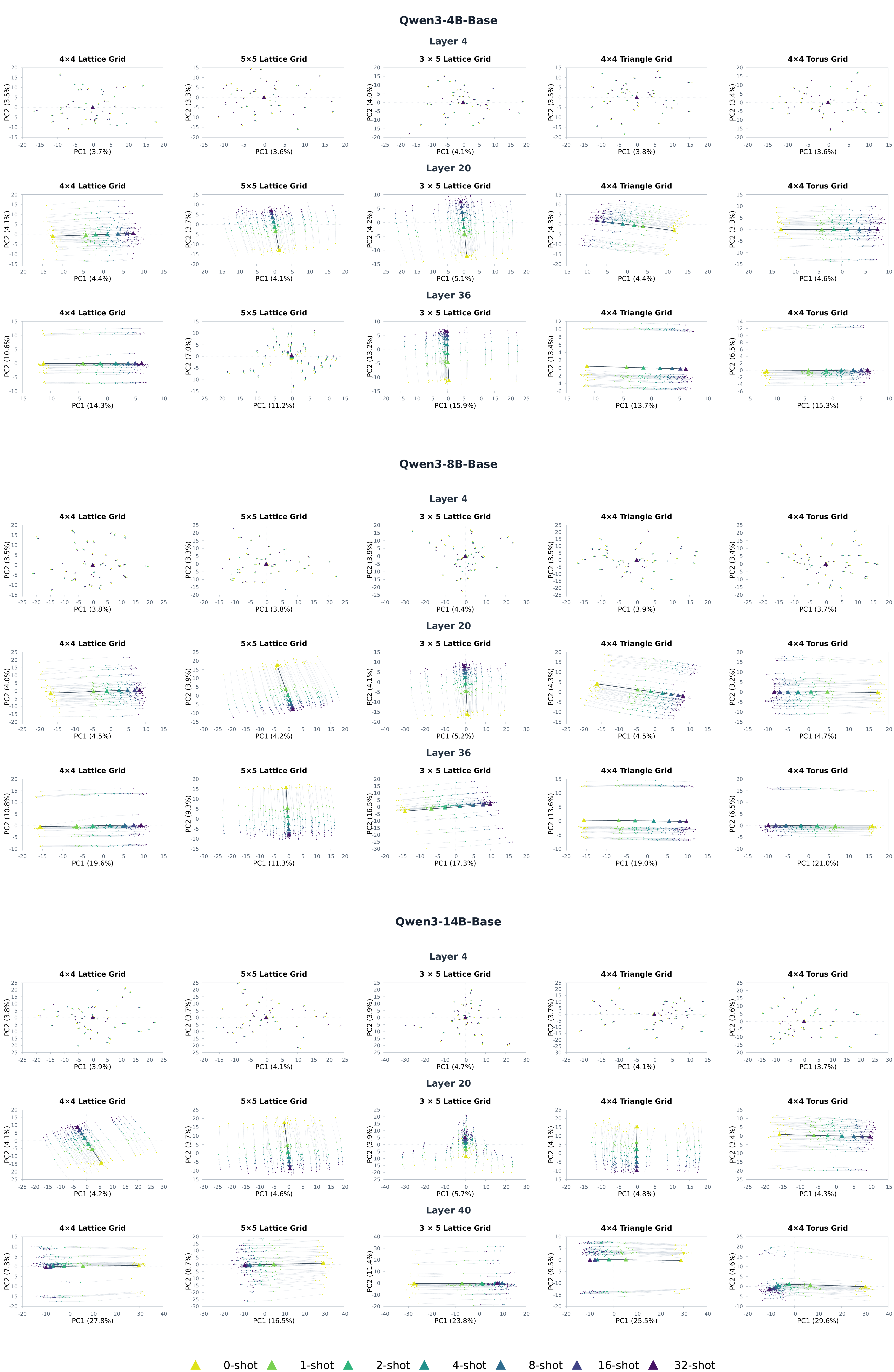}
  \caption{\textbf{PCA of Shifts in Hidden-State Centroids across Random Walks Across Different Number of Shots.} PCA of 300 test samples per shot. Points represent individual random walks and triangles indicate their means. In each panel, PCA is fit jointly to all activations shown. Results are from three Qwen3 models on five graph topologies.}
  \label{fig:pca-example}
\end{figure}

\Cref{fig:relative-node-position} suggests that, as long as the query remains the same, increasing the number of shots does not substantially change the relative positions of nodes in the graph representation. The similarity to the 0-shot setting remains nearly constant as the number of shots increases, and the variation is small compared with the case where different fruit vocabularies are assigned to nodes with the same topology. The PCA visualization on the left of \Cref{fig:shot-vector-illustration} also visualizes that the relative node configuration remains largely unchanged.

\clearpage
\subsection{Shot Vector}
\begin{figure}[h]
  \centering
  \includegraphics[width=0.95\linewidth]{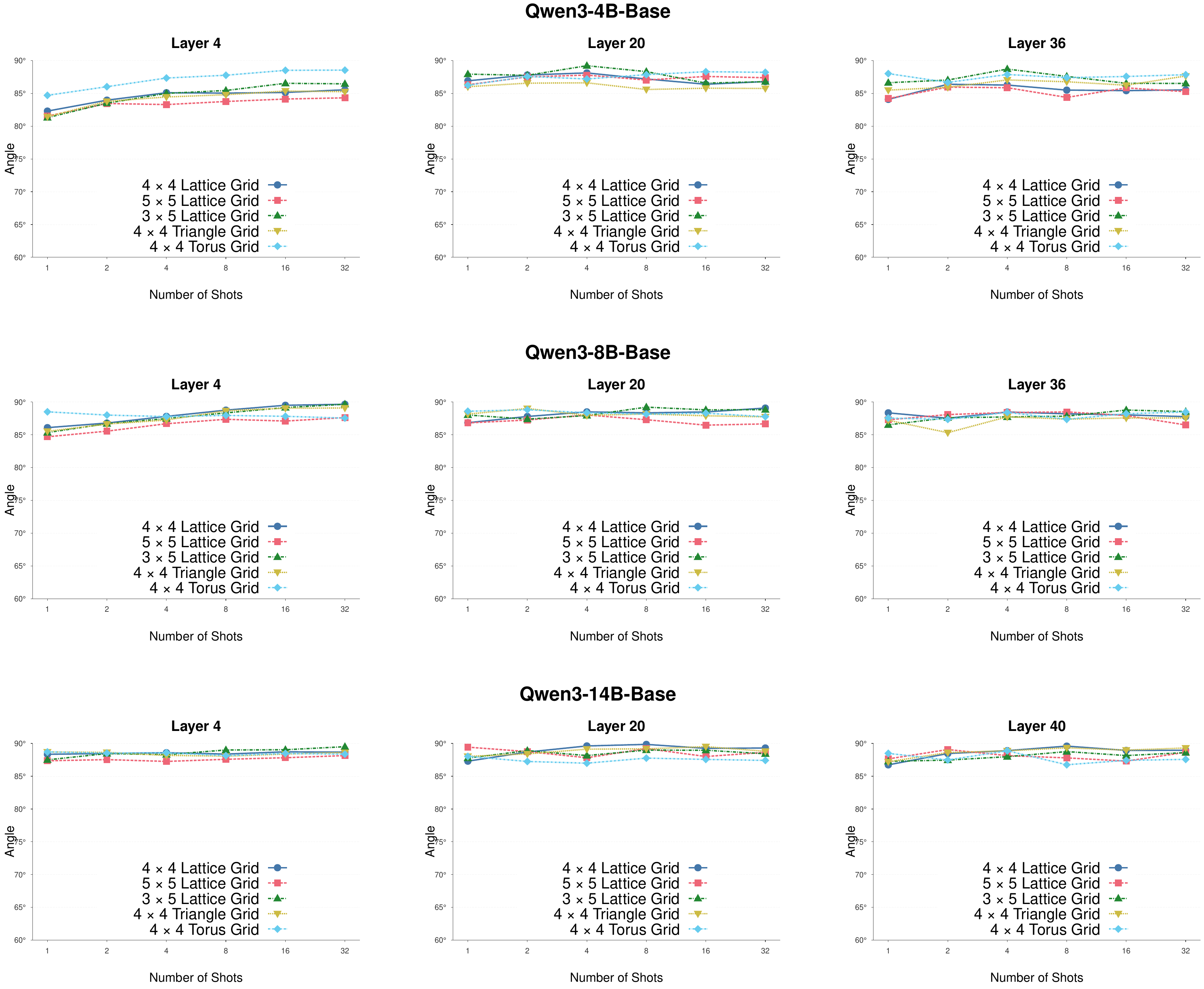}
  \caption{\textbf{Angle Between the World Representation Subspace and the Shot Vector.} Angles to the corresponding probe subspaces, shown separately for each topology. Results are from three Qwen3 models on five graph topologies.}
  \label{fig:fs-zs}
\end{figure}

\begin{figure}[h]
  \centering
  \includegraphics[width=0.95\linewidth]{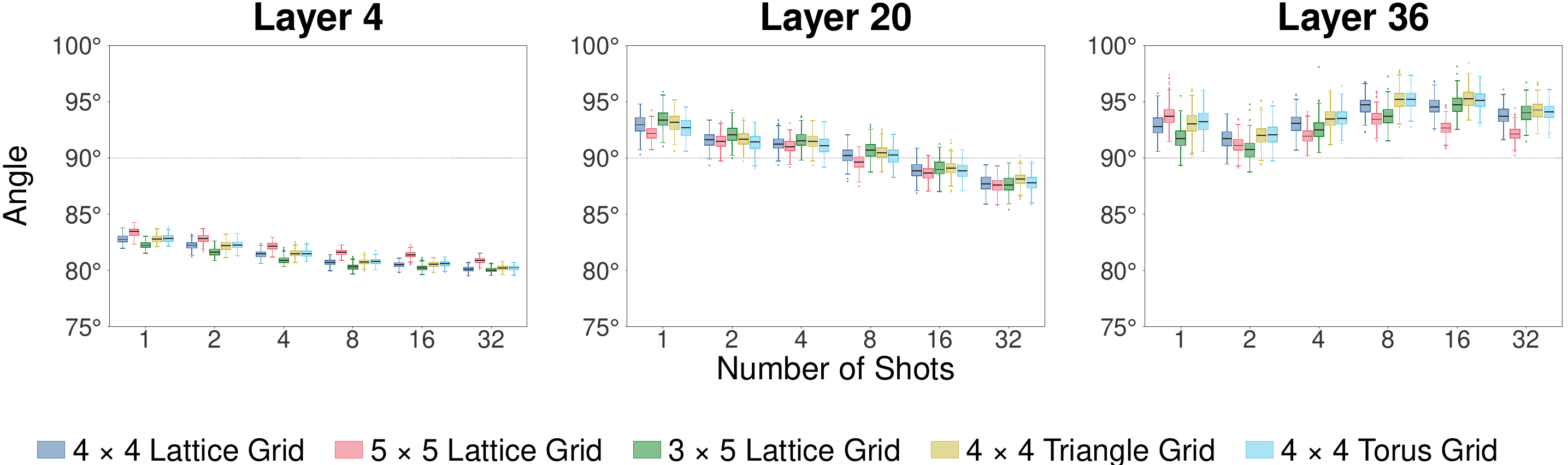}
  \caption{\textbf{Angle Between the Shot Vector and the Task Vector.} The angle between the two vectors is in degrees from $0^\circ$ to $180^\circ$. Results from Qwen3-4B-Base on five topologies.}
  \label{fig:shot-task}
\end{figure}

\label[appendix]{appendix:shot-vector}
We define the Shot Vector $\Delta \bar{h}^{(l)}$ as the mean difference between few shot and zero shot hidden states for corresponding nodes in the query random walk, computed over the test set. \Cref{fig:fs-zs} reports detailed results on the overlap between the Shot Vector and the World Representation Subspace. We also show in \Cref{fig:shot-task} that the Task Vector \citep{hendel2023context}, defined as the few shot hidden state immediately before the answer ("\texttt{\textbackslash n}" in our graph tracking task), is distinct from the Shot Vector and nearly orthogonal to it. Note that \citet{hendel2023context} define the Task Vector for tasks with input--output pairs, including algorithmic, translation, linguistic, and knowledge tasks. This differs from our few-shot setting, where each demonstration consists of an observation, few-shot examples as an instruction, and an answer. Nevertheless, we extract the Task Vector using the same definition.

\begin{figure}[h]
  \centering
\includegraphics[width=0.98\linewidth]{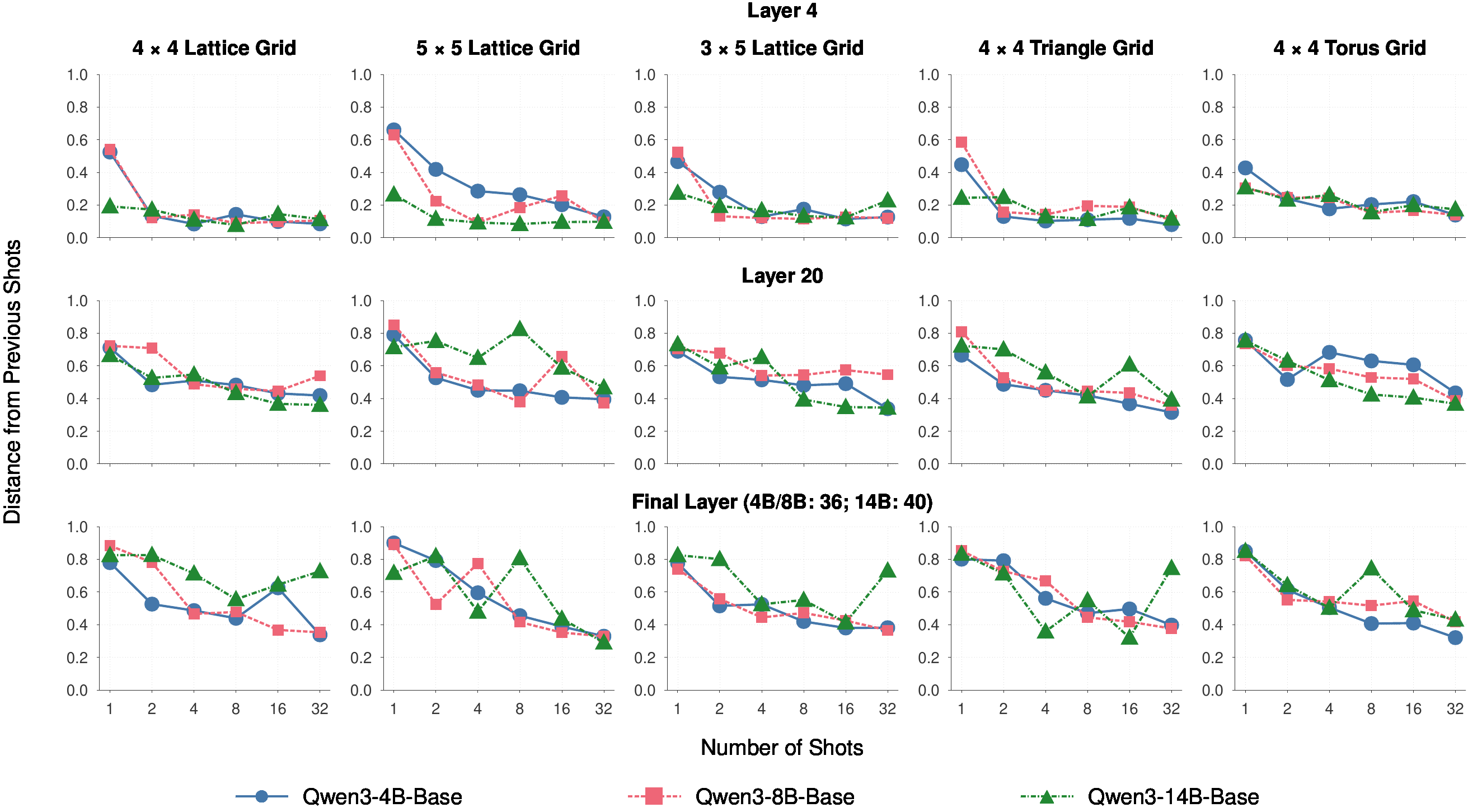}
  \caption{\textbf{Representation Subspace Distance Results.} Normalized projection distance between probe subspaces at consecutive tested shot counts, shown separately for each topology. Results are from three Qwen3 models on five graph topologies.}
  \label{fig:distance}
\end{figure}

\rev{
\subsection{Probing and Representation of Instructions}\label[appendix]{appendix:probe-representation-instruction}
}

\rev{In addition to measuring how well random walk nodes align with the graph topology (\Cref{fig:auroc-detail}), we report the normalized DE of instruction nodes in few-shot demonstrations across varying number of shots in \Cref{fig:de-instruction}. As discussed in \Cref{section:representation}, the normalized DE for random walk nodes remains largely constant regardless of number of shots. In contrast, for instruction nodes across four models and three topologies, it decreases as the number of shots increases. This suggests that few-shot demonstrations encourage instruction nodes to align more closely with the latent graph topology.}
\rev{
\begin{figure}[h]
  \centering
  \includegraphics[width=0.95\linewidth]{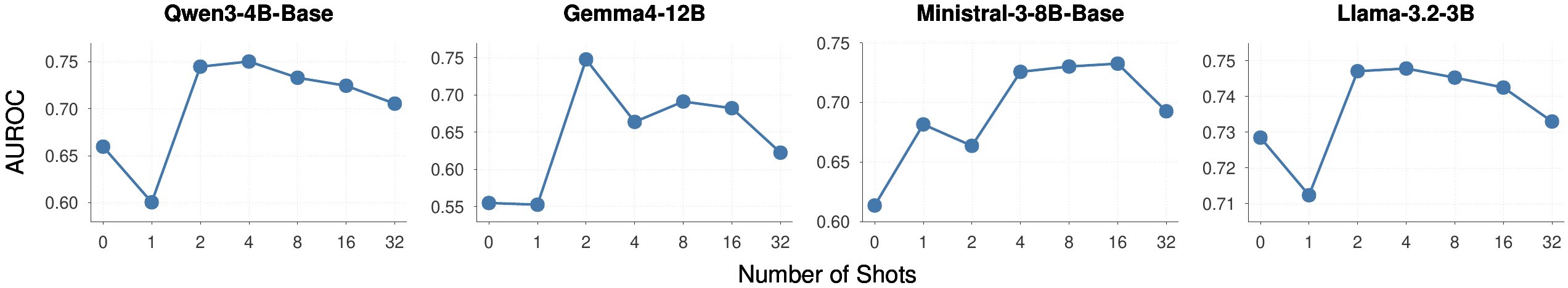}
  \caption{\rev{\textbf{Cross Probe Performance on Instructions.}
  Probe AUROC across shots for the final layer of four models on the $4 \times 4$ lattice grid. Probes trained on random walk in the query at each shot count are applied to instruction representations at the same shot count. Each condition is evaluated on 300 test samples.}}
  \label{fig:cross-probe-instruction}
\end{figure}
}
\rev{
\begin{figure}[h]
  \centering
  \includegraphics[width=0.9\linewidth]{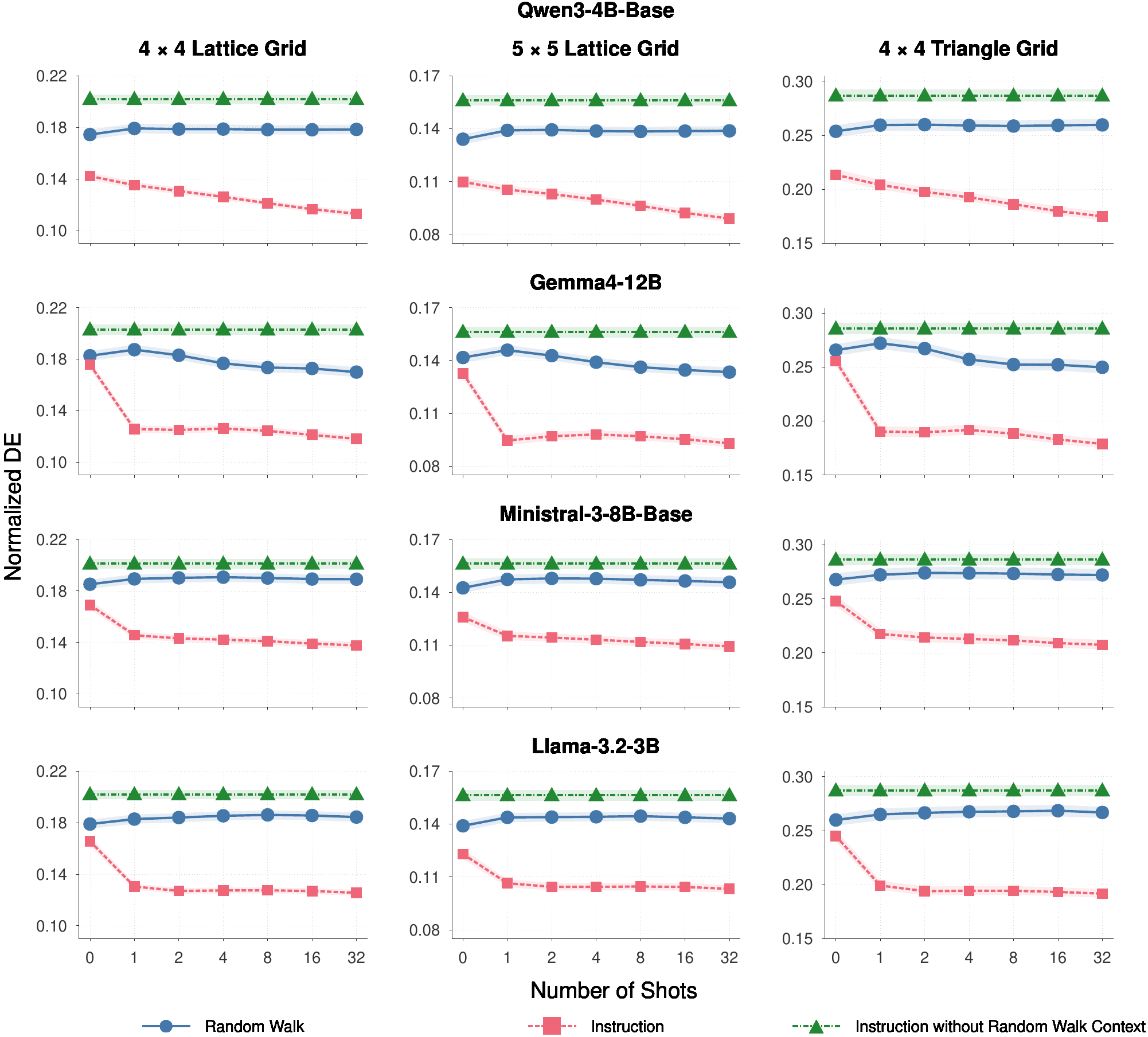}
  \caption{\rev{\textbf{Normalized Dirichlet Energy (DE) of Instruction Across Different Number of Shots.} Normalized DE of final layer across shots and three graph topologies for four models. Blue and red curves correspond to representations from the query random walk and instruction examples, respectively. Green curves show an uncontextualized baseline without random walk context. Each condition is evaluated on 300 test samples.}}
  \label{fig:de-instruction}
\end{figure}
}
\rev{Moreover, \Cref{fig:cross-probe-instruction} reports the cross probe performance when applying a probe layer, trained on query random walks, to instructions across different number of shots. Compared to the 0-shot setting, providing few-shot demonstrations improves AUROC. Together with \Cref{fig:de-instruction}, this suggests that few-shot demonstrations lead the hidden states of fruit vocabulary in instructions to align more strongly with the graph topology behind the random walk.}

\subsection{Intervention}
\begin{figure}[h]
  \centering
  \includegraphics[width=\linewidth]{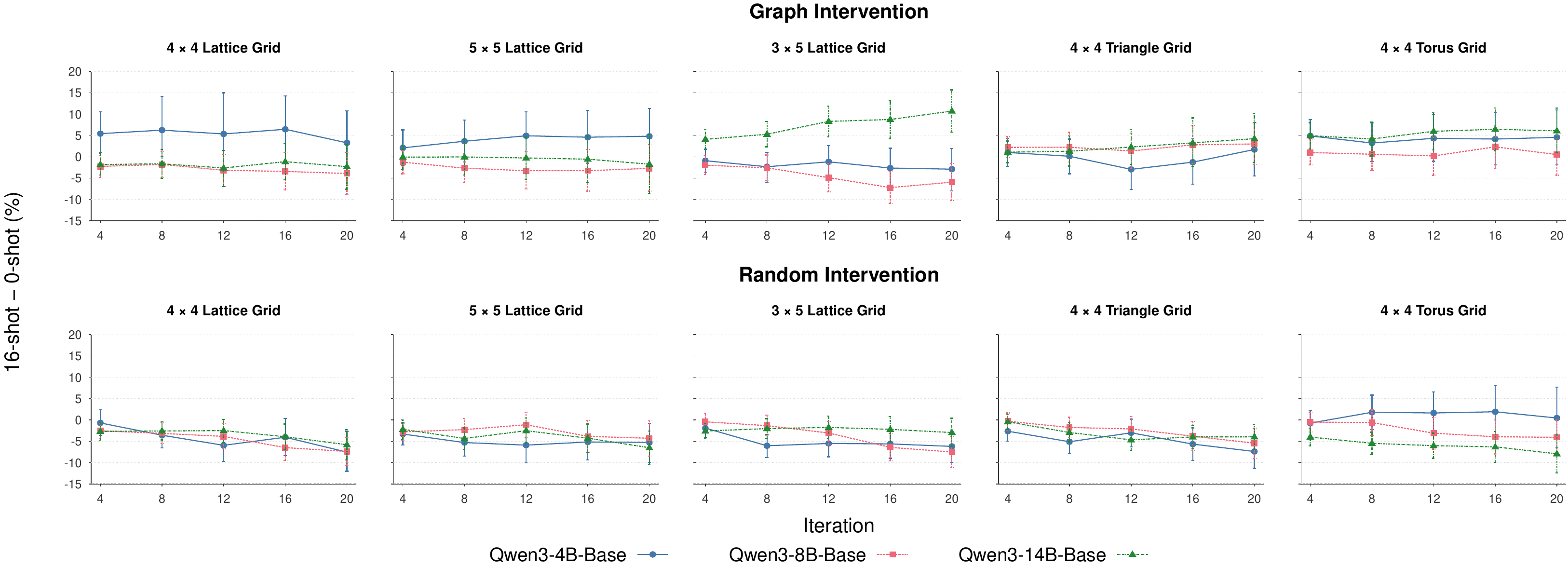}
  \caption{\textbf{Intervention Results across Intervention Iterations.} Plotted points represent the average paired difference in this degradation between the 16-shot and 0-shot settings across 300 samples in test set. The top is iterative nullspace projection targeted to graph structure, and the bottom is random orthogonal subspace intervention matched in rank. For each prompt condition, relative degradation is computed per test sample as the drop in the model's assigned probability for the correct answer after intervention, normalized by the probability before intervention. Positive values indicate that intervention causes a larger relative performance drop under 16-shot prompting than under 0-shot setting.}
  \label{fig:intervention-details}
\end{figure}

\label[appendix]{appendix:intervention}
We report the degradation in probability of the correct answer token caused by graph intervention and random intervention across five topologies for three Qwen3 models in \Cref{fig:intervention-details}. Graph intervention tends to cause larger degradation in the 16-shot setting than in the 0-shot setting, whereas random intervention shows the opposite trend.

\begin{figure}[h]
  \centering
  \includegraphics[width=\linewidth]{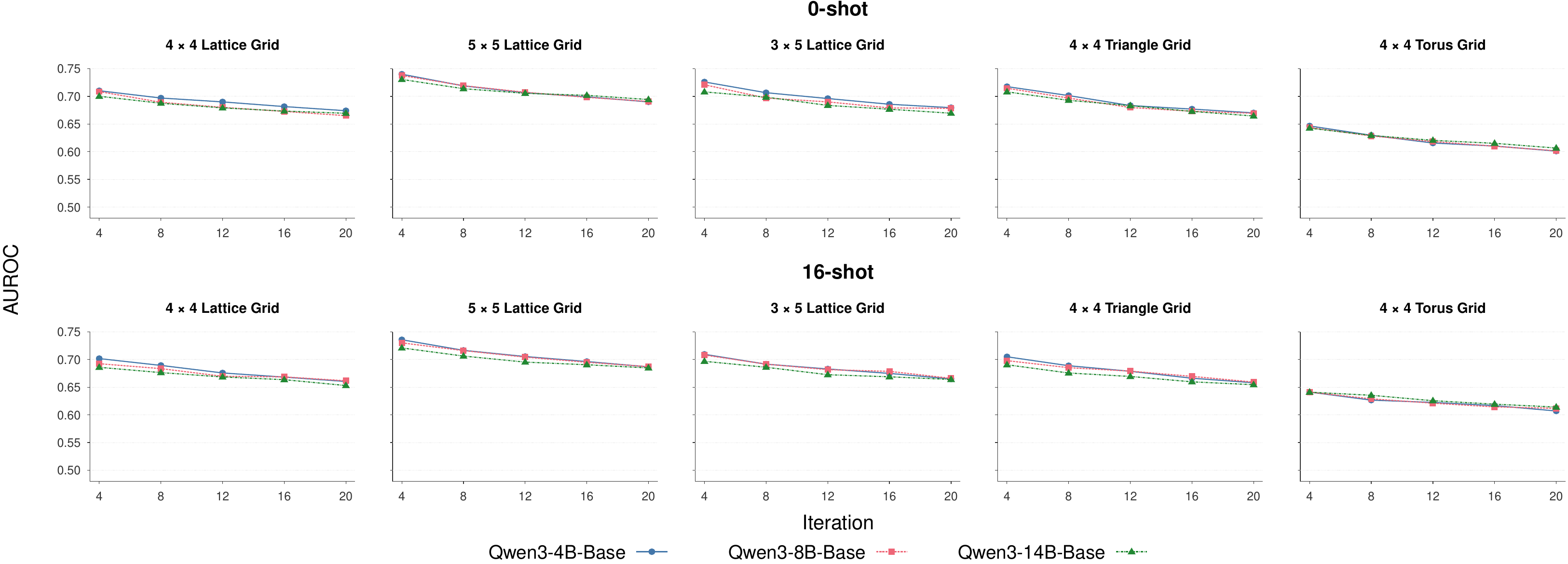}
  \caption{\textbf{Probe Performance during Intervention Iterations.} At each cumulative intervention step (4, 8, 12, 16, and 20), a new linear probe is trained on the intervened residual stream activations and evaluated on a held-out test set. Plotted points indicate the average of test AUROC across the layers 2–36 for Qwen3-4B-Base and Qwen3-8B-Base, and layers 2–40 for Qwen3-14B-Base.}
  \label{fig:intervention-probe}
\end{figure}

Test probe AUROC \Cref{fig:intervention-probe} also gradually decreases over the intervention iterations. Our intervention removes graph information by subtracting the component of each centered node representation that lies in the learned world representation subspace. Since we intervene on a very low dimensional subspace with $r=2,4$ relative to the ambient dimension, the effect on the representation is limited. We therefore apply the intervention multiple times, following \citep{elazar2021amnesic}.

\paragraph{\rev{Hypothesis Testing.}}
\rev{We tested whether the difference between graph and random intervention effects changes from 0-shot to 16-shot. We measured the reduction in correct answer sequence log probability for the same 300 samples under both shot conditions. Let $G_{i,s}$ and $R_{i,s}$ denote the reductions under graph and random interventions for sample $i$ at a number of shot $s$. We defined the interaction difference and its mean as $D_i=(G_{i,16}-R_{i,16})-(G_{i,0}-R_{i,0}), \quad \Delta = {1}/{300}\sum_{i=1}^{300}D_i$.}
\rev{We tested $H_0:\mathbb{E}[D_i]=0$ using a two-sided one-sample $t$-test on $D_i$. The test statistic was $t=\Delta/(s_D/\sqrt{300})$, where $s_D$ is the sample standard deviation of $D_i$, with 299 degrees of freedom. We applied Holm correction across the 15 model and topology sets, with an adjusted significance threshold of 0.05.}
\rev{All 15 comparisons had positive interaction estimates, and 12 remained significant. This indicates an increase in the graph intervention effect relative to random intervention from 0-shot to 16-shot.}

\begin{table*}[t]
\centering
\begingroup\color{black}
\caption{\rev{\textbf{Absolute Effects of Graph and Random Interventions,
and Their Interaction with the Number of Shots.}
All effects are mean reductions in the correct answer sequence log probability.
$p_{\mathrm{Holm}}$ denotes p-values from two-sided paired $t$-tests,
adjusted using Holm's method across 15 comparisons. Boldface denotes
statistical significance at $p_{\mathrm{Holm}}<0.05$.}}
\label{tab:absolute-intervention-shot-interaction}
\small
\setlength{\tabcolsep}{5pt}
\renewcommand{\arraystretch}{1.12}
\vspace{4pt}
\scalebox{0.88}{%
\begin{tabular*}{\textwidth}{@{\extracolsep{\fill}}llrrrrrr@{}}
\toprule
 & & \multicolumn{2}{c}{0-Shot} & \multicolumn{2}{c}{16-Shot} & & \\
\cmidrule(lr){3-4}\cmidrule(lr){5-6}
Model & Topology & Graph & Random & Graph & Random & $\Delta$ & $p_{\mathrm{Holm}}$ \\
\midrule
Qwen3-4B & $4\times4$ Lattice & $0.1166$ & $0.2206$ & $0.1061$ & $0.0522$ & $+0.1578$ & $\boldsymbol{7.2481\times10^{-5}}$ \\
 & $5\times5$ Lattice & $0.0880$ & $0.2067$ & $0.1140$ & $0.0587$ & $+0.1740$ & $\boldsymbol{6.0502\times10^{-6}}$ \\
 & $3\times5$ Lattice & $0.1445$ & $0.1967$ & $0.1173$ & $0.0484$ & $+0.1212$ & $\boldsymbol{0.00066}$ \\
 & $4\times4$ Triangle & $0.1067$ & $0.2090$ & $0.0703$ & $0.0622$ & $+0.1104$ & $\boldsymbol{0.00336}$ \\
 & $4\times4$ Torus & $0.2125$ & $0.5120$ & $0.2196$ & $0.2376$ & $+0.2815$ & $\boldsymbol{7.2695\times10^{-9}}$ \\
\midrule
Qwen3-8B & $4\times4$ Lattice & $0.1291$ & $0.1237$ & $0.0502$ & $0.0234$ & $+0.0213$ & $0.86015$ \\
 & $5\times5$ Lattice & $0.0710$ & $0.0808$ & $0.0100$ & $0.0126$ & $+0.0071$ & $0.86015$ \\
 & $3\times5$ Lattice & $0.1090$ & $0.1146$ & $0.0305$ & $0.0044$ & $+0.0317$ & $0.72732$ \\
 & $4\times4$ Triangle & $0.0418$ & $0.1295$ & $0.0444$ & $0.0452$ & $+0.0869$ & $\boldsymbol{0.01596}$ \\
 & $4\times4$ Torus & $0.1879$ & $0.2642$ & $0.1680$ & $0.1350$ & $+0.1094$ & $\boldsymbol{0.00336}$ \\
\midrule
Qwen3-14B & $4\times4$ Lattice & $0.0228$ & $0.1018$ & $0.0258$ & $0.0388$ & $+0.0660$ & $\boldsymbol{0.04426}$ \\
 & $5\times5$ Lattice & $-0.0230$ & $0.0853$ & $-0.0039$ & $0.0128$ & $+0.0915$ & $\boldsymbol{0.01412}$ \\
 & $3\times5$ Lattice & $-0.0223$ & $0.0694$ & $0.0828$ & $0.0181$ & $+0.1562$ & $\boldsymbol{5.7289\times10^{-8}}$ \\
 & $4\times4$ Triangle & $-0.0126$ & $0.0992$ & $0.0364$ & $0.0456$ & $+0.1026$ & $\boldsymbol{0.00122}$ \\
 & $4\times4$ Torus & $0.0964$ & $0.2259$ & $0.1665$ & $0.1346$ & $+0.1614$ & $\boldsymbol{6.0502\times10^{-6}}$ \\
\bottomrule
\end{tabular*}}
\endgroup
\end{table*}

\rev{\subsection{Attention Analysis}}\label[appendix]{appendix:attention}
\rev{For each layer and attention head, we analyze attention from the final input position, which predicts the first answer token. Let $S_{\mathrm{walk}}$ and $S_{\mathrm{inst}}$ denote the random walk tokens and the six instruction examples (instruction tokens), respectively. We compute the \textit{random walk share} of attention weights as $(\sum_{j \in S_{\mathrm{walk}}} \alpha_j) / (\sum_{j \in S_{\mathrm{walk}}} \alpha_j + \sum_{j \in S_{\mathrm{inst}}} \alpha_j)$ and the random walk share of vector norms as $(\sum_{j \in S_{\mathrm{walk}}}| \alpha_j W_O^{(h)} v_j |_2) / (\sum_{j \in S_{\mathrm{walk}}}| \alpha_j W_O^{(h)} v_j |_2 + \sum_{j \in S_{\mathrm{inst}}}| \alpha_j W_O^{(h)} v_j |_2)$, following \citet{kobayashi2020attention}, where $\alpha_j$ is the attention weight, $v_j$ is the value vector for head $h$, and $W_O^{(h)}$ is its output projection.}

\rev{We calculate these ratios for each head and input, then average them across heads and 300 samples per topology, separately for the 0-shot and 16-shot settings. We measure attention at every layer and contribution norms at selected layers, before residual addition and any post-attention normalization.}

\begin{figure}[h]
  \centering
  \includegraphics[width=\linewidth]{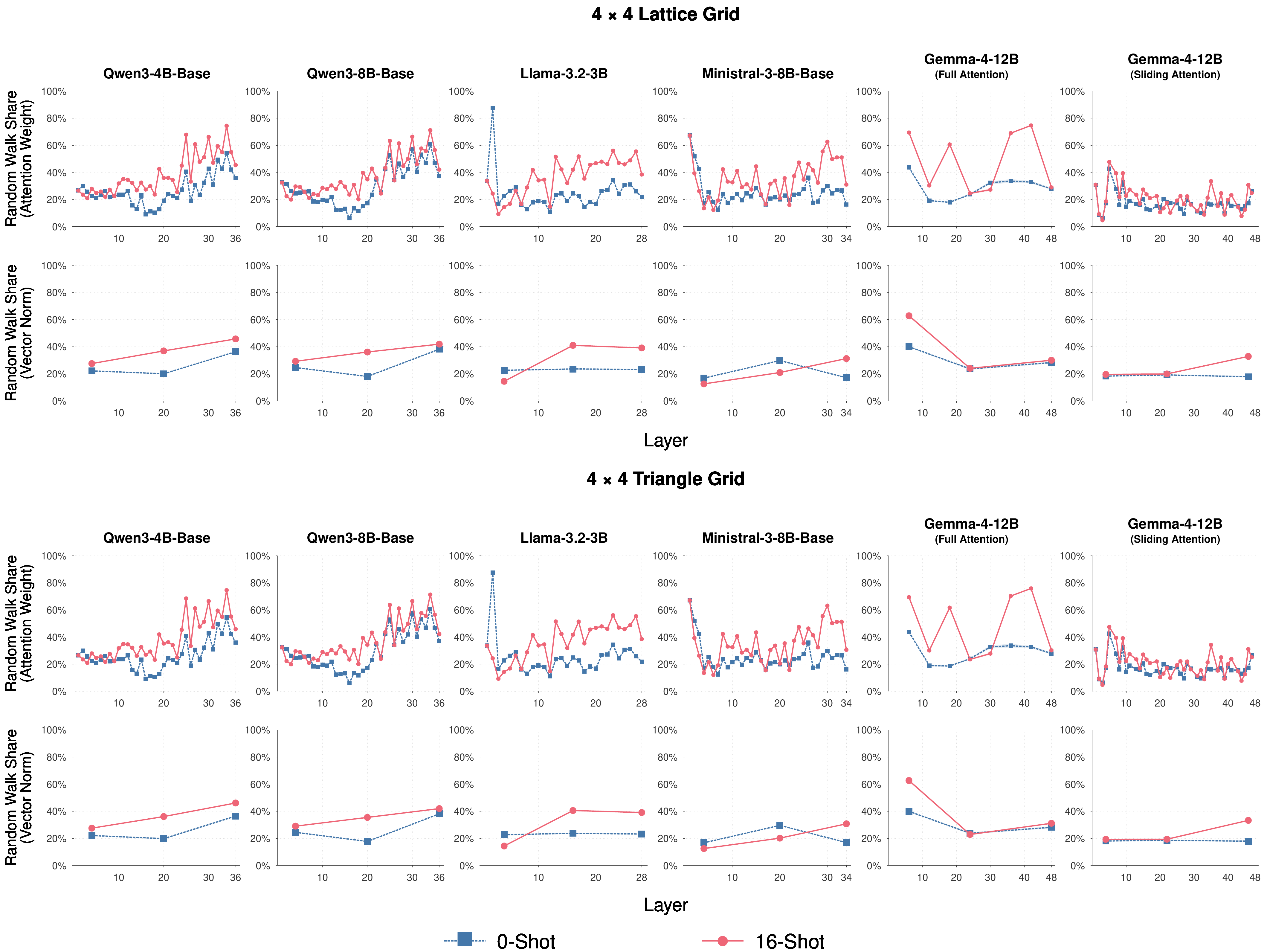}
  \caption{\rev{\textbf{Attention to Random Walk and Instruction Tokens.} Random walk share at the answer prediction position across five models and two graph topologies. Upper rows show attention to random walk tokens divided by attention to random walk and instruction tokens within the query. Lower rows show the corresponding share of projected value contribution norms at selected layers. Blue and red lines indicate 0 shot and 16 shot settings. Results average per head ratios over 300 test samples. For Gemma-4-12B, full and sliding attention layers are shown separately.}}
  \label{fig:attention-appendix}
\end{figure}

\rev{As shown in \Cref{fig:attention-appendix}, the random walk share (the fraction of attention or norm contributions assigned to random walk tokens relative to random walk and instruction tokens when predicting the first answer token (the final input token position)) is higher in the 16-shot than in the 0-shot setting, especially in the middle and later layers, for both attention weights and vector norms. This suggests that few-shot demonstrations shift the model's focus toward the random walk representations which we have probed in this paper.}

\section{Additional Experiments}

\subsection{Experiments on Llama, Gemma, and Ministral}
\begin{figure}[h]
  \centering
  \includegraphics[width=\linewidth]{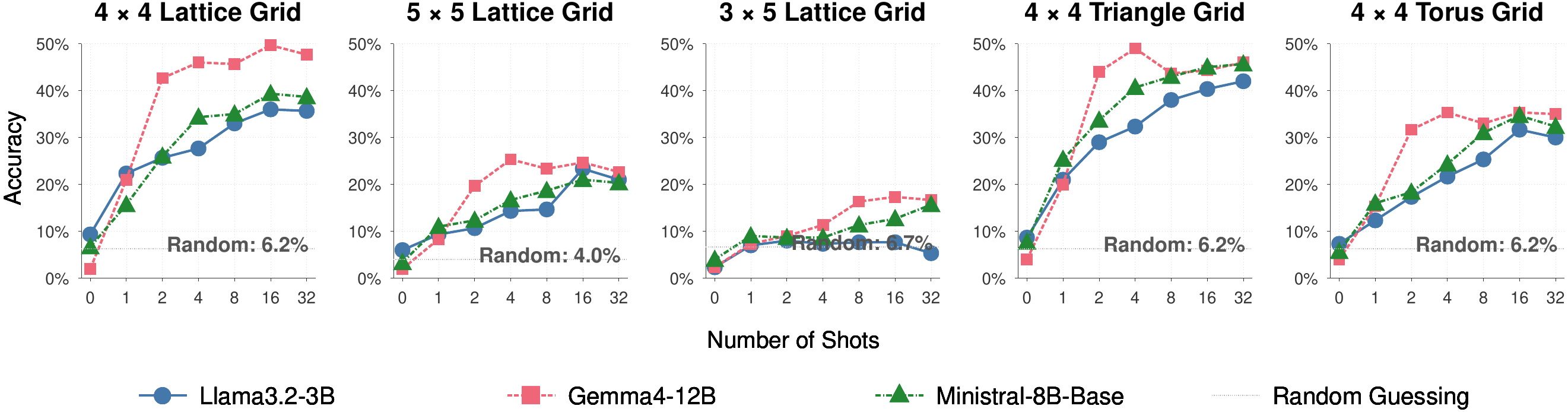}
  \caption{\textbf{Few-Shot Performance.} Accuracy of Llama-3.2-3B, Gemma-4-12B, and Ministral-3-8B-Base-2512 on the two-step-right prediction task across five graph topologies. Each point reports accuracy on 300 test samples with 0, 1, 2, 4, 8, 16, or 32 few-shot demonstrations. Dotted gray lines indicate random guessing over the candidate nodes for each topology.}
  \label{fig:few-shot-performance-additional}
\end{figure}

\begin{figure}[h]
  \centering
  \includegraphics[width=0.95\linewidth]{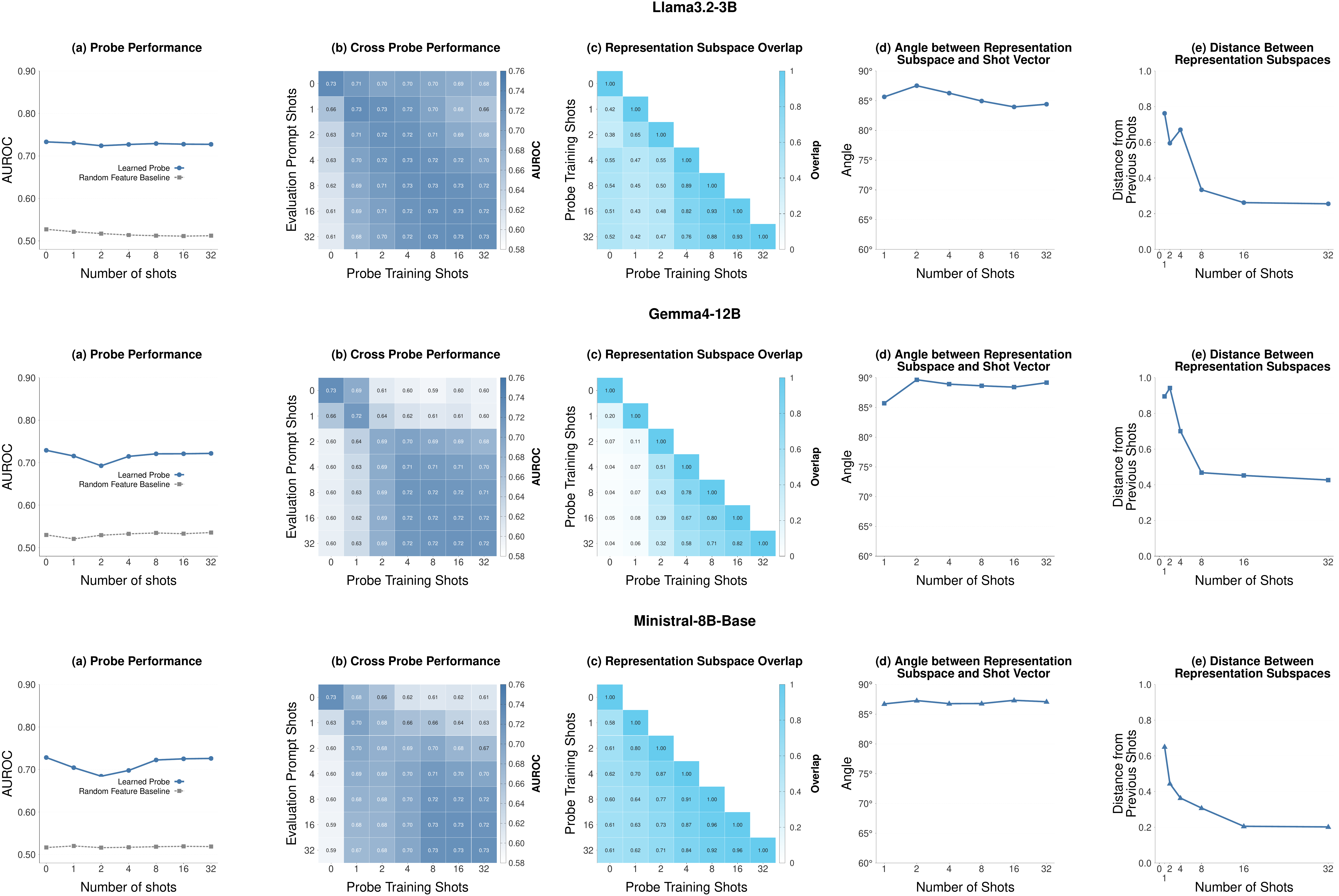}
  \caption{\textbf{(a) Probe Performance.} Probe AUROC. Gray lines show random feature probe baselines. \textbf{(b) Cross Probe Performance.} AUROC of probes trained at one shot count and evaluated at another. \textbf{(c) World Representation Subspace Overlap.} Overlap between probe subspaces across shot counts. \textbf{(d) Angle Between Shot Vectors and the World Representation Subspace.} Angles to the corresponding probe subspaces. \textbf{(e) Distance between Representation Subspaces across Shots.} Normalized projection distance between probe subspaces at consecutive tested shot counts. Results are from layer 28 of Llama-3.2-3B, layer 48 of Gemma-4-12B, and layer 34 of Ministral-3-8B-Base-2512 on $4\times4$ Lattice Grid.}
  \label{fig:relocation-additional}
\end{figure}

\begin{figure}[h]
  \centering
  \includegraphics[width=0.80\linewidth]{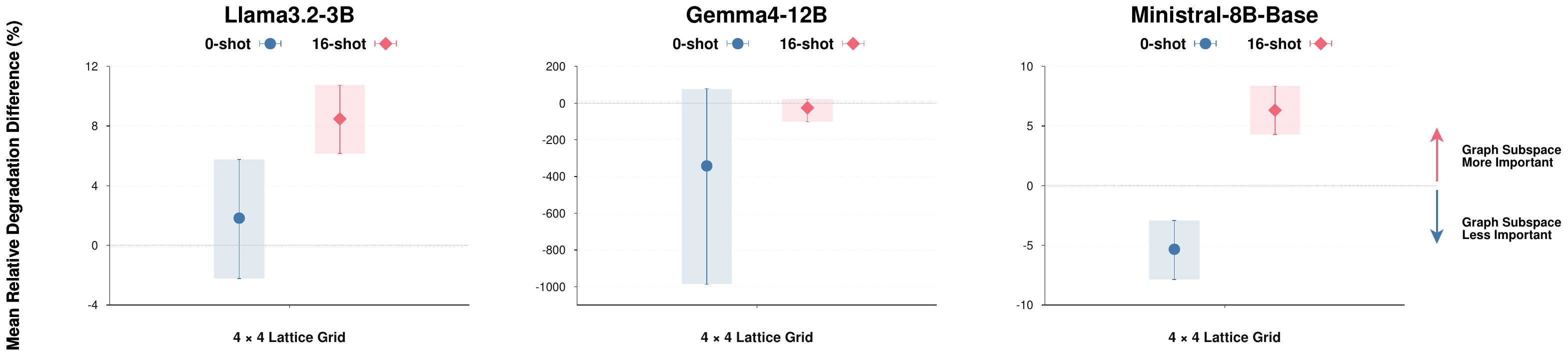}
  \caption{\textbf{Intervention Results.} Graph subspace erasure at iteration 20 in Llama-3.2-3B, Gemma-4-12B, and Ministral-3-8B-Base-2512 on $4\times4$ Lattice Grid. Interventions target node subtokens in the final query’s walk at every Transformer block output. The random intervention uses an orthogonal subspace with matched rank and perturbation norm per node. Points show the mean difference in correct answer probability reduction between graph and random interventions, normalized by each query’s unperturbed probability, across 300 test samples. Probabilities are normalized over candidate answer sequence likelihoods. Positive values indicate greater degradation from graph erasure. Horizontal bars show 95\% bootstrap confidence intervals obtained by resampling test samples 5000 times. Note that the wide confidence intervals for 0-shot Gemma-4-12B are because of near-zero baseline probabilities in the denominator.}
  \label{fig:intervention-additional}
\end{figure}

\label[appendix]{appendix:other-families}
In addition to the results for Qwen3-4B-Base, Qwen3-8B-Base, and Qwen3-14B-Base \citep{yang2025qwen3}, we report the few-shot performance of Llama-3.2-3B \citep{grattafiori2024llama3}, Gemma-4-12B \citep{gemmateam2026gemma4}, and Ministral-3-8B-Base-2512 \citep{liu2026ministral3} across the five topologies in \Cref{fig:few-shot-performance-additional}. We further report the results of our probing and representation analyses on the $4\times4$ lattice grid in \Cref{fig:relocation-additional}, and the intervention results in \Cref{fig:intervention-additional}. Across all of these experiments, we observe results consistent with the results from Qwen3.

\subsection{Experiments on Different Ranks}\label[appendix]{appendix:rank}

\begin{figure}[h]
  \centering
  \includegraphics[width=0.90\linewidth]{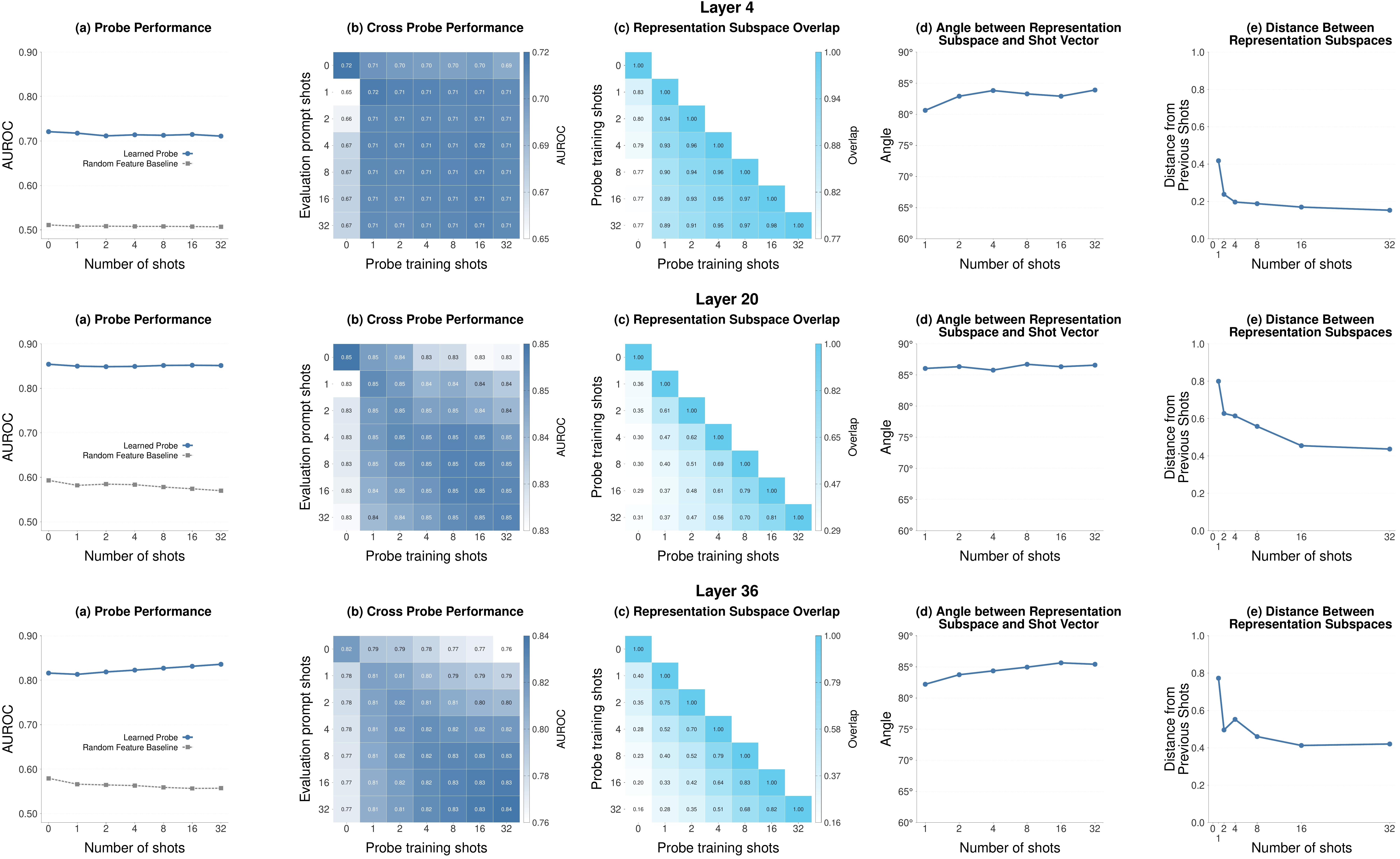}
  \caption{\textbf{(a) Probe Performance.} Probe AUROC. Gray lines show random feature probe baselines. \textbf{(b) Cross Probe Performance.} AUROC of probes trained at one shot count and evaluated at another. \textbf{(c) World Representation Subspace Overlap.} Overlap between probe subspaces across shot counts. \textbf{(d) Angle Between Shot Vectors and the World Representation Subspace.} Angles to the corresponding probe subspaces. \textbf{(e) Distance between Representation Subspaces across Shots.} Normalized projection distance between probe subspaces at consecutive tested shot counts. Results are from Qwen3-4B-Base on $4\times 4$ Lattice Grid, where $r=8$.}
  \label{fig:probe-rank-8}
\end{figure}

\begin{figure}[h]
  \centering
  \includegraphics[width=0.90\linewidth]{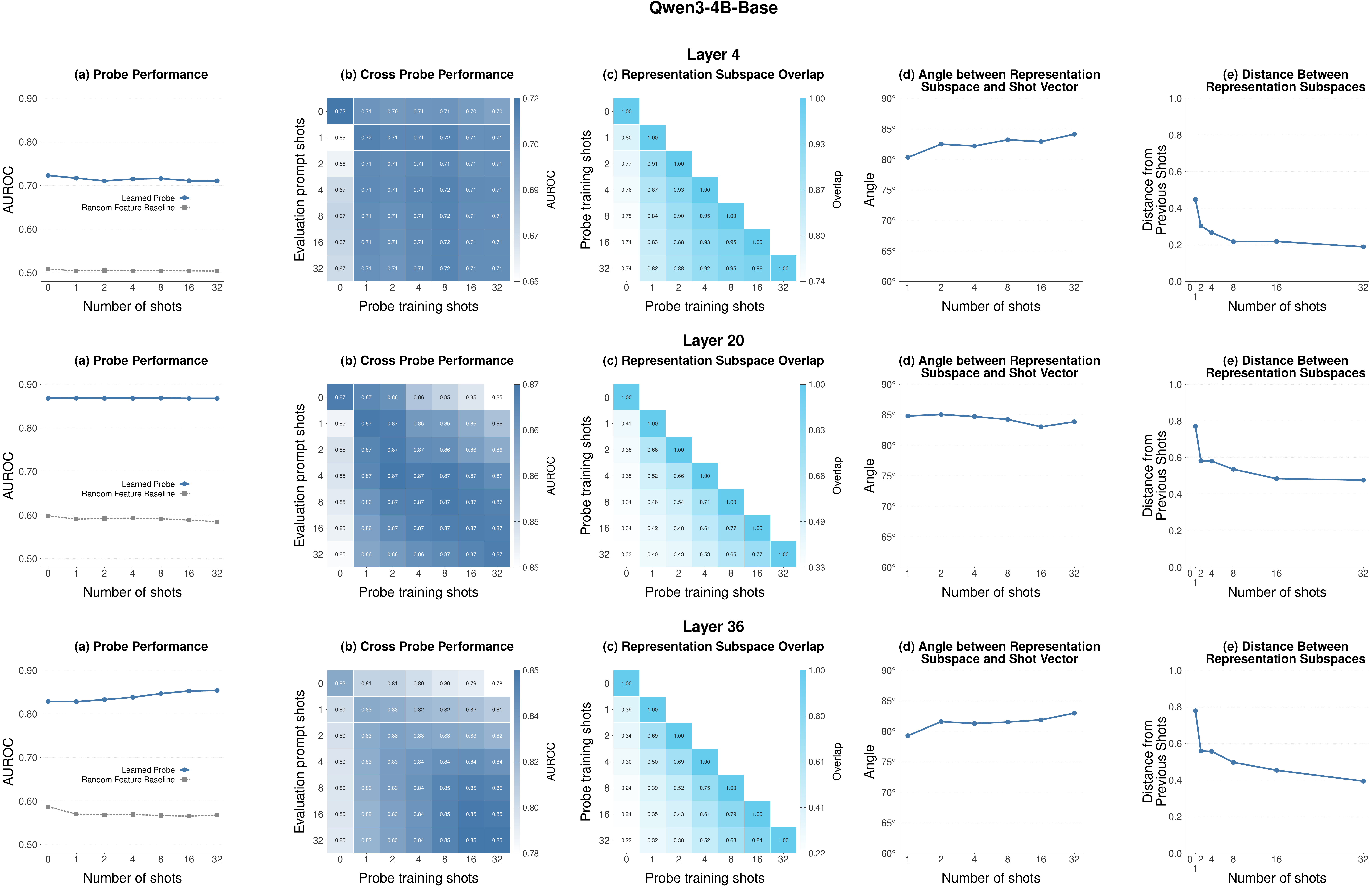}
  \caption{\textbf{(a) Probe Performance.} Probe AUROC. Gray lines show random feature probe baselines. \textbf{(b) Cross Probe Performance.} AUROC of probes trained at one shot count and evaluated at another. \textbf{(c) World Representation Subspace Overlap.} Overlap between probe subspaces across shot counts. \textbf{(d) Angle Between Shot Vectors and the World Representation Subspace.} Angles to the corresponding probe subspaces. \textbf{(e) Distance between Representation Subspaces across Shots.} Normalized projection distance between probe subspaces at consecutive tested shot counts. Results are from Qwen3-4B-Base on $4\times 4$ Lattice Grid, where $r=16$.}
  \label{fig:probe-rank-16}
\end{figure}

\begin{figure}[h]
  \centering
  \includegraphics[width=0.75\linewidth]{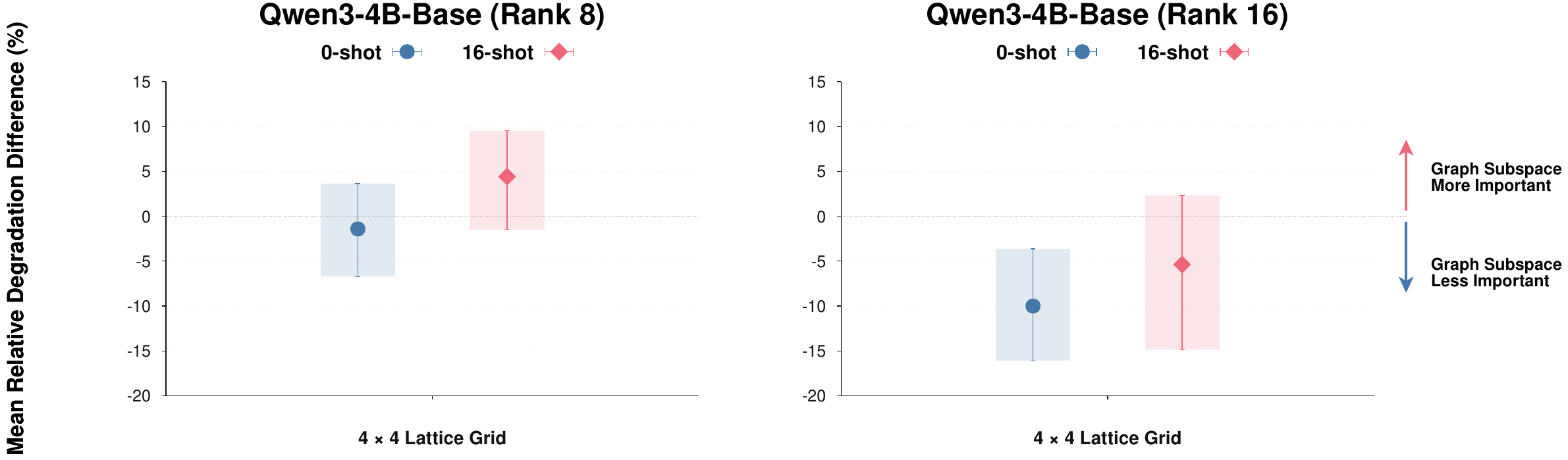}
  \caption{\textbf{Intervention Results.} Graph subspace erasure at iteration 4 in Qwen3-4B-Base on $4\times4$ Lattice Grid, where $r=8,16$. Interventions target node subtokens in the final query’s walk at every Transformer block output. The random intervention uses an orthogonal subspace with matched rank and perturbation norm per node. Points show the mean difference in correct answer probability reduction between graph and random interventions, normalized by each query’s unperturbed probability, across 300 test samples. Probabilities are normalized over candidate answer sequence likelihoods. Positive values indicate greater degradation from graph erasure. Horizontal bars show 95\% bootstrap confidence intervals obtained by resampling test samples 5000 times.}
  \label{fig:intervention-rank}
\end{figure}

For probing, we set the rank $r$ of the probe layer $W^{(l)} \in \mathbb{R}^{d \times r}$ to $2,2,2,2,$ and $4$ for the $4\times4$ Lattice Grid, $5\times5$ Lattice Grid, $3\times5$ Lattice Grid, $4\times4$ Triangle Grid, and $4\times4$ Torus Grid, respectively. We use $r=4$ for the $4\times4$ Torus Grid because encoding its two periodic coordinates with sine and cosine pairs yields a four-dimensional embedding that exactly separates adjacent and non-adjacent nodes by a Euclidean distance threshold. We use $r=2$ for the other grids because they admit two-dimensional coordinate representations. We additionally evaluate higher ranks, $r=8$ and $16$, on the $4\times4$ Lattice Grid with Qwen3-4B-Base. \Cref{fig:probe-rank-8,fig:probe-rank-16} report the probing and representation analysis results, and \Cref{fig:intervention-rank} reports the intervention results.

\subsection{Sensitivity to Random Seed}\label[appendix]{appendix:seed}
To quantify variability across seeds, we report results over five seeds for probing, representation analysis, and intervention on the $4\times4$ Lattice Grid with Qwen3-4B-Base.

\begin{table}[h]
\centering
\small
\setlength{\tabcolsep}{5pt}
\renewcommand{\arraystretch}{1.12}
\caption{\textbf{Probe Performance.} Results from Qwen3-4B-Base evaluated on 300 test samples on $4\times4$ Lattice Grid. “L4, 0-Shot” denotes the probe performance at Layer 4 with 0-shot prompting.}
\label{tab:qwen4b-five-seed-probe-auroc}
\begin{tabular}{@{}lrrrrrr@{}}
\toprule
Setting & Seed 1 & Seed 2 & Seed 3 & Seed 4 & Seed 5 & Mean $\pm$ Std \\
\midrule
L4, 0-Shot & 0.6903 & 0.7072 & 0.7060 & 0.7048 & 0.7013 & $0.7019 \pm 0.0069$ \\
L4, 16-Shot & 0.6867 & 0.6824 & 0.7046 & 0.6763 & 0.6941 & $0.6888 \pm 0.0109$ \\
L20, 0-Shot & 0.7812 & 0.7761 & 0.7792 & 0.7784 & 0.7777 & $0.7785 \pm 0.0019$ \\
L20, 16-Shot & 0.7664 & 0.7573 & 0.7591 & 0.7688 & 0.7722 & $0.7647 \pm 0.0064$ \\
L36, 0-Shot & 0.7675 & 0.7508 & 0.7629 & 0.7591 & 0.7656 & $0.7612 \pm 0.0066$ \\
L36, 16-Shot & 0.7407 & 0.7359 & 0.7419 & 0.7406 & 0.7412 & $0.7401 \pm 0.0024$ \\
\bottomrule
\end{tabular}
\end{table}

\begin{table}[h]
\centering
\small
\setlength{\tabcolsep}{5pt}
\renewcommand{\arraystretch}{1.12}
\caption{\textbf{Cross Probe Performance.} Results from Qwen3-4B-Base on $4\times4$ Lattice Grid. “L4, $0 \to 16$-Shot” denotes the Layer 4 probe trained on 500 0-shot samples and evaluated on 300 16-shot samples.
}
\label{tab:qwen4b-five-seed-cross-probe-auroc}
\begin{tabular}{@{}lrrrrrr@{}}
\toprule
Setting & Seed 1 & Seed 2 & Seed 3 & Seed 4 & Seed 5 & Mean $\pm$ Std \\
\midrule
L4, $0 \to 16$-Shot & 0.6406 & 0.6487 & 0.6525 & 0.6483 & 0.6425 & $0.6465 \pm 0.0049$ \\
L4, $16 \to 32$-Shot & 0.6857 & 0.6821 & 0.7036 & 0.6755 & 0.6942 & $0.6882 \pm 0.0109$ \\
L20, $0 \to 16$-Shot & 0.7191 & 0.7112 & 0.7028 & 0.7136 & 0.7061 & $0.7106 \pm 0.0064$ \\
L20, $16 \to 32$-Shot & 0.7669 & 0.7580 & 0.7592 & 0.7692 & 0.7731 & $0.7653 \pm 0.0065$ \\
L36, $0 \to 16$-Shot & 0.6329 & 0.6727 & 0.6551 & 0.6641 & 0.6538 & $0.6557 \pm 0.0148$ \\
L36, $16 \to 32$-Shot & 0.7428 & 0.7375 & 0.7439 & 0.7436 & 0.7415 & $0.7419 \pm 0.0026$ \\
\bottomrule
\end{tabular}
\end{table}

\begin{table}[h]
\centering
\small
\setlength{\tabcolsep}{5pt}
\renewcommand{\arraystretch}{1.12}
\caption{\textbf{Representation Subspace Overlap.} Results from Qwen3-4B-Base on $4\times4$ Lattice Grid. “L4, $0 \leftrightarrow 16$-Shot” denotes the overlap between the rank-2 representation subspaces of probe layers trained on 500 samples.
}
\label{tab:qwen4b-five-seed-subspace-overlap}
\begin{tabular}{@{}lrrrrrr@{}}
\toprule
Setting & Seed 1 & Seed 2 & Seed 3 & Seed 4 & Seed 5 & Mean $\pm$ Std \\
\midrule
L4, $0 \leftrightarrow 16$-Shot & 0.6295 & 0.8563 & 0.7353 & 0.7794 & 0.7285 & $0.7458 \pm 0.0826$ \\
L4, $16 \leftrightarrow 32$-Shot & 0.9804 & 0.9728 & 0.9681 & 0.9238 & 0.9928 & $0.9676 \pm 0.0262$ \\
L20, $0 \leftrightarrow 16$-Shot & 0.3329 & 0.4381 & 0.3081 & 0.3230 & 0.2904 & $0.3385 \pm 0.0580$ \\
L20, $16 \leftrightarrow 32$-Shot & 0.7970 & 0.7931 & 0.8709 & 0.8691 & 0.8247 & $0.8310 \pm 0.0376$ \\
L36, $0 \leftrightarrow 16$-Shot & 0.1776 & 0.2669 & 0.1812 & 0.1554 & 0.1601 & $0.1882 \pm 0.0453$ \\
L36, $16 \leftrightarrow 32$-Shot & 0.8654 & 0.8654 & 0.8993 & 0.8901 & 0.8861 & $0.8813 \pm 0.0153$ \\
\bottomrule
\end{tabular}
\end{table}

\begin{table}[h]
\centering
\small
\setlength{\tabcolsep}{5pt}
\renewcommand{\arraystretch}{1.12}
\caption{\textbf{Mean Relative Degradation Difference after Interventions.}
Results from Qwen3-4B-Base on the $4\times4$ Lattice Grid.
Each row reports the mean difference in correct answer probability reduction between graph and random interventions after 20 iterations for the corresponding shot setting, normalized by the corresponding clean probability before interventions.
Positive values indicate greater degradation under the graph intervention.}
\label{tab:qwen4b-five-seed-intervention-probability-pct}
\begin{tabular}{@{}lrrrrrr@{}}
\toprule
Setting & Seed 1 & Seed 2 & Seed 3 & Seed 4 & Seed 5 & Mean $\pm$ Std \\
\midrule
0-Shot  & -0.401 & -0.649 & -0.380 & -0.359 & -0.648 & $-0.487 \pm 0.148$ \\
16-Shot &  1.010 &  1.293 &  1.069 &  1.246 &  1.635 & $ 1.251 \pm 0.245$ \\
\bottomrule
\end{tabular}
\end{table}

\begin{table}[h]
\centering
\small
\setlength{\tabcolsep}{5pt}
\renewcommand{\arraystretch}{1.12}
\caption{\textbf{Probe Performance after Graph Interventions.} Results from Qwen3-4B-Base evaluated on 300 test samples on $4\times4$ Lattice Grid after graph intervention for 20 iterations. “L4, 0-Shot” denotes the probe performance at Layer 4 with 0-shot prompting.}
\label{tab:qwen4b-five-seed-audit-auroc-layer-mean}
\begin{tabular}{@{}lrrrrrr@{}}
\toprule
Setting & Seed 1 & Seed 2 & Seed 3 & Seed 4 & Seed 5 & Mean $\pm$ Std \\
\midrule
L4, 0-Shot & 0.5862 & 0.6337 & 0.5985 & 0.6015 & 0.5806 & $0.6001 \pm 0.0207$ \\
L4, 16-Shot & 0.5798 & 0.5897 & 0.5543 & 0.5758 & 0.5582 & $0.5716 \pm 0.0149$ \\
L20, 0-Shot & 0.7158 & 0.6900 & 0.6920 & 0.7095 & 0.7000 & $0.7015 \pm 0.0111$ \\
L20, 16-Shot & 0.6658 & 0.6783 & 0.6947 & 0.6891 & 0.6845 & $0.6825 \pm 0.0111$ \\
L36, 0-Shot & 0.6905 & 0.7208 & 0.7263 & 0.7059 & 0.7109 & $0.7109 \pm 0.0139$ \\
L36, 16-Shot & 0.6795 & 0.7056 & 0.6901 & 0.6915 & 0.6914 & $0.6916 \pm 0.0093$ \\
\bottomrule
\end{tabular}
\end{table}

\end{document}